\documentclass[lettersize,journal]{IEEEtran}
\usepackage{amsmath,amsfonts}
\usepackage{algorithmic}
\usepackage{algorithm}
\usepackage{array}
\usepackage{textcomp}
\usepackage{stfloats}
\usepackage{url}
\usepackage{verbatim}
\usepackage{graphicx}
\usepackage{cite}
\usepackage{booktabs}
\usepackage{threeparttable}
\usepackage{bm}
\usepackage{ulem}
\usepackage{comment}
\usepackage{balance}
\usepackage{amsmath}
\usepackage{graphicx}
\usepackage{multirow}
\usepackage{hyperref}
\usepackage{soul}
\usepackage{ulem}
\usepackage[caption=false,font=scriptsize,labelfont=sf,textfont=sf]{subfig}
\usepackage{listings}
\usepackage{color}
\usepackage{etoolbox}
\makeatletter
\patchcmd{\maketitle}
	{\@maketitle}
    {\@maketitle}
	{}
	{}
\makeatother
\usepackage{inconsolata}

\definecolor{dkgreen}{rgb}{0,0.6,0}
\definecolor{gray}{rgb}{0.5,0.5,0.5}
\definecolor{mauve}{rgb}{0.58,0,0.82}
\usepackage{ulem}
\usepackage{comment}
\usepackage{balance}
\usepackage{amsmath}
\usepackage{graphicx}
\usepackage{multirow}
\usepackage{amssymb}
\usepackage{algorithm, algorithmic}
\usepackage{hyperref}
\usepackage{soul}
\usepackage{ulem}
\usepackage[caption=false,font=scriptsize,labelfont=sf,textfont=sf]{subfig}
\usepackage{listings}
\usepackage{color}
\usepackage{etoolbox}
\makeatletter
\patchcmd{\maketitle}
	{\@maketitle}
    {\@maketitle}
	{}
	{}
\makeatother
\usepackage{inconsolata}

\definecolor{dkgreen}{rgb}{0,0.6,0}
\definecolor{gray}{rgb}{0.5,0.5,0.5}
\definecolor{mauve}{rgb}{0.58,0,0.82}
\usepackage[table,xcdraw,dvipsnames]{xcolor}
\usepackage{tabularx}

\begin{document}

\title{\Large \bf {\it SRLR}: Symbolic Regression based Logic Recovery to Counter\\ Programmable Logic Controller Attacks}

\author{
    \IEEEauthorblockN{Hao Zhou$^{a}$, Suman Sourav$^{b}$, Binbin Chen$^{c}$, Ke Yu$^{a}$}\\
    \IEEEauthorblockA{$^a$ Beijing University of Posts and Telecommunications, China}\\
    \IEEEauthorblockA{$^b$ Aalborg University, Denmark}\\
    \IEEEauthorblockA{$^c$ Singapore University of Technology and Design, Singapore}\\
    \IEEEauthorblockA{\{zhouh, yuke\}@bupt.edu.cn, sumansourav@cs.aau.dk, binbin\_chen@sutd.edu.sg}
    \thanks{
    This article was accepted by IEEE Transactions on Information Forensics and Security. Corresponding author: Ke Yu (yuke@bupt.edu.cn). Digital Object Identifier 10.1109/TIFS.2025.3634027
    
    ©2025 IEEE. Personal use of this material is permitted. Permission from IEEE must be obtained for all other uses, in any current or future media, including reprinting/republishing this material for advertising or promotional purposes, creating new collective works, for resale or redistribution to servers or lists, or reuse of any copyrighted component of this work in other works.
    }
}

\markboth{IEEE Transactions on Information Forensics and Security}%
{Hao Zhou \MakeLowercase{\textit{et al.}}: \Large \bf {\it SRLR}: Symbolic Regression based Logic Recovery to Counter\\ Programmable Logic Controller Attacks}


\maketitle

\begin{abstract}
Programmable Logic Controllers (PLCs) are critical components in Industrial Control Systems (ICSs). Their potential exposure to external world makes them susceptible to cyber-attacks. Existing detection methods against controller logic attacks use either specification-based or learnt models. However, specification-based models require experts' manual efforts or access to PLC's source code, while machine learning-based models often fall short of providing explanation for their decisions. We design {\it SRLR} --- a {\it Symbolic Regression based Logic Recovery} solution to identify the logic of a PLC based only on its inputs and outputs. The recovered logic is used to generate explainable rules for detecting controller logic attacks. 
{\it SRLR} enhances the latest deep symbolic regression methods using the following ICS-specific properties: (1) some important ICS control logic is best represented in frequency domain rather than time domain; (2) an ICS controller can operate in multiple modes, each using different logic, where mode switches usually do not happen frequently;
(3) a robust controller usually filters out outlier inputs as ICS sensor data can be noisy; and (4) with the above factors captured, the degree of complexity of the formulas is reduced, making effective search possible.
Thanks to these enhancements, {\it SRLR} consistently outperforms all existing methods in a variety of ICS settings that we evaluate. 
In terms of the recovery accuracy, {\it SRLR}'s gain can be as high as $39\%$ in some challenging environment. We also evaluate {\it SRLR} on a distribution grid containing hundreds of voltage regulators, demonstrating its stability in handling large-scale, complex systems with varied configurations.
\end{abstract}


\section{Introduction}
In the era of Industry 4.0, critical infrastructures are integrated with advanced communication networks and physical processes, leading to the emergence of diverse ICSs such as water treatment plants and power control systems. However, this integration in ICSs exposes industrial process to various vulnerabilities, thus precipitating severe security challenges~\cite{giraldo2018survey, zhu2024honeyjudge}. Internet-connected PLCs play a pivotal role in ICSs. High-profile attacks, such as Stuxnet~\cite{kushner2013real}, manipulate the configuration of PLCs or tamper with PLC's output commands to compromise physical processes.

Various Intrusion Detection Systems (IDSs) have been developed to detect malicious activities in ICSs. Many of them focus on detecting manipulated PLC logic. State-of-the-art (SOTA) deep learning-based IDSs~\cite{feng2021time, tuli2022tranad, deng2021graph} gather sensor and actuator readings within ICSs to construct profiles for normal or anomalous data. However, these techniques are black-boxes where the reasoning behind the detection results is non-explainable, even for domain experts. In contrast, specification-based IDSs~\cite{stellios2018survey}, are explainable but rely on rules derived from expert knowledge. 
Crafting and maintaining these manual rules is labor-intensive, error-prone, and difficult. Recent advancements (e.g., ~\cite{tan2022cotoru}) include the design of automatic rule generators by parsing PLCs' code (which can thereby be used by an IDS); however, these codes are often proprietary in nature and inaccessible to third-party security vendors. 
Specifically, based on our first-hand experiences with several state-owned ICS operators, their PLC code is classified as ``CONFIDENTIAL" by regulation, so their policy strictly prohibits sharing the code outside their organizations, even to contracted security-vendors; moreover, some core PLC logic is the intellectual-property of solution-vendors and is not accessible even to the ICS operators themselves.
Consequently, without access to the logical information of PLCs, these methods encounter significant challenges in effectively addressing complex attack detection problems in real-world ICS environments.
On the other hand, ICS operators are usually open for contracted security-vendors to instrument their system in a non-invasive way, i.e. allowing security-vendors to passively gather input/output data of PLCs. Additionally, high-fidelity in-house testing facilities (running the same PLC logic as in the field) are usually provided for security-vendors to calibrate their solutions for pre-deployment testing. In this regard, there is a need for accurate attack detection that (1) doesn't rely on code availability from PLC vendors; (2) instead, leverages available system log and network trace; and (3) provides transparent and explainable detection decisions.

\begin{table*}[!t]
    \caption{Summary of comparisons with different methods.}
    \begin{center}
    \label{table_comparison_sota}
    \resizebox{0.8\textwidth}{!}{
    \begin{tabular}{ c|cc|cc}
        \hline
        \multirow{2}{*}{Logic recovery models} & \multicolumn{2}{c|}{Time domain} & \multicolumn{2}{c}{Frequency domain} \\
        & Single-mode logic & Multi-mode logic & Single-mode logic& Multi-mode logic\\
        \hline
        NARMAX, NARX, FIR~\cite{ljung2020deep} & $\checkmark$ & $\times$ & $\checkmark$ & $\times$\\
        Genetic Programming~\cite{schmidt2009distilling, mundhenk2021symbolic} & $\checkmark$ & $\times$ & $\times$ & $\times$\\
        Cluster SR~\cite{ly2012learning} & $\checkmark$ & $\checkmark$ & $\times$ & $\times$\\
        SRLR (ours) & $\checkmark$ & $\checkmark$ & $\checkmark$ & $\checkmark$\\
        \hline
        \hline
        \multirow{2}{*}{Attack detection models} &  Explainable & \multicolumn{2}{|c|}{Minimal prior knowledge} & Easy installation\\
        & with human-understandable rules & \multicolumn{2}{|c|}{without knowledge of attack signatures} & without access to PLC code\\
        \hline
        Specification based methods~\cite{khraisat2019survey, giraldo2018survey} & $\checkmark$ & \multicolumn{2}{|c|}{$\times$} & $\times$\\
        Machine Learning (ML) based methods~\cite{han2022learning, tuli2022tranad, feng2021time, audibert2020usad} & $\times$ & \multicolumn{2}{|c|}{$\checkmark$} & $\checkmark$\\
        CoToRu~\cite{tan2022cotoru} & $\checkmark$ & \multicolumn{2}{|c|}{$\checkmark$} & $\times$\\
        SRLR (ours) & $\checkmark$ & \multicolumn{2}{|c|}{$\checkmark$} & $\checkmark$\\
        \hline
    \end{tabular}
    }
    \end{center}
\end{table*}

In this work, we propose {\it SRLR (Symbolic Regression based Logic Recovery)} --- a method designed to utilize PLCs' inputs (sensor readings) and outputs (actuator commands) to identify PLCs' key control logic and generate a series of rules for attack detection. The design of SRLR is feasible because: For many PLCs, the initiation of their control commands is contingent upon some key decision variables, whose values are determined by mathematical equations of input sensor readings and governed by physical laws and control loop mechanisms. While there are complicated PLC logic (e.g., state-based logic or those with dozens of conditional branches) that do not offer succinct mathematical representation (hence, beyond the scope of our work), we believe that by recovering PLC logic that can be expressed as mathematically deterministic equations between PLC inputs and outputs, one can already cover a large subset of PLC control behaviors in many domains ---- from power systems to other important industrial control systems, as we will validate through our experiments. This observation inspires us to employ Deep Symbolic Regression (DSR)~\cite{petersen2021deep} to infer the formulaic mappings, wherein PLCs' outputs are expressed as mathematical functions of sensor readings. 
SRLR enhances the design of DSR based on the following ICS-unique properties: (1) For some important ICS domains, 
the control logic to be recovered often has specific patterns and expected level of complexity (e.g., including integral and differentiation components), and the simplest representation of such logic is not in the time domain, but in the frequency domain. One of our contributions in designing SRLR is to use such domain knowledge and map the original time-domain data into the complex-valued frequency domain, or known as s-domain, before recovering the control logic. (2) A PLC may be configured to operate in multiple modes, with each mode running different logic. The mode switches in ICS are not frequent. 
This observation inspires us to infer multi-mode logic using a continuity constraint to comprehensively profile the PLCs' control logic. (3) In ICSs, the sensor readings that are fed into PLCs often contain outliers or noise. To deal with that, a robust PLC usually has built-in logic to filter out such outliers. 
We also design an outlier-aware training algorithm for SRLR to prevent outlier data from driving the search in the wrong direction. (4) Lastly, 
we propose an effective regularization method to guide the search towards the appropriate complexity level. 

Thanks to the above enhancements, SRLR is able to recover the controller logic with higher accuracy compared to the DSR method. The derived formulas by SRLR can be leveraged to formulate invariant rules that describe the normal operational profile of PLCs. Deviation from these rules can then indicate in an explainable manner anomalous behavior observed in the PLCs' operation. We summarize the difference between SRLR and the main SOTA methods in Table~\ref{table_comparison_sota}.

Using the control logic recovered by SRLR, we generate a series of rules that accurately describe the normal operational profile of PLCs. Deviation from these rules can be used to detect PLC attacks.
To show the effectiveness of SRLR in detecting controller logic attacks, we extensively evaluate SRLR using a variety of ICSs in different settings. We show that SRLR outperforms all the existing methods by up to $39\%$ in terms of accuracy. For the  Secure Water Treatment (SWaT) dataset \cite{goh2017dataset}, SRLR not only improves the attack detection accuracy on recent SOTA schemes, and it also makes the decision more explainable by mapping to the corresponding invariant that is being violated.

\section{Background and Threat Model}
\subsection{Logic Recovery}
There are industrial processes that operate under a singular mode (or state) with static control logic. We define the recovery of such control logic as a {\it single-mode logic recovery}. Making it more challenging, PLCs in many ICSs encompass multi-mode logic, which carry out different logic based on state-transition conditions. We refer to the problem of profiling the switched control logics, i.e., recovering the control logic for each operating mode/configuration of the considered system as {\it multi-mode logic recovery}. We assume in each mode (be it single or multi-mode system), there is a succinct mathematical equation that governs the relationship between the output of a PLC and its inputs.

One fundamental concept we use to recover logic is symbolic regression (SR). SR has been extensively studied in the field of discovering physical laws~\cite{wang2023scientific} aiming to derive mathematical formulas that accurately represent data generated by physical systems. Given a dataset $(X,y)$, where $X_i \in \mathbb{R}^n, y_i \in \mathbb{R}$, the objective of SR is to identify the function $f: \mathbb{R}^n \rightarrow \mathbb{R}$, such that $y_i=f(X_i)$ for each data point.

\begin{figure}[t]
    \centering
    \includegraphics[scale=0.45]{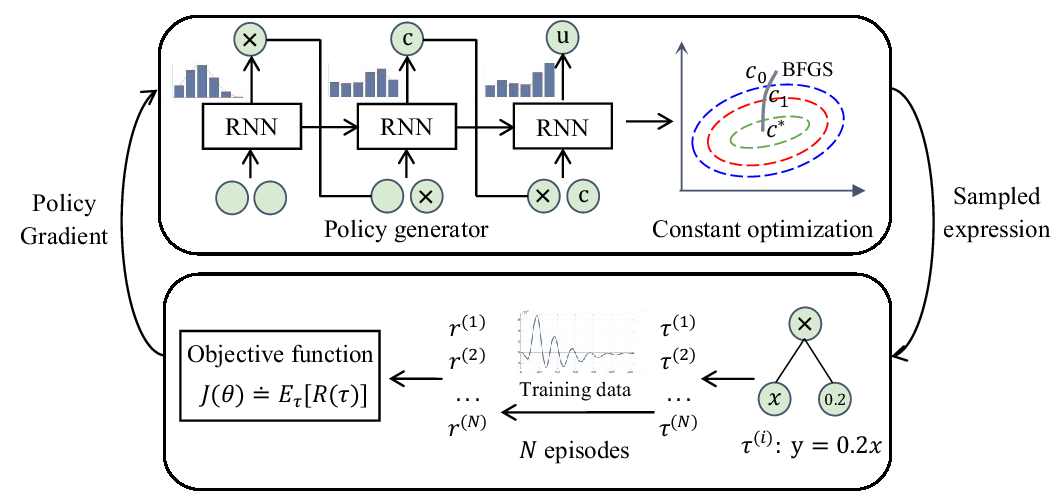}
    \caption{Framework of the DSR model. An RNN generates sample probabilities, which are used to construct an expression tree. The predictive errors of the generated expressions serve as a reward signal to further train the RNN, improving expression generation over time.}
    \label{fig_framwork_dsr}
\end{figure}

In this work, we adopt the recent SOTA model, namely DSR~\cite{petersen2021deep}, as the backbone framework for identifying control logic. As illustrated in Figure~\ref{fig_framwork_dsr}, DSR operates within a reinforcement learning framework, where the Recurrent Neural Network (RNN) serves as the policy. The symbolic representations act as the state, while emitting the probabilities of each symbol within a library (e.g.$\{+,-,$ etc.
$\}$) constitutes the action. During an episode, the sampled symbolic sequence can construct an expression tree, representing a mathematical expression $\tau$ through a pre-order traversal. The risk-policy gradient is adopted in the objective function to promote the generation of optimal mathematical expressions as follows:
\begin{equation}
    \label{eq_objfun_std}
    \begin{aligned}
    J_{risk}(\theta) &\doteq \mathbb{E}_{\tau\sim p(\tau|\theta)}[R(\tau)|R(\tau) \ge R_{\varepsilon}(\tau)]\\
    R(\tau) &= \frac{1}{\frac{1}{\sigma_y}\sqrt{\frac{1}{T}\sum_{t=1}^T(y_t-f(x_t))^2}} 
    \end{aligned}
\end{equation}
where $R(\tau)$ is the reward, $R_{\varepsilon}(\tau)$ is the ($1-\varepsilon$)-quantile of the rewards, and $\sigma_y$ is the standard deviation of the observations. 

Other key components of the DSR design include constant optimization, constraints for avoiding invalid formulas, etc. (c.f. \cite{petersen2021deep} for further details).

\subsection{Threat Model}
PLCs control the physical processes in ICSs. A PLC contains control logic and firmware that drives its operation. Configuration and updates to the PLC code are performed via a workstation PC equipped with PLC management software like TwinCAT 3, using a dedicated interface.

Similar to previous work on PLC security (e.g.,~\cite{tan2022cotoru}), we assume attackers can manipulate the PLC's control logic or modify its firmware to tamper with the PLC's output values. Remote attackers can gain access to a compromised workstation and use that to modify the control logic~\cite{spenneberg2016plc}. Other attack vectors, such as those at the firmware level, can be executed through direct hardware access, such as SD cards or the Joint Test Action Group (JTAG) interface~\cite{tan2022cotoru}, or through supply chain attacks. These advanced attacks can alter the output from PLCs in a more stealthy manner without leaving a visible trace on the PLC code.

We assume that the sensor readings are not modified in this work. Various false data detection and sensor reading protection methods have been extensively studied~\cite{aoudi2018truth}, and SRLR can complement these methods by enhancing control logic protection. Hence, in our threat model, SRLR is used to detect deviation of PLC's behavior, regardless of whether such deviation arises due to changes in application logic or modification in underlying firmware, and regardless of whether deviation happens persistently or occasionally.

\section{SRLR: Symbolic Regression for PLC Logic Recovery \& Attack Detection}

We will present the design of SRLR in this section. After a brief introduction of DSR, we will describe the key techniques we incorporate into SRLR to make DSR work more effectively for PLC logic recovery, and consequently generate explainable attack detection rules.

\subsection{
Design 1: Using the Right Domain}

\subsubsection{Single-mode Logic Recovery}
The DSR method can be readily applied for single-mode logic recovery through the collection of input and output data from dynamic systems. However, oftentimes, specifically in ICSs, the system's time domain exhibits considerable complexity. As such, it becomes challenging for DSR to accurately generate the correct expressions, particularly for functions containing integral or derivative expressions. 

Considering the Linear Time-invariant System (LTI), which can be described as an ordinary differential equation:
\begin{equation}
    \label{eq_lit}
    \sum_{p=1}^P a_p\frac{\mathrm{d}^p}{\mathrm{d}t^p}y_t=\sum_{q=1}^Qb_q\frac{\mathrm{d}^q}{\mathrm{d}t^q}x_t
\end{equation}
where $a_p$ and $b_q$ are constant coefficients, and $P, Q$ represent derivative orders related to inputs and outputs, respectively.

Time-domain systems often involve complex integral and derivative equations in ICSs, posing a challenge for DSR method to generate complex expressions. Nevertheless, these integral and derivative formulations can be readily transformed into simpler expressions in the \textit{frequency domain or the $s$ domain}. Applying  Laplace transform to both sides of Equation (\ref{eq_lit}), results in the following expression:
\begin{equation}
    \label{eq_lit_sdomain}
    Y(s)\sum_{p=1}^P a_p s^p=X(s)\sum_{q=1}^Qb_q s^q
\end{equation}
where $s=\sigma+iw$ is the complex $s$-domain parameter, $\sigma, w$ are real constants. Thus, LTI can be expressed in a much simpler and clear fashion as the transfer function in $s$ domain:
\begin{equation}
    \label{eq_lit_transferfn}
    H(s)=\frac{Y(s)}{X(s)}=\frac{b_0+b_1s+...+b_Qs^Q}{a_0+a_1s+...+a_Ps^P}
\end{equation}

\subsubsection{Model Construction and Training} 
\begin{figure}[t]
    \centering
    \includegraphics[scale=0.4]{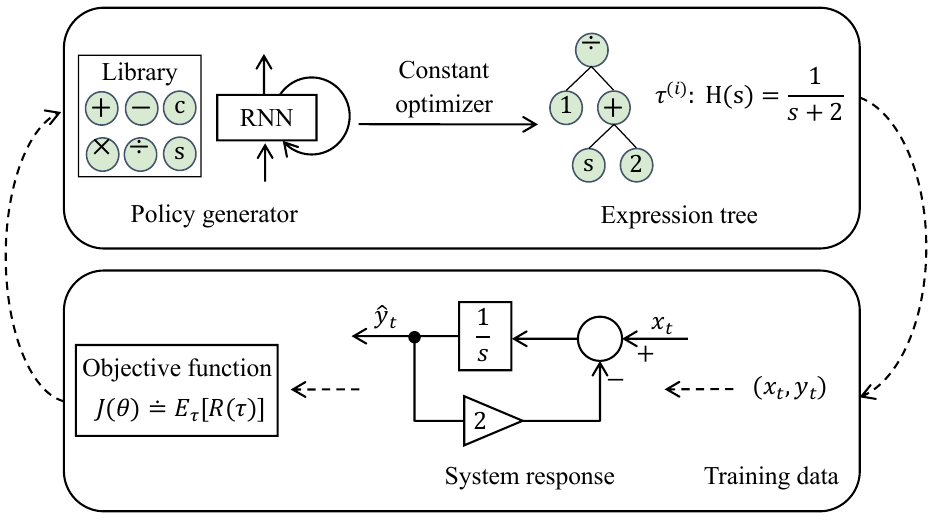}
    \caption{Framework for single-mode process identification in the s-domain using the DSR model. The generated expressions represent transfer functions, and system responses are simulated to calculate the reward.}
    \label{fig_framework_identcontsys}
\end{figure}

We adopt the DSR model as the foundation for directly inferring the system's transfer function from observational data, without relying on expert knowledge. A schematic illustration of the general framework for identifying control systems is depicted in Figure~\ref{fig_framework_identcontsys}. The search library comprises not only basic mathematical symbols, such as $\{+,-,\times, \div, constant\}$, but also contains the complex parameter $s$. Unlike the logic recovery in the time domain, the input variables (e.g. $x_1, x_2$) are not a part of the library. This is because our goal is to predict the actual expression of the transfer function in the $s$ domain.

{In our approach, an RNN acts as the policy to sequentially generate the predictive transfer function. To evaluate system behavior during training, we compute the corresponding time-domain system response. This involves two main steps. First, the transfer function is converted into a state-space representation using the controllable canonical form. Next, a numerical integration technique, such as the Euler method, is applied to produce the system’s output in the time domain. This entire computation process can be efficiently implemented using standard control libraries~\cite{python-control2021}.}

{Subsequently, the prediction errors are utilized to calculate the rewards, we get:}
\begin{equation}
    \label{eq_contsys_rewards}
    \begin{aligned}
    R(\tau) &= \frac{1}{\frac{1}{\sigma_{y}}\sqrt{\frac{1}{T}\sum_{t=1}^{T}(y_t-\hat{f}(x_t))^2}}\\
    \end{aligned}
\end{equation}
{where $y_t, \hat{f}(x_t)$ are the ground truth and prediction outputs respectively.}

SRLR can operate in either the time domain or the s domain. The choice between these domains can be guided by slight domain knowledge about the system's behavior.
In the time domain, it models the system’s dynamics directly, while in the s domain, it generates the corresponding transfer function. Despite these differences, both approaches use the same data representation: time-domain input data are fed into the simulated system to produce time-domain outputs. The prediction error between the simulated and actual outputs is then used to compute a reward signal, which guides model training through a risk-aware policy gradient method.

\subsection{
Design 2: Use of Continuity Constraint}
\subsubsection{Multi-mode Logic Recovery:}In this section, we introduce the details of multi-mode logic recovery.
Consider a controller comprised of $K$ modes. For example, the notation $s_t=k$ indicates that the mode $k$ is operational in the time step $t$. We can describe the dynamics of the $k$-th mode through the equation $y_{t}=f_k(x_t,s_t=k;\zeta_k)$, and the multi-mode control logic can be formulated as follows:
\begin{equation}
    \label{eq_hybridsys_define}
        y_{t} = {f}(x_{t},s_{t};\zeta ) = \left\{ {\begin{array}{*{20}{c}}
    {{f_1}(x_t,s_t=1;\zeta_1)}, (x_t,y_t)\in \mathcal{D}_1\\
    {{f_2}(x_t,s_t=2;\zeta_2)}, (x_t,y_t)\in \mathcal{D}_2\\
    {...}\\
    {{f_K}(x_t,s_t=K;\zeta_K)}, (x_t,y_t)\in \mathcal{D}_K
    \end{array}} \right.
\end{equation}
where the function $f(\zeta)$ represents the equation governing the mode and $\zeta$ is the parameters, and the dynamics $f(\zeta)$ can be linear or nonlinear. $\mathcal{D}_k=\{(x_t,y_t)|s_t=k\}$ denotes the input/output set of the $k$-th mode.

The mode equation $f_k(x_t,s_t=k;\zeta_k)$ and the mode dataset $\mathcal{D}_k$ are not accessible apriori. To correctly determine the logic of such a multi-mode system, we need to identify the equation $f(\zeta_k)$ of each mode, the sub-dataset $\mathcal{D}_k$ of each mode, and the total number $K$ of modes.

The multi-mode process identification problem can be formulated as two optimization tasks. The first task involves (correctly) partitioning the dataset into sub-datasets corresponding to each operating mode, while the second employs DSR on that sub-dataset to identify the accurate equation governing that mode. However, solving the joint optimization problem presents a non-trivial challenge. We introduce a binary vector $\gamma_{k} \in \mathbb{R}^T$ as the membership value, where the membership $\gamma_{k,t}$ indicates whether the input-output pair $(x_t,y_t)$ belongs to $k$-th mode. If true, $\gamma_{k,t}=1$; otherwise, $\gamma_{k,t}=0$. The joint objective function is formulated as follows:
\begin{equation}
    \label{eq_objfun_hybridsys}
    \begin{aligned}
    &\min_{\gamma,\zeta} \quad \sum_{t=1}^T\sum_{k=1}^K{\gamma_{k,t}l(y_t,\hat{y}_t;\zeta_k)+ \phi L(\hat{y}_t) + g(\gamma_{k,t};\lambda)}\\
    &\begin{array}{r@{\quad}r@{}l@{\quad}l}
        s.t.
        &(x_t,y_t) &\in \mathcal{D}\\
        &\gamma_k &\in \{0,1\}^{|\mathcal{D}|}\\
         &\hat{y}_t &= f_k(x_t,s=k;\zeta_k)
    \end{array}
    \end{aligned}
\end{equation}
where $\hat{y}_t$ represents the predicted output at time step $t$. $\mathcal{D} = \bigcup_{k=1}^K \mathcal{D}_k$ and $|\mathcal{D}|$ is the total number of input-output pairs. The first term $l(y_t,\hat{y}_t;\zeta_k)$ is the prediction error for the equation of $k$-th mode (e.g. square error). The second term $ L(\hat{y}_t)$ is the regularization for the prediction model with hyperparameter $\phi$. The third term $g(\gamma_{k,t};\lambda)$ is the regularization assigning the input-output pairs to a specific mode in training.

This is a general framework that can be utilized for multi-mode process identification tasks by specifying a regressor to predict system's output and defining a regularizer for partitioning the sub-dataset. We employ DSR as the regressor, and choose a hard regularizer that is widely adopted in previous studies~\cite{kumar2010self}. The regularizer is formulated as $g(\gamma;\lambda)=-\lambda \gamma$.

We employ the Alternative Optimization Strategy (AOS)~\cite{kumar2010self} to address the mixed-integer programming problem. AOS is an alternatively iterative method where we first fix the $\gamma$ value and update the regressor model. Then, we fix the regressor parameters and update the $\gamma$ value. The objective function for optimizing the regularizer is as follows:
\begin{equation}
    \label{eq_objfun_regularizer}
    \begin{aligned}
    &\min_{\gamma} \quad \sum_{t=1}^T\sum_{k=1}^K{\gamma_{k,t}l_{k,t} - \lambda \gamma_{k,t}}\\
    &\begin{array}{r@{\quad}r@{}l@{\quad}l}
        s.t.
        &\gamma_k &\in \{0,1\}^{|\mathcal{D}|}\\
    \end{array}
    \end{aligned}
\end{equation}
where $l_{k,t}$ is the abbreviation of $l(y_t,\hat{y}_t)$ for convenience.
When the regressor parameters are fixed, the loss of prediction is a scalar. Thus, the optimal $\gamma_{k,t}$ can be obtained as follows:
\begin{equation}
    \label{eq_solution_regularizer}
    \gamma_{k,t}^*(\lambda,l_{k,t})=\left\{
    \begin{aligned}
        & 0 & if\ \ l_{k,t}\le \lambda\\
        & 1 & if\ \ l_{k,t}> \lambda
    \end{aligned}
    \right.
\end{equation}

\begin{figure}[t]
    \centering
    \includegraphics[scale=0.35]{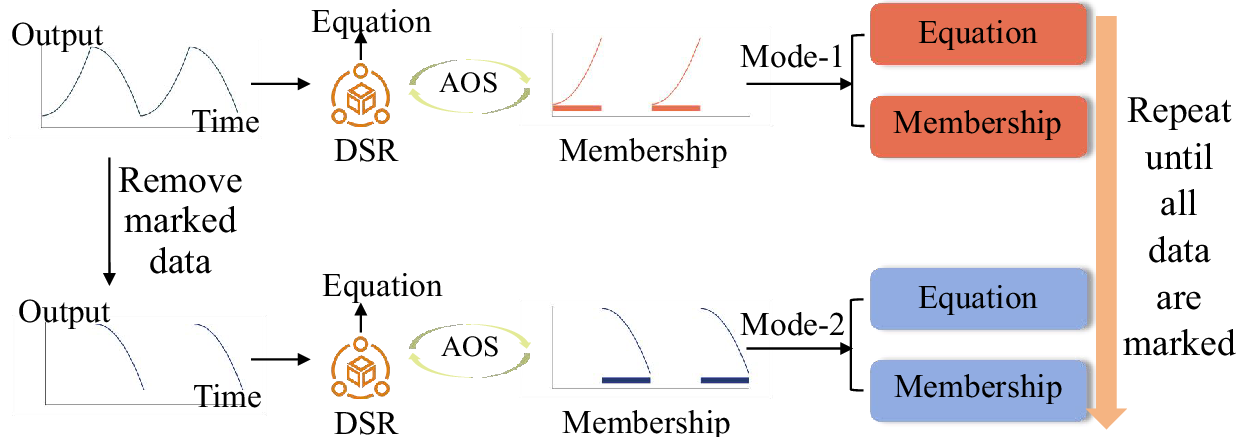}
    \caption{Example of multi-mode logic recovery. The DSR model is iteratively optimized alongside a binary variable. Identified mode points are progressively removed, and the process continues until all points are assigned to a mode.}
    \label{fig_framework_identmultimode}
\end{figure} 

\subsubsection{Refined Optimization Approach}
If the prediction loss $l_{k,t}$ is less than the pre-defined threshold $\lambda$, then the input-output pair $(x_t,y_t)$ is assigned to mode $k$ ($\gamma_{k,t}=1$). Conversely, if the loss surpasses the threshold, it indicates that $(x_t,y_t)$ is not generated by $k$-th mode ($\gamma_{k,t}=0$). 
Given that fewer points in a multi-mode process provide limited information about the subsystem, we make the following basic hypothesis for constructing the identification method. This hypothesis holds for most ICS as mode switches are often non-transient in nature.

\newtheorem{assumption}{Assumption}
\begin{assumption} \label{assm:w}
At least $w$ input-output pairs are generated during each mode activation 
(i.e., a mode switch can only occur after $w$ time slots), where $w$ is some small constant. 
\end{assumption}

Considering that the data generation process in dynamic systems follows a temporal sequence, we leverage this smooth information and incorporate a regularization term to prompt the algorithm to commence the identification process with a greater relaxation for the first $w$ data points, i.e., a higher value of $\lambda$. When we are identifying the $k$-th mode, the objective function can be expressed as follows:
\begin{equation}
    \label{eq_objfun_regularizer2}
    \begin{aligned}
    &\min_{\gamma_k} \quad \sum_{t=1}^T{\gamma_{k,t}l_{k,t} - \sum_{t=1}^{w}\lambda_1 \gamma_{k,t} - \sum_{t=w+1}^{T}\lambda_2 \gamma_{k,t}}\\
    &\begin{array}{r@{\quad}r@{}l@{\quad}l}
        s.t.
        &\gamma_k &\in \{0,1\}^{|\mathcal{D}|}\\
    \end{array}
    \end{aligned}
\end{equation}
where $\lambda_1$ is a larger threshold that relaxes the condition for the initial points to be divided to mode $k$, and $\lambda_2$ is a smaller value that limits the predictions with minor errors.
In our experiments, the parameters $\lambda_1$ and $\lambda_2$ are empirically set to 1 and 0.001, respectively.
When the $\gamma_k$ is fixed, we proceed to train the DSR model using the selected training data specified by $\gamma_k$. The objective function is defined as follows:
\begin{equation}
    \label{eq_objfun_dsr}
    \begin{aligned}
    &\min_{\zeta_k} \quad \sum_{t=1}^T{\gamma_{k,t}l(y_t,\hat{y}_t;\zeta_k)+ \phi L(\hat{y}_t)}\\
    &\begin{array}{r@{\quad}r@{}l@{\quad}l}
        s.t.
        &(x_t,y_t) &\in \mathcal{D}\\
         &\hat{y}_t &= f_k(x_t,s=k;\zeta_k)
    \end{array}
    \end{aligned}
\end{equation}

We can employ risk policy gradient to optimize the DSR problem, and the gradient is computed as follows when identifying mode $k$:
\begin{equation}
    \label{eq_solution_dsr}
    \nabla_{\theta}J_{risk}(\zeta_k;\varepsilon) \approx \frac{1}{\varepsilon T}\sum_{t=1}^{T} \mathbb{I}\cdot [R(\tau^{(t)})-\widetilde{R}_{\varepsilon}(\zeta_k)]\nabla_{\zeta_k}\log p(\tau^{(t)}|\zeta_k)
\end{equation}
where $\widetilde{R}_{\varepsilon}(\zeta_k)$ is the $(1-\varepsilon)$ quantile of the rewards in a training batch. $\mathbb{I}$ is an indicator variable, where if ${R(\tau^{(t)})\ge \widetilde{R}_{\varepsilon}(\zeta_k)}$, then $\mathbb{I}=1$; otherwise, $\mathbb{I}=0$.

As shown in Figure~\ref{fig_framework_identmultimode}, the proposed method iteratively identifies each mode and removes the index once the sub-dataset is successfully partitioned.

\subsection{Design 3: Outlier-aware Training}
{While DSR is optimized in the previous section to handle the right domain and multiple modes for controller logic recovery, it may still be susceptible to outliers in the training data. These outliers can arise from environmental interference, noise during data collection, or internal system vibrations. Such sensitivity can result in DSR generating overly complex expressions, leading to overfitting or even failure to recover control logic. To mitigate this issue, we proposed a data dropout approach to address potential outliers.}

Rewards are calculated upon the generation of an expression by DSR, and these rewards are then utilized in the risk-seeking policy gradient mechanism to update the parameters of DSR. Throughout this process, all training data, including outliers, contribute to the reward calculation. Typically, the evolutionary models attempt to initially fit easier data in the early iterations and then tackle more challenging data (such as outliers) in subsequent iterations. Consequently, the normal data are likely to be fitted by generated candidate expressions, resulting in higher rewards for these data points and lower rewards for outliers. Drawing from this observation, we implement a data dropout approach where we remove the data representing the lowest $\alpha$ percent in terms of reward for a chosen constant $\alpha$, as these are deemed more likely to be outliers. 

\subsection{Design 4: Complexity Regularization}
In general, high-order systems featuring a more complex transfer function can emulate the response of lower-order systems when provided with input data. Consequently, it is essential to impose additional constraints on DSR to ensure the generation of expressions with suitable complexity levels.

We quantify complexity based on the number of mathematical symbols~\cite{petersen2021deep}. Specifically, the symbols $[\div, \times]$ exhibit higher complexity compared to $[+,-]$ as detailed in Table~\ref{table_complexity}, and the complexities are summed as the final complexity of the expression. We utilize the Akaike Information Criterion (AIC)~\cite{akaike1974new} as a regularization term to encourage our model to generate less complex expressions. AIC is an indicator of predictive error derived from information theory, serving as a criterion for model selection in regression and balancing precision with simplicity. Its formulation is as follows:
\begin{equation}
    \label{eq_aic}
    \begin{aligned}
    AIC = (2m-ln(\hat{L}))/n
    \end{aligned}
\end{equation}
where $m$ is the number of model parameters, $\hat{L}$ is the predictive likelihood and $n$ is the total number of data.

The preferred candidate model typically exhibits a lower AIC score, indicating either a lower complexity or a higher likelihood. Assuming the error residuals follow an independent identical normal distribution, the maximum value of a model's log-likelihood can be formulated by the residual. As our calculation of the complexity of generated expressions is based on the number of symbols, the complexity term tends to be small. To enhance the impact of expression complexity, we introduce an additional complexity term into the AIC formula. The final modified AIC is calculated as follows:
\begin{equation}
    \label{eq_modifiedaic}
    \begin{aligned}
    ln(\hat{L}) &= -\frac{n}{2}\cdot ln(2\pi) - \frac{n}{2}\cdot ln\left(\frac{sse}{n}\right) - \frac{n}{2} \\
    AIC &= (2m-ln(\hat{L}))/n + m
    \end{aligned}
\end{equation}
where $sse$ is the residual error.

The negative AIC is a regularization term within the reward function that encourages the model to generate expressions that strike a balance between complexity and precision.

\begin{table}[t]
    \caption{The measurement of symbol complexity.}
    \begin{center}
    \label{table_complexity}
    \resizebox{0.85\linewidth}{!}{
    \begin{tabular}{ c|c|c|c|c }
        \hline
        Complexity & 1 & 2 & 3 & 4\\
        \hline
        \multirow{2}*{Symbol} & $+,-,\times,$ & \multirow{2}*{$\div$} & \multirow{2}*{$\sin, \cos$} & \multirow{2}*{$\log, \exp$}\\
        & constant,variable & & & \\
        \hline
    \end{tabular}}
    \end{center}
\end{table}

\begin{figure}[t]
    \centering
    \includegraphics[scale=0.45]{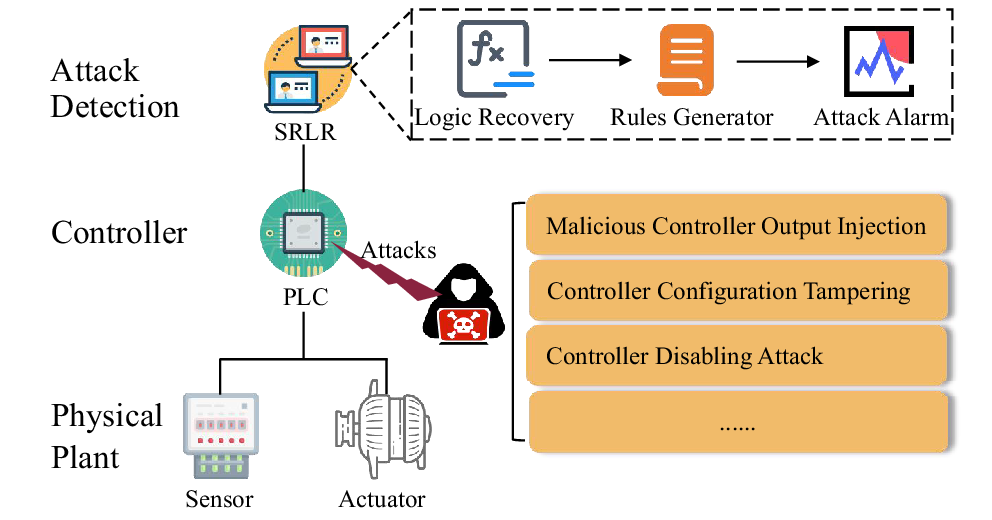}
    \caption{Attack detection framework based on SRLR. The model learns normal PLC logic from sensor and actuator data, then generates rules to monitor and ensure PLC security.}
    \label{fig_framwork_attack}
\end{figure}

\subsection{Using SRLR for Attack Detection}
We now present how our proposed SRLR can be applied to detect different attacks in ICSs. The overall framework is depicted in Figure~\ref{fig_framwork_attack}, where the ICSs comprise both the physical process and the controller module. Sensors measure the states of the physical system, such as water level, flow rate, etc., 
and these measurements are fed into PLCs, which analyze the system states and emit control commands to actuators. The actuators then execute these commands to control the physical system. However, PLCs are susceptible to vulnerabilities and may be targeted by attackers, leading to the transmission of incorrect or even harmful commands.

Notice that, once we have correctly identified the normal system process (in either single-mode or multi-mode), we can leverage that to check the behavior of  
the system. If there is a deviation from the expected behavior, the system can be considered as behaving anomalously and ``under-attack".  
These predictions are compared with the actual output from the actuators, with anomaly score calculated as:
\begin{equation}
    \label{eq_anomaly_score}
    \begin{aligned}
    e_t = \frac{1}{T}\sum_{t=1}^T{\lvert y_t-\hat{y}_t \rvert}
    \end{aligned}
\end{equation}
where $y_t$ represents the observed PLCs output and $\hat{y}_t$ denotes the prediction of the identified system.

The output commands may exhibit some fluctuations, especially at the state-switching time points (during change in the operating mode), resulting in spikes in the anomaly scores. To prevent these points from being erroneously classified as an attack, we employ the Exponentially Weighted Moving Average (EWMA) smoothing algorithm to smooth the anomaly scores. Points are only flagged as attacks when the smoothed anomaly score exceeds a predefined threshold determined through a non-parametric threshold~\cite{hundman2018detecting}.

{The SRLR model can be integrated into monitoring systems within ICSs to deliver accurate numerical results that support timely and efficient alarms. In addition to its predictive capabilities, SRLR generates human-understandable rules, enabling operators to trace potential threats by analyzing these rules. Furthermore, it identifies compromised variables based on the generated rules, which is the critical information that helps security teams investigate and determine the root cause of incidents.}

\section{Experiments on Logic Recovery}
\begin{table*}[t]
    \caption{The recovery results of SRLR for the EWMA equation. Ground truth: $y(t)=0.8x(t)+0.16x(t-1)+0.032x(t-2)$.}
    \begin{center}
    \label{table_ewma}
    \resizebox{\textwidth}{!}{
    \begin{tabular}{ c|c|c|c|c|c }
        \hline
        Dataset & AIC & Outlier-aware & Identified mode & Complexity & BFR\\
        \hline
        \multirow{4}*{Raw data} & $\times$ & $\times$ & $y(t)=0.2307x(t-1)+x(t)/(0.2585x(t-1)/(x(t-1)+x(t-2))+1.1842)$ & 19 & 0.9966\\
        & $\times$ & $\checkmark$ & $y(t)=0.9064x(t-2)x(t)/(0.873x(t-2)-0.1269x(t-1)+0.1673x(t)+0.1178)$ & 22 & 0.9963\\
        & $\checkmark$ & $\times$ & $y(t)=0.7958x(t)+0.1653x(t-1)+0.0308x(t-2)$ & 15 & 0.9959\\
        & $\checkmark$ & $\checkmark$ & $y(t)=0.7917x(t)+0.1706x(t-1)+0.0296x(t-2)$ & 17 & 0.9957\\
        \hline
        \multirow{4}*{Contanimated data} & $\times$ & $\times$ & $y(t)=0.0029x(t-1)(x(t-2)(-x(t-2)+x(t)+641.6587)-303.4484x(t-2)+x(t)-8.2933)/x(t-2)$ & 29 & 0.8293\\
        & $\times$ & $\checkmark$ & $y(t)=0.7999x(t)+0.2x(t-1)(0.0005x(t-2)-0.0005x(t-1)+0.96)-0.0007$ & 24 & 0.9959\\
        & $\checkmark$ & $\times$ & $y(t)=0.9919x(t)-0.2275x(t-1)+0.2275x(t-2)$ & 10 & 0.9897\\
        & $\checkmark$ & $\checkmark$ & $y(t)=0.7985x(t)+0.1628x(t-1)+0.0305x(t-2)$ & 15 & 0.9945\\
        \hline
    \end{tabular}
    }
    \end{center}
\end{table*}
\begin{table*}[htbp]
    \begin{center}
    \caption{The identification results of SRLR and baselines in multi-mode logic recovery. (noise level $N_p=0.02$)}
    \label{table_compare_multimode}
    \resizebox{\textwidth}{!}{
    \begin{tabular}{ c|c|c|c|c|c }
        \hline
        Dataset & Mode & The ground truth mode & SRLR & Cluster SR & Cluster DSR\\
        \hline
        \multirow{2}*{Hysteresis Relay} & 1 & $y=1$ & $y=1$ & $y=1$ & $y=1$\\
        & 2 & $y=-1$ & $y=-1$ & $y=-1$ & $y=-1$\\
        \hline
        \multirow{2}*{Continuous Hysteresis} & 1 & $y=0.5x^2+x-0.5$ & $y=0.4988x^2+0.9993x-0.4994$ & $y=0.013+x$ & $y=0.4988x^2+0.9993x-0.4994$\\
        & 2 & $y=-0.5x^2+x+0.5$ & $y=-0.4987x^2+0.9998x+0.4996$ & $y=x$ & $y=-0.4987x^2+0.9997x+0.4995$\\
        \hline
        \multirow{3}*{Phototaxic Robot} & 1 & $y=x_2-x_1$ & $y=x_2-x_1$ & $y=x_2-x_1$ & $y=x_2-1.0009x_1$\\
        & 2 & $y=1/(x_1-x_2)$ & $y=0.9647/(1.0332x_1-1.0332x_2)$ & $y=0.969/(x_1-x_2)$ & $y=0.083x_1-0.083x_2-0.0873$\\
        & 3 & $y=0$ & $y=0$ & $y=-0.005$ & $y=0$\\
        \hline
        \multirow{3}*{Non-linear System} & 1 & $y=x_1x_2$ & $y=x_1x_2$ & $y=x_1x_2$ & $y=1.0827x_1x_2-0.4131x_1$\\
        & 2 & $y=6x_1/(6+x_2)$ & $y=5.9206x_1/(5.9153+x_2)$ & $y=x_1x_2-(x_1-5.169)x_1/(-6.699)$ & $y=0.4774x_1^2+0.4774x_1x_2$\\
        & 3 & $y=(x_1+x_2)/(x_1-x_2)$ & $y=(x_1+x_2)/(x_1-x_2)$ & $y=(x_1+x_2)/(x_1-x_2)$ & $y=1.2327x_1x_2-1.2797x_1$\\
        \hline
    \end{tabular}
    }
    \end{center}
\end{table*}
In this section, we conduct experiments to validate the performance of SRLR in terms of correctly identifying the operating mode of an ICS system.


\subsection{{Metrics}}
{We evaluate the correctness of the recovered logic from two key perspectives: fidelity and explainability.}

{Fidelity refers to how accurately the recovered logic reproduces the true behavior of the system. To quantify this, we use the Best Fit Ratio (BFR) as defined in~\cite{masti2021learning}. The BFR is calculated as follows:}
\begin{equation}
    \label{eq_bfr}
    \begin{aligned}
    \text{BFR} = \max \{0,1-\frac{\|y-\hat{y}\|_2}{\|y-\bar{y}\|_2}\}
    \end{aligned}
\end{equation}
{where $y=[y(1),...,y(T)], \hat{y}=[\hat{y}(1),...,\hat{y}(T)]$ are the ground truth and predicted system outputs, respectively. The BFR ranges from 0 to 100\%, with higher values indicating closer alignment with the true system dynamics. This metric also serves as an indicator of model precision: if the recovered logic consistently achieves high BFR scores across test data, it suggests strong generalization and reliable predictive accuracy.}

{Explainability evaluates how easily a human can interpret the recovered expressions and understand the relationships among variables. First, we assess the overall symbolic complexity of each mode identified by SRLR, as shown in Table~\ref{table_complexity}. This complexity measure considers both the number and diversity of symbols used. Lower complexity generally implies greater explainability. Second, we examine symbolic equivalence, i.e. whether the recovered expression matches the known system dynamics. When the true dynamics are available, we can use intuitive visual inspection or formal tools such as SymPy\footnote{\url{docs.sympy.org}}} to determine equivalence.

{This focus on explainability is a key advantage of SRLR. Unlike many machine learning-based approaches that yield only black-box numerical predictions, SRLR produces explicit symbolic expressions that not only approximate system behavior but also provide insight into the underlying control logic and variable relationships.}

\subsection{Single-mode Logic Recovery}
For the first evaluation, we tasked SRLR to recover the EWMA equations in the time domain. EWMA is a popular and prevalent method employed on PLCs to preprocess data for smoothing in ICSs~\cite{newhart2019data}. EWMA smooths time series data through an exponential window, expressed as follows:
\begin{equation}
    \label{eq_ewma}
    \begin{aligned}
    y(t) &= \alpha x(t) + (1-\alpha)y(t-1) \\
     &\approx \alpha x(t)+\alpha(1-\alpha)x(t-1)+\alpha(1-\alpha)^2x(t-2)
    \end{aligned}
\end{equation}
where $\alpha$ is the smoothing factor, $x(t)$ is the input data at $t$-th time step and $y(t)$ is the output after EWMA process. Notice from the definition of EWMA -- earlier inputs have progressively less influence on the output than recent ones. For our experiments, we truncate the input with a finite window size 2 and set the $\alpha$ to 0.8 to keep the resulting equation simple. However, our method also works when considering a larger number of inputs.
Additionally, to simulate a contamination situation, we randomly introduce outliers to the raw data points as: $\tilde{x}(t) = \bar{x}\cdot \epsilon + x(t)$,
where $\bar{x}$ is the local mean, and $\epsilon\sim \mathcal{N}(-0.1,0.1)$.

The dataset utilized in this study is sourced from the real-world SWaT testbed~\cite{goh2017dataset}. 
The identification results are presented in Table~\ref{table_ewma}, revealing that the model can generate expressions with high BFR when outliers are not injected. However, without AIC regularization, the model tends to generate complex expressions that deviate significantly from the ground truth. The AIC regularization helps constrain the model to generate the correct expression with a lower complexity.
In contaminated data, i.e., data with injected outliers AIC regularization alone results in incorrect expressions. However, when equipped with both the AIC regularization and outlier-aware method, the model can generate correct expressions with lower complexity and a high BFR. Experimental results confirm the effectiveness of our proposed AIC regularization and outlier-aware method, especially in cases where the training data contains outliers.

\subsection{Multi-mode Logic Recovery}
\begin{figure*}[!t]
    \centering
    \subfloat[Continuous Hysteresis]{
    \begin{minipage}[b]{0.49\textwidth}
    \includegraphics[width=1\linewidth]{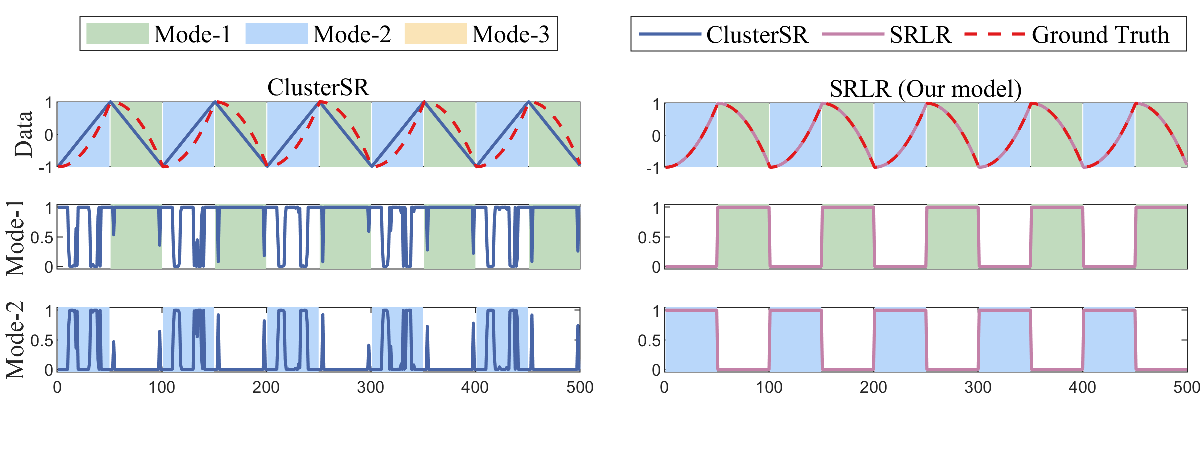}
    \end{minipage}}
    \subfloat[Nonlinear System]{
    \begin{minipage}[b]{0.49\textwidth}
    \includegraphics[width=1\linewidth]{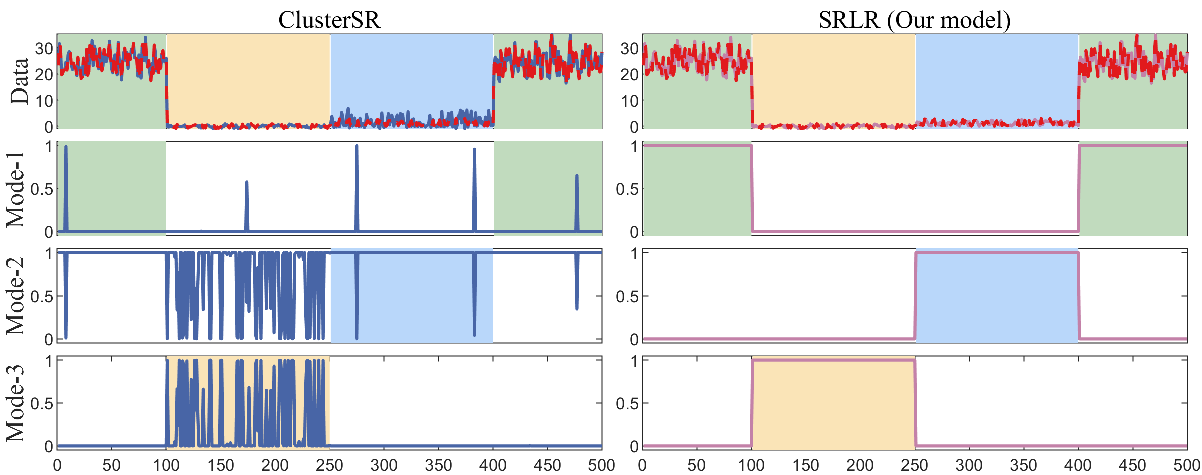}
    \end{minipage}}
    \caption{Comparison of mode identification results between Cluster SR and the proposed SRLR method. The first row shows simulated data from both methods alongside ground truth. Subsequent rows present the identified mode indices for each method.}
    \label{fig_compare_index}
\end{figure*}
Building upon the findings from single-mode logic recovery, we evaluate the performance of SRLR on identifying hybrid ICS systems having multiple operating modes.

\noindent\textbf{Dataset. } We consider four classic hybrid dynamic systems as in \cite{ly2012learning}. These hybrid dynamic systems include hysteresis relay, continuous hysteresis loop, phototaxic robot, and nonlinear system. The hysteresis relay and continuous hysteresis loop represent discrete and continuous switch systems, respectively. The phototaxic robot system comprises three modes to track the movement of a light-interacting robot. The nonlinear system comprises three modes with nonlinear equations. For further datasets details, please refer to~\cite{ly2012learning}.

We also evaluate the efficacy of SRLR across varying levels of noise in the data, comparing its performance against several baseline models. To this end, we consider an addition of zero-mean Gaussian noise to the input data, with noise levels ranging from 0\% to 10\%~\cite{ly2012learning}. The noise level is proportional to the standard deviation of the raw data, denoted as $N_p=\sigma_{noise}/\sigma_y$, where $\sigma_{noise}$ represents the standard deviation of the noisy data and $\sigma_y$ is the standard deviation of the noiseless data. We use the noisy data as the training set and the noiseless data as the testing set following the evaluation in~\cite{ly2012learning} to ensure a fair comparison.
\begin{figure}[t]
    \centering
    \includegraphics[scale=0.45]{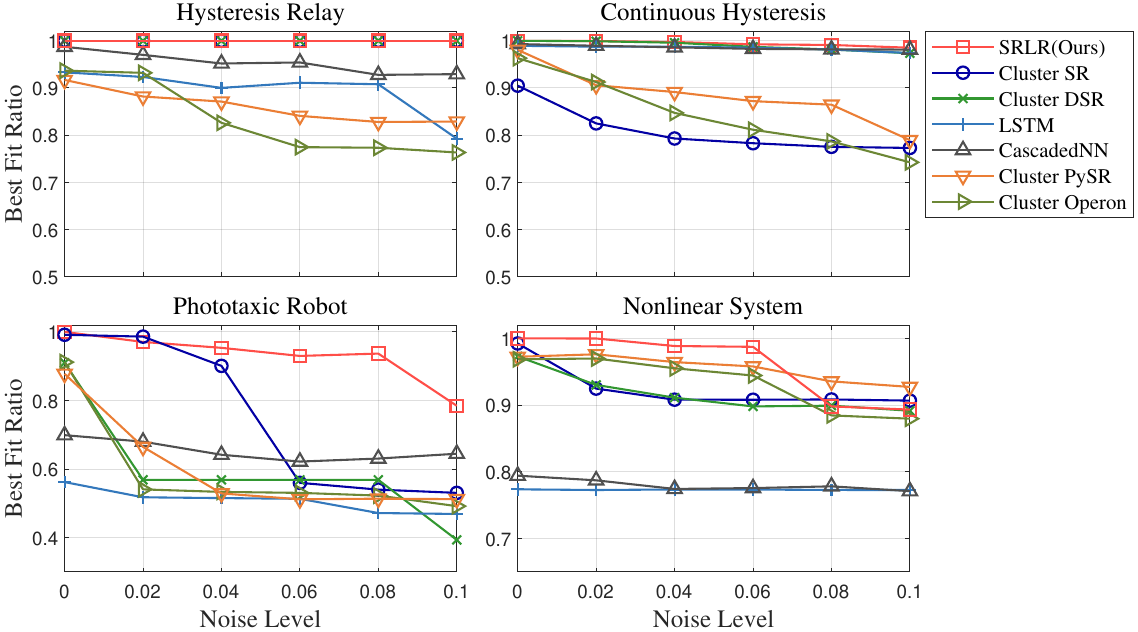}
    \caption{The comparative BFR across different noise levels.}
    \label{fig_compare_hybrid}
\end{figure}

\noindent\textbf{Baselines. } 
We evaluate the performance of SRLR against six baseline methods. The first, Cluster SR~\cite{ly2012learning}, is a symbolic regression technique that combines Generic Programming (GP) with the expectation-maximization algorithm for clustering and membership estimation. To benchmark SRLR against SOTA symbolic regression tools, we also create three variants: Cluster DSR, Cluster PySR, and Cluster Operon. These versions replace the GP component in Cluster SR with more advanced symbolic regression algorithms, such as DSR, PySR~\cite{cranmer2023interpretable}, and Operon~\cite{burlacu2020operon}, respectively. For broader comparison, we also include two black-box deep learning models: Long Short-term Memory Network (LSTM) and a Deep Cascaded Neural Network (CascadedNN)~\cite{ljung2020deep}.

\noindent\textbf{Comparison results. } 
As shown in Figure~\ref{fig_compare_hybrid}, SRLR consistently outperforms all baseline methods in terms of average BFR. The black-box models, LSTM and CascadedNN, demonstrate competitive results on the discrete and continuous hysteresis datasets, where their performance closely matches SRLR. However, their effectiveness declines on more complex tasks, such as modeling nonlinear systems. In contrast, SRLR not only maintains high predictive accuracy but also provides interpretable equations that reveal the underlying system dynamics.

The performance of the cluster-based symbolic regression models varies across datasets. For example, Cluster SR achieves better results on the phototaxic robot dataset but struggles with the continuous hysteresis system, leading to a significantly lower BFR, as shown in Table~\ref{table_compare_multimode}. This inconsistency can be attributed to the shared membership estimation method across all cluster-based models, which fails to account for the temporal continuity present in system-generated data. As a result, these models underperform compared to SRLR, particularly on tasks that require capturing complex or sequential patterns.

Furthermore, we conducted a comparative analysis of the mode indexes identified by SRLR (ours) and Cluster SR. Both approaches are trained using noisy data, and the resulting index identification results are illustrated in Figure~\ref{fig_compare_index}. 
A key difference we observe is that Cluster SR does not account for the temporal dynamics of systems and therefore overlooks the local continuity of the point distribution within each mode.

As shown in Figure~\ref{fig_compare_index}, Cluster SR yields a sparse membership distribution in the continuous hysteresis system, despite most values being correct. In the case of the nonlinear system, Cluster SR even misclassifies the membership due to the presence of noisy training data. In contrast, as SRLR leverages temporal information and employs a larger $\lambda_1$ (see Equation~\ref{eq_objfun_regularizer2}), along with an effective fine-tuning mechanism, it helps the correct identification of the initial points of each mode, providing the algorithm with essential prior information to guide its decisions. This leads to the correct identification of initial segments, and inference of the correct mathematical structure which results in overall better performance.

\subsection{Logic Recovery in Distribution Grid}
\label{app:distribution_grid}
Next we evaluate SRLR on a distribution grid with hundreds of voltage regulators are deployed and updated independently to stabilize the local voltages delivered to users. To simulate this distributed network, we randomly generated the gains and time constants for each subsystem within the normal range, as referenced in the literature~\cite{saadat1999power}. The corresponding parameters of Proportional-Integral-Derivative (PID) controllers were automatically tuned using MATLAB PID Tuner App.

As illustrated in Table~\ref{table_distribution_grid}, we simulate 40 distributed voltage regulator systems, each consisting of four physical subsystems and one PID controller. {The recovery performance of SRLR in physical subsystems (amplifier, exciter, generator, and sensor) achieved an average BFR 0.9999, demonstrating SRLR's stability under diverse conditions. Additionally, the average BFR for the controllers was 0.9969, indicating SRLR's effectiveness in identifying complex logic (e.g. second-order mathematical expression in PID controllers) with a high degree of accuracy. In total, 395 subsystems were identified across 79 distributed systems, and the high success rate highlights SRLR’s robustness and suitability for deployment in large-scale, distributed real-world systems.}

\begin{table}[t]
    \caption{The parameters and results of the distribution grid.}
    \begin{center}
    \label{table_distribution_grid}
    \resizebox{\linewidth}{!}{
    \begin{tabular}{ c|c|c|c|c }
        \hline
        System & Gain & Time Constant & Total Number & Average BFR\\
        \hline
        Amplifier & (10, 400) & (0.02, 0.1) & 79 & 0.9999\\
        \hline
        Exciter & (0.7, 1) & (0.02, 1) & 79 & 0.9999\\
        \hline
        Generator & (0.7, 1) & (1, 2) & 79 & 0.9999\\
        \hline
        Sensor & (0.7, 1) & (0.01, 0.06) & 79 & 0.9999\\
        \hline
        Controller & \multicolumn{2}{|c|}{Automatically tuned} & 79 & 0.9969\\
        \hline
    \end{tabular}
    }
    \end{center}
\end{table}

\section{Experiments on Attack Detection}
In this section, we evaluate the performance of SRLR to detect controller logic attacks in different ICS environments, including a power grid frequency control system~\cite{obaid2016fuzzy} and the SWaT testbed~\cite{goh2017dataset}.

\subsection{{Metrics and Baselines}}
\noindent\textbf{{Metrics.}}
{We evaluate detection performance using widely-used metrics: F1 score, precision, and recall. Following established practice in prior work, we enumerate all possible threshold values and report the highest achieved F1 score, denoted as F1, Precision, and Recall. Additionally, we adopt the adjust-point evaluation strategy~\cite{su2019robust, audibert2020usad, feng2021time}, which considers an entire attack segment correctly detected if at least one alarm is triggered within its duration. Metrics calculated using this approach are reported as F1$^*$, Presion$^*$, and Recall$^*$. The average performance metrics along with their standard deviations are reported.}

\noindent\textbf{{Baselines.}}
{We compare SRLR against several neural System Identification (SI)} methods, including RNN~\cite{ljung2020deep}, CascadedNN~\cite{ljung2020deep}, and NSIBF~\cite{feng2021time} along with its variants NSIBF-RECON and NSIBF-PRED. RNN and CascadedNN directly predict system outputs using neural networks. NSIBF constructs a state-space model via neural networks and updates system states through Bayesian filtering. It detects anomalies based on a combination of reconstruction and prediction errors. In addition to these neural SI methods, we also evaluate SRLR against deep learning-based anomaly detection models. These include the encoder-decoder architecture EncDec-AD~\cite{malhotra2016lstm}, the density-based approach DAGMM~\cite{zong2018deep}, the adversarially trained method USAD~\cite{audibert2020usad}, and the probabilistic modeling framework OmniAnomaly~\cite{su2019robust}.

\subsection{Power Grid Frequency Control System}
\begin{figure}[t]
    \centering
    \includegraphics[scale=0.3]{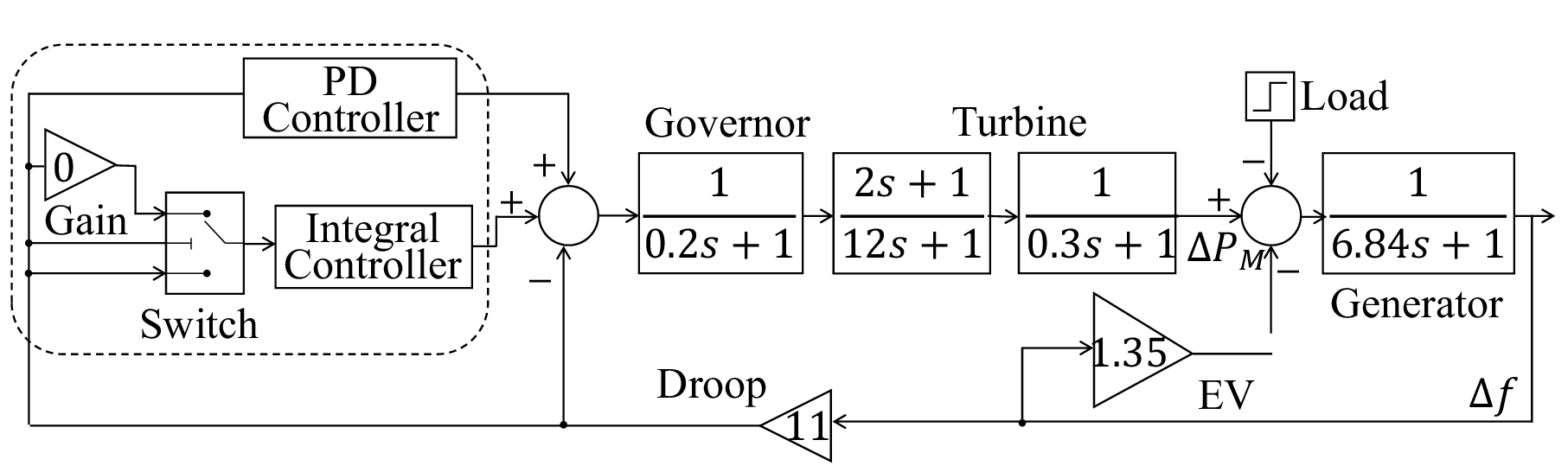}
    \caption{Framework of the simulated LFC system. The system uses an integral-separated PID controller, with logic that switches based on feedback signals.}
    \label{fig_framework_flc}
\end{figure}

\noindent\textbf{System Description. } We simulated a power grid Load Frequency Control (LFC) system as the first ICS case study. 
The system parameters are aligned with prior research simulating a power system of Great Britain~\cite{obaid2016fuzzy}. As depicted in Figure~\ref{fig_framework_flc}, the controller is modeled using an integral-separated PID controller. The integral component of the PID is crucial for eliminating static errors; however, in instances where the input errors are substantial, the integral controller accumulates these errors, potentially driving the system's actuators into a state of saturation -- a phenomenon extensively studied and referred to as controller windup in the literature~\cite{shin2011anti, da2018analysis, jia2019improved}. Integral-separated PID control is an effective strategy to circumvent windup; it operates by switching the controller to Proportional-Derivative (PD) mode when the controller's input errors exceed a predefined threshold, and reverting to PID mode when inputs fall below this threshold.
{The training dataset consists of 180,000 simulated points, while the testing dataset includes 182,500 points, with a 39.6\% anomaly rate generated through various attack scenarios described in the attack model section.}

\noindent\textbf{Attack Model.} We launch three types of attacks on the LFC system, aimed at disturbing the frequency deviation and potentially causing the frequency control to fail. The first attack is \textit{Malicious Controller Output Injection Attack}. Similar to previous methods of injecting malicious data~\cite{liu2019event}, we assume these attacks occur randomly, governed by a nonlinear function, formulated as follows:
\begin{equation}
    \label{eq_malicious_injection}
    \begin{aligned}
    \bar{y}(t) &= \beta(t)g(y(t))+(1-\beta (t))y(t)\\
    g(y) &= -tanh(Gy)
    \end{aligned}
\end{equation}
Here, $G$ represents the constant for the upper bound, and $\beta\in \{0,1\}$ is a Bernoulli variable with an expectation ranging from $[0,0.3]$. $g(y)$ generates the malicious data with the restriction $\parallel g(y) \parallel^2 \le \parallel Gy \parallel^2$.
 
The second attack is the \textit{Controller Configuration Tampering Attack}, where the parameters of the PID controller are altered based on a random variable:
\begin{equation}
    \label{eq_tampering_config}
    \begin{aligned}
    & KP=KP+\xi_1, KI=KI+\xi_2, KD=KD+\xi_3\\
    & KP\ge 0, KI\ge 0, KD\ge 0
    \end{aligned}
\end{equation}
where $\xi_1, \xi_2, \xi_3$ follow a Gaussian distribution $\mathcal{N}(0,1)$. $KP$, $KI$, and $KD$ represent the proportional, integral, and derivative parameters of a PID controller, respectively.

The third attack is \textit{Controller Disabling Attack}. Specifically, $KP=0, KI=0, KD=0$. While this attack shares similarities with tampering with the PID configurations, it differs in that the system's PID configuration returns to normal after a short period. This attack can be formulated as follows:
\begin{equation}
    \label{eq_disable_config}
    \begin{aligned}
    & \text{PID config = normal}, t \in [0, T_{off})\\
    & \text{PID config = attack}, t \in [T_{off}, T],
    \end{aligned}
\end{equation}
where the controller outputs are divided into $N$ periods, and each period lasts for $T$ time slots. $T_{off}$ is the starting time step for attacks in one 
period.

\begin{table}[t]
    \caption{The controller identification results.}
    \begin{center}
    \label{table_results_lfc}
    \resizebox{0.9\linewidth}{!}{
    \begin{tabular}{|c|c|c|}
        \hline
        Mode & Ground truth & Prediction\\
        \hline
        Controller-1 & $\frac{3.3s+2.1}{s+1}$ & $\frac{3.874s+2.599}{1.182s+1.231}$\\
        \hline
        {Controller-2} & $\frac{3.3s^2+2.7s+0.6}{s^2+s}$  & $\frac{3.027s^2+2.099s+0.4469}{s^2+0.6936s}$\\
        \hline
        {Switch threshold} & $0.1$ & $0.0979$\\
        \hline
    \end{tabular}
    }
    \end{center}
    
\end{table}

\begin{table*}[htbp]
    \begin{center}
    \caption{The comparison of attack detection in the load frequency control system.}
    \label{table_comparison_lfcattack}
    \resizebox{\textwidth}{!}{
    \begin{threeparttable}
    \begin{tabular}{ c|c|ccc|ccc }
        \toprule
        & Model & F1 & Precision & Recall & F1$^*$ & Precision$^*$ & Recall$^*$\\
        \midrule
        \multirow{10}{*}{\shortstack{Malicious\\Controller\\Output\\Injection}} & CascadedNN & $0.9916\pm 0.0126$ & $0.9937\pm 0.011$ & $0.9896\pm 0.0166$ & $0.9939\pm 0.0096$ & $1.0\pm 0.0$ & $0.988\pm 0.0189$\\
        & RNN & $0.9795\pm 0.0315$ & $0.9901\pm 0.0222$ & $0.9700\pm 0.0463$ & $0.9822\pm 0.0273$ & $0.9927\pm 0.0192$ & $0.9723\pm 0.0381$\\
        & EncDec-AD & $0.3357\pm 0.1028$ & $0.2262\pm 0.0706$ & $0.6690\pm 0.1966$ & $0.3543\pm 0.0973$ & $0.2430\pm 0.0735$ & $0.6749\pm 0.1526$\\
        & DAGMM & $0.2732\pm 0.1333$ & $0.1646\pm 0.0889$ & $1.0\pm 0.0$ & $0.2826\pm 0.0152$ & $0.1646\pm 0.1054$ & $1.0\pm 0.0$\\
        & USAD & $0.8521\pm 0.1802$ & $0.7772\pm 0.2359$ & $0.9991\pm 0.0003$ & $0.8943\pm 0.1169$ & $0.8323\pm 0.1768$ & $0.9915\pm 0.0159$\\
        & OmniAnomaly & $0.8121\pm 0.1011$ & $0.6964\pm 0.1595$	& $0.9963\pm 0.0063$ & $0.9874\pm 0.0213$	& $0.9961\pm 0.0094$ & $0.9797\pm 0.0416$\\
        & NSIBF & $0.8727\pm 0.0929$ & $0.8683\pm 0.1776$ & $0.8862\pm 0.0037$ & $0.9391\pm 0.0860$ & $0.9107\pm 0.1262$ & $0.9717\pm 0.0398$\\
        & NSIBF-RECON & $0.9245\pm 0.1009$& $0.8695\pm 0.1768$& $0.9981\pm 0.0012$& $0.9318\pm 0.0964$& $0.8940\pm 0.1498$ & $0.9776\pm 0.0316$\\
        & NSIBF-PRED & $0.9245\pm 0.1009$ & $0.8695\pm 0.1768$ & $0.9981\pm 0.0012$ & $0.9383\pm 0.0872$ & $0.9030\pm 0.1370$ & $0.9803\pm 0.0278$\\
        & SRLR (Ours) & $\bm{0.9980\pm 0.0013}$ & $1.0\pm 0.0$ & $0.9961\pm 0.0027$ & $\bm{0.9982\pm 0.0015}$ & $1.0\pm 0.0$ & $0.9964\pm 0.0030$\\
        
        \midrule
        \multirow{10}{*}{\shortstack{Controller\\Configuration\\Tampering\\Attack}} & CascadedNN & $0.9736\pm 0.0571$ & $0.9582\pm 0.0963$ & $0.9951\pm 0.0141$ & $0.9999\pm 0.0$ & $1.0\pm 0.0$ & $1.0\pm 0.0$\\
        & RNN & $0.9286\pm 0.0841$ & $0.9069\pm 0.1285$ & $0.9662\pm 0.0655$ & $0.9999\pm 0.0$ & $1.0\pm 0.0$ & $1.0\pm 0.0$\\
        & EncDec-AD & $0.7721\pm 0.1402$ & $0.6881\pm 0.2218$ & $0.9544\pm 0.0914$ & $0.9960\pm 0.0112$ & $0.9922\pm 0.0214$ & $1.0\pm 0.0$\\
        & DAGMM & $0.7014\pm 0.0923$ & $0.5485\pm 0.1074$ & $0.9950\pm 0.0161$ & $0.7087\pm 0.0915$ & $0.5560\pm 0.1074$ & $1.0\pm 0.0$\\
        & USAD & $0.7555\pm 0.1274$ & $0.6956\pm 0.2275$ & $0.9201\pm 0.1359$ & $0.9824\pm 0.0242$ & $0.9664\pm 0.0456$ & $1\pm 0.0$\\
        & OmniAnomaly & $0.7058\pm 0.0887$ & $0.5552\pm 0.1015$ & $0.9872\pm 0.0319$ & $0.9399\pm 0.0848$	& $0.8964\pm 0.1279$ & $1.0\pm 0.0$\\
        & NSIBF & $0.751\pm 0.0621$ & $0.6717\pm 0.0657$ & $0.8519\pm 0.0541$ & $0.9877\pm 0.0151$ & $0.9759\pm 0.0295$ & $1.0\pm 0.0$\\
        & NSIBF-RECON & $0.7080\pm 0.0458$ & $0.5491\pm 0.0551$ & $0.9996\pm 0.0004$ & $0.9744\pm 0.0258$ & $0.9755\pm 0.0141$ & $0.9735\pm 0.0374$\\
        & NSIBF-PRED & $0.7082\pm 0.0462$ & $0.5493\pm 0.0554$ & $1.0\pm 0.0$ & $0.9713\pm 0.0204$ & $0.9695\pm 0.0036$ & $0.9735\pm 0.0374$\\
        & SRLR (Ours) & $\bm{0.9955\pm 0.0077}$ & $0.9981\pm 0.0043$ & $0.9929\pm 0.0114$ & $0.9999\pm 0.0$ & $1.0\pm 0.0$ & $1.0\pm 0.0$\\
        \midrule
        \multirow{10}{*}{\shortstack{Controller\\Disabling\\Attack}} & CascadedNN & $0.9727\pm 0.0392$ & $0.9707\pm 0.0455$ & $0.9757\pm 0.0433$ & $0.9999\pm 0.0$ & $1.0\pm 0.0$ & $1.0\pm 0.0$\\
        & RNN & $0.8857\pm 0.0823$ & $0.8853\pm 0.0706$ & $0.8899\pm 0.1079$ & $0.9888\pm 0.0185$ & $0.9801\pm 0.0318$ & $0.9981\pm 0.0083$\\
        & EncDec-AD & $0.5026\pm 0.0796$ & $0.3946\pm 0.1230$ & $0.8703\pm 0.2398$ & $0.9139\pm 0.0454$ & $0.8912\pm 0.0499$ & $0.9444\pm 0.0860$\\
        & DAGMM & $0.4824\pm 0.0790$ & $0.3213\pm 0.0680$ & $0.9991\pm 0.0020$ & $0.4904\pm 0.0805$ & $0.3284\pm 0.0696$ & $1.0\pm 0.0$\\
        & USAD & $0.5056\pm 0.0734$ & $0.3778\pm 0.1307$ & $0.9394\pm 0.1854$ & $0.8285\pm 0.0817$ & $0.7738\pm 0.1005$ & $0.9122\pm 0.1207$\\
        & OmniAnomaly & $0.7229\pm 0.1020$ & $0.6004\pm 0.1238$	& $0.9324\pm 0.0912$ & $0.9864\pm 0.0276$	& $0.9745\pm 0.0506$ & $1.0\pm 0.0$\\
        & NSIBF & $0.5839\pm 0.0206$ & $0.5165\pm 0.0518$ & $0.6907\pm 0.1476$ & $0.8593\pm 0.0085$ & $0.8483\pm 0.0066$ & $0.8705\pm 0.0105$\\
        & NSIBF-RECON & $0.5012\pm 0.0137$ & $0.3344\pm 0.0122$ & $1.0\pm 0.0$ & $0.7578\pm 0.0151$ & $0.6221\pm 0.0119$ & $0.9691\pm 0.0204$\\
        & NSIBF-PRED & $0.5009\pm 0.0133$ & $0.3342\pm 0.0119$ & $0.9994\pm 0.0007$ & $0.7699\pm 0.0323$ & $0.6379\pm 0.0456$ & $0.9725\pm 0.0030$\\
        & SRLR (Ours) & $\bm{0.9822\pm 0.0264}$ & $0.9975\pm 0.0045$ & $0.9685\pm 0.047$ & $\bm{0.9999\pm 0.0001}$ & $1.0\pm 0.0$ & $0.9999\pm 0.0001$\\
        \bottomrule
    \end{tabular}
        \begin{tablenotes}
        \footnotesize
        \item[1] F1$^*$, Precision$^*$ and Recall$^*$ refer to the metrics calculated using the point-adjusted evaluation strategy.
        \end{tablenotes}
    \end{threeparttable}
    }
    \end{center}
\end{table*}

\noindent\textbf{Attack Detection. } We utilize the SRLR in multi-mode logic recovery setting to identify the switched controller logic (i.e., logic switch from PD control to PID control) utilizing just the controller inputs and outputs as training data. Subsequently, we apply the identified controllers' logic to the dataset and compute the errors, determining the time points where a mode switch occurs. Employing the Mahalanobis distance-based filtering method~\cite{estimator1999fast}, we then utilize the DSR model to determine the logic switch threshold. {The training process requires 5.7 hours.} The results presented in Table~\ref{table_results_lfc}, demonstrate the accurate recovery of the switched controller logic and the corresponding switch results.

The results are presented in Table~\ref{table_comparison_lfcattack}. It is evident from the table that our model consistently outperforms the baselines across all three types of attack detection scenarios. This superiority can be attributed to the accuracy with which our model recovers the switched controllers' logic and the associated switch conditions. Notably, the CascadedNN exhibits the second-best performance owing to its fully connected characteristics. {SI based methods, including our proposed model, RNN, CascadedNN and NSIBF, demonstrate superior performance compared to autoencoder-based approaches and generative approaches.}

In the case of malicious controller output injection, our model effectively detects nearly all instances of malicious data. This success stems from the model's ability to accurately identify between malicious and normal data. Conversely, autoencoder methods struggle to distinguish between normal and malicious data, as they primarily focus on capturing the distribution of training and testing data without explicitly considering controller logic. When the controller configuration is tampered with, particularly if the tampered configuration closely resembles the normal configuration, the performance of our model, RNN, and CascadedNN experiences a slight decrease. {However, EncDec-AD and DAGMM methods exhibit improved performance in this scenario, as the tampered configuration persists for an extended period, leading to a continuous presence of attacked points. In contrast, the controller disabling attack poses a stealthier threat, as it only lasts for a brief duration and may go undetected by system identification-based methods. Moreover, the relatively fewer attacked points in comparison to controller configuration tampering also contribute to decreased performance for NSIBF, EncDec-AD, DAGMM and USAD. OmniAnomaly demonstrates relatively stable performance across three types of attacks, due to its ability to model the dynamic characteristics of time series data. Additionally, the point-adjusted metrics yield higher values compared to standard metrics, as they evaluate continuous attack segments as single anomaly events. This approach increases the likelihood of detection, particularly in cases such as controller configuration tampering, where identifying any point within the segment is sufficient to flag the entire attack.} In summary, leveraging the identified switch controllers' logic and associated switch conditions enables accurate simulation of system behavior, facilitating efficient detection of attacks.

\subsection{Secure Water Treatment Testbed}
The SWaT platform, developed by iTrust SUTD~\cite{goh2017dataset}, represents a scaled-down emulation of real-world water treatment systems with six-stage processes. In the first 7 days, the SWaT system operated under normal conditions, followed by four days of operation under attack conditions. {The dataset, collected from 51 sensors and actuators, comprises 495,000 training samples and 449,919 testing samples, with approximately 12.14\% of the testing set containing attacked points.} These attacks were initiated by tampering with sensor or actuator values, introducing perturbations into the system dynamics.

The SWaT dataset has been extensively utilized in the machine learning community~\cite{tuli2022tranad, feng2021time}. 
Prior research primarily focuses on detecting attacks and triggering alarms without providing a comprehensive explanation for the occurrence of these attacks to system operators. Moreover, these studies often treat all attacks as a single category, overlooking the diverse nature of attack types. We address these limitations by conducting an in-depth analysis of attack behaviors, categorizing them into three distinct types.  
Subsequently, we apply our mode process identification method to recover controller logic, enabling the generation of rules for attack detection. Leveraging these generated rules, we detect attacks and provide detailed explanations to system operators when alarms are triggered. This approach not only enhances the accuracy of attack detection but also facilitates a deeper understanding of the underlying causes behind these security breaches.


\noindent\textbf{Attack Categories. } We classify the attacks observed in the SWaT dataset into three distinct types: a) \textit{Actuator Value Tampering Attack}: The actuators themselves are targeted, resulting in the manipulation of their values. b) \textit{Actuator-Dependent Sensor Value Tampering Attack}: The attacker tampers with sensor values that share a consistent relationship with specific actuators. This relationship becomes disrupted upon tampering with the sensor values. c) \textit{Actuator-Independent Sensor Value Tampering Attack}: The attacker targets sensor values that are not directly related to the control logic of the PLCs governing the actuators. 
For our analysis, SWaT dataset
encompasses 36 consecutive attacks, with 19 attacks classified as actuator value tampering, 6 attacks categorized as actuator-dependent sensor value tampering, and 11 attacks falling under the actuator-independent sensor value tampering category. This classification allows for a comprehensive understanding of the diverse attack scenarios present within the SWaT dataset.

\noindent\textbf{Attack Detection. } 
\begin{figure*}[htbp]
    \centering
    \subfloat[Actuator value tampering Attack.]{
    \begin{minipage}[b]{0.33\textwidth}
    \includegraphics[width=1\linewidth]{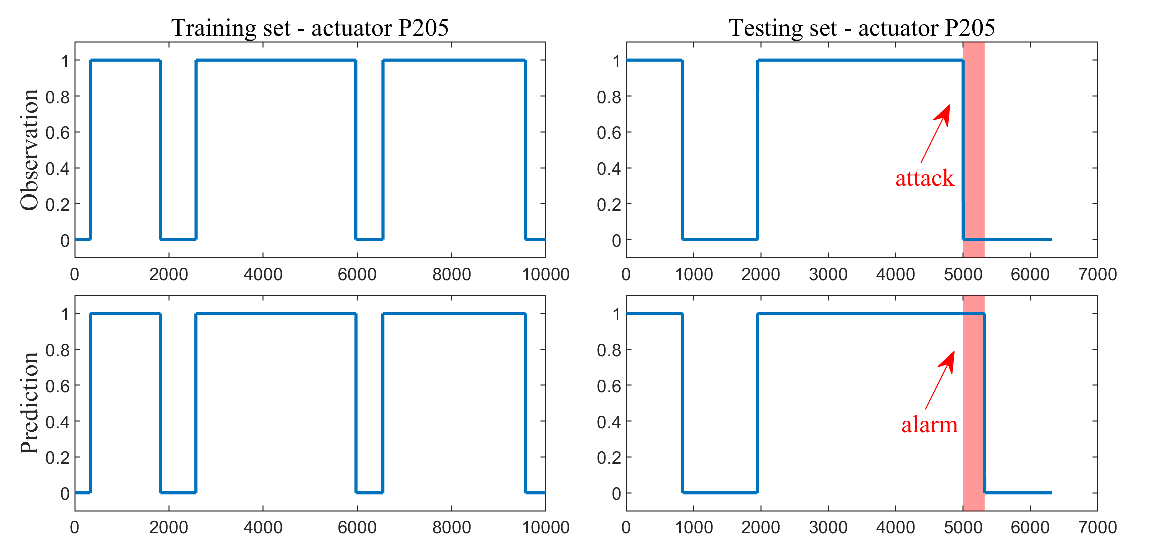}
    \end{minipage}}
    \subfloat[Actuator-dependent sensor value tampering]{
    \begin{minipage}[b]{0.33\textwidth}
    \includegraphics[width=1\linewidth]{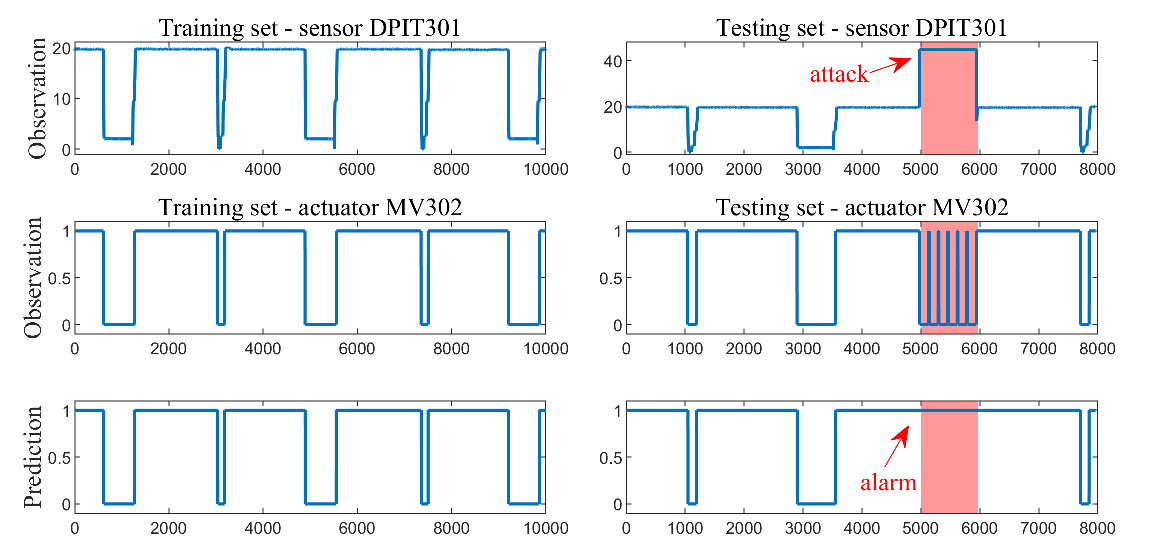}
    \end{minipage}}
    \subfloat[Actuator-independent sensor value tampering]{
    \begin{minipage}[b]{0.33\textwidth}
    \includegraphics[width=1\linewidth]{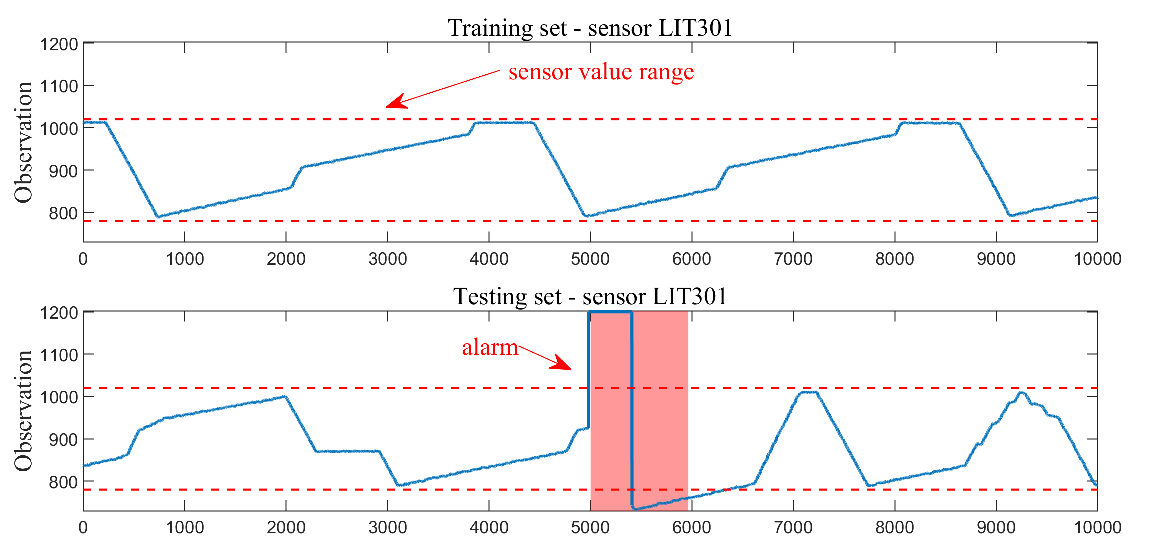}
    \end{minipage}}
    \caption{Examples of attack detection results on the SWaT dataset with red areas highlighting the intervals where actual attacks occur. (a) The observed and predicted values for the pump P205. The left sub-figure shows 
    the training set and the right sub-figure shows the results on the testing set. (b) observations of differential pressure DPIT301 and the observed and predicted behavior of valve MV302 on the training and testing sets. (c) Observations of level transmitter LIT301 on training and testing sets.}
    \label{fig_swat_attack}
\end{figure*}
{We grouped the 51 variables into six categories corresponding to the distinct physical processes within the system. Each group encompasses both actuators and sensors that relate to its specific physical process. It is assumed that some of the actuators are controlled by the value of sensors and this can be expressed through a mathematical formulation. Upon examining the actuator data, it was observed that actuators operate in binary states, either "on" or "off", with these states typically determined by time-domain control logic. To reflect this behavior, SRLR is operated within the time domain on the SWaT system. To further enhance modeling accuracy, we introduced a step function. This adjustment not only better captures the binary nature of actuator outputs but also reduces the SRLR search space. With this configuration, SRLR is capable of capturing nonlinear relationships.}

The dynamic relationship between an actuator and its sensors is represented by the following equation:
\begin{equation}
    \label{eq_swat_logic}
    \begin{aligned}
    y(t) &= Step(f(x_1(t),x_1(t-\tau),x_2(t),x_2(t-\tau)...))\\
    &= \left\{
        \begin{aligned}
        1, f(x)\ge 0 \\
        0, f(x)< 0
        \end{aligned}
        \right.  
    \end{aligned}
\end{equation}
where $y(t)$ represents the actuator value at time step $t$, $x(t)$, and $x(t-\tau)$ denote the current and delayed sensor values within the same group, respectively. $step()$ represents the step function. The function $f(x)$ reveals the relationship between the sensors and the actuator, with the actuator activated if $f(x)\ge 0$, and deactivated otherwise.

Employing the proposed single-mode logic recovery approach, we endeavored to recover the truth logic of the function $f(x)$ by identifying all relationships that invariantly hold true within the training dataset. {Training on the SWaT dataset takes approximately 4.08 hours.} This process generated 15 definitive rules that remained valid across the training set. Subsequently, these rules were applied to simulate the corresponding actuator values within the testing set, facilitating the detection of anomalies indicative of potential attacks.

\begin{lstlisting}[caption={Examples of invariant rules on the training dataset},captionpos=b, label=lst:swat_rules, basicstyle=\small]
def rule_1(P205_obs):
    x1=AIT202; x2=AIT203; x3=FIT201
    x4=AIT203_delay; x5=FIT201_delay
    P205_pred = step(x1**3*x3-x2*x5+x2-2*x4)
    error = abs(P205_pred-P205_obs)
    return error
def rule_2(MV302_obs):
    x1=DPIT301; x2=FIT301;
    x3=LIT301; x4=FIT301_delay
    MV302_pred = step(x1-0.0115*x3)*(x2+x3+x4)
    error = abs(MV302_pred-MV302_obs)
    return error
def rule_3(LIT301_obs):
    if LIT301_obs>1020 or LIT301_obs<780:
        error = 1
    else: 
        error = 0
    return error
\end{lstlisting}

For the first type of attack, i.e., the Actuator Value Tampering Attack, detection becomes feasible if rules governing actuator value generation are correctly established by our proposed SRLR model. Leveraging the single-mode process identification method, we derive the formula governing the relationship between actuator P205 and sensors in process 2, subsequently yielding Rule 1. As depicted in List~\ref{lst:swat_rules}, rows 2 and 3 are sensor observations assignments, row 5 denotes the generated formula, and row 6 represents the error between predictive P205 and observed P205. As illustrated in Figure~\ref{fig_swat_attack}(a), predictive P205 consistently aligns with observations in the training data. However, in the test data, the actuator P205 is maliciously manipulated from ``on" to ``off" during the attack, resulting in a discrepancy between predictive and observed P205 values and triggering an alarm. Thus, this actuator tampering attack is effectively detected.

For the second type of attack, i.e., the Actuator-Dependent Sensor Value Tampering Attack, the sensor value is linked to the actuator. Although the sensor value does not directly control the output of the PLC, when the sensor value is tampered with, the system's status becomes abnormal, disrupting the previous invariant relationship between the sensor and actuator. Therefore, identifying the relationship between sensors and their corresponding actuators can detect this type of attack. As illustrated in List~\ref{lst:swat_rules} rows 9 to 14, we establish the relationship between the actuator MV302 and the sensor value DPIT301. This equation consistently holds true in the training data, as depicted in Figure~\ref{fig_swat_attack}(b). In the testing data, the attacker tampers with the DPIT301 value, causing oscillations in the MV302 value. However, when simulating the logic in List~\ref{lst:swat_rules} row 12, the MV302 is predicted to remain consistently open, triggering the alarm and thus, detecting the attack successfully.

For the third type of attack, i.e., the Actuator-Independent Sensor Value Tampering Attack, the sensor value neither controls the PLC's output nor has an invariant relationship with actuators. Therefore, we cannot employ the mode equation identification method to detect this type of attack. However, we can utilize the normal range observed in the training data to constrain the corresponding sensor values in the testing data. As illustrated in Figure~\ref{fig_swat_attack}(c), although the LIT301 participates in the calculation of MV302, we find that tampering with the LIT301 value does not affect the prediction for the MV302. Then, we analyze the range of LIT301 in the training data and generate a rule to limit the range of LIT301 in the testing data. As shown in List~\ref{lst:swat_rules} rows 16 to 21, we observe that LIT301 in the training data falls within the range of [780,1020]. In the testing data, when the value of LIT301 is tampered to exceed the normal range, this attack can succeed, as depicted at the bottom of Figure~\ref{fig_swat_attack}(c). Thus, while this type of attack cannot be detected by the proposed model, we can still apply the range rule to detect out-of-range values.

We observe some (false) alarms when there is no attack. For instance, SRLR identifies a relationship between MV101 and sensor FIT101. However, there are instances of delayed responses / updates in the testing data, leading to triggering of alarms. In this case, SRLR recovers correct logic, but there is an additional task (beyond SRLR) to distinguish whether the violation of the rule is due to attack or some data anomaly.

\begin{figure}[t]
    \centering
    \includegraphics[scale=0.365]{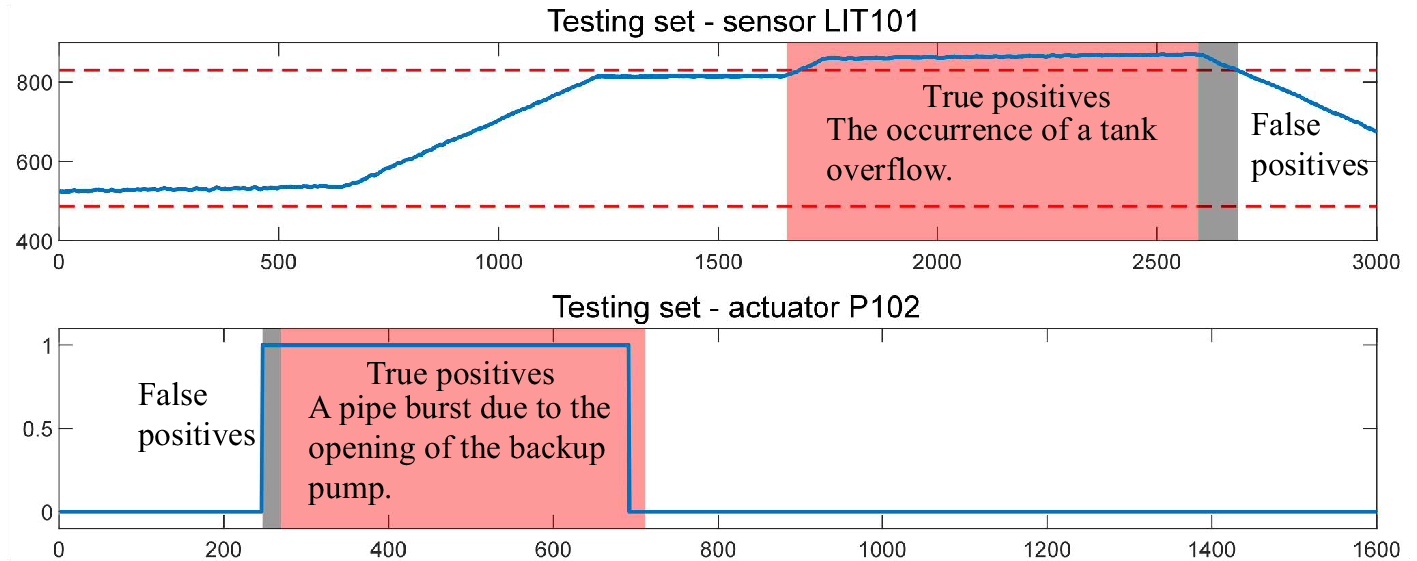}
    \caption{The case study for false positives.}
    \label{fig_false_positives}
\end{figure}

\begin{table*}[htbp]
    \begin{center}
    \caption{The comparison of attack detection in the SWaT system.}
    \label{table_comparison_swatattack}
    \resizebox{\textwidth}{!}{
    \begin{threeparttable}
    \begin{tabular}{ c|ccc|ccc }
        \toprule
        Model & F1 & Precision & Recall & F1$^*$ & Precision$^*$ & Recall$^*$\\
        \midrule
        CascadedNN & $0.6829\pm 0.0619$ & $0.7473\pm 0.1647$ & $0.6433\pm 0.0296$ & $0.9010\pm 0.0062$ & $0.9499\pm 0.0144$ & $0.8573\pm 0.0170$\\
        RNN & $0.6963\pm 0.0735$ & $0.9386\pm 0.0457$ & $0.5612\pm 0.1032$ & $0.9048\pm 0.0184$ & $0.9471\pm 0.0242$ & $0.8668\pm 0.0340$\\
        EncDec-AD & $0.6940\pm 0.0405$ & $0.8272\pm 0.1517$ & $0.6080\pm 0.0506$ & $0.8762\pm 0.0185$ & $0.9132\pm 0.0264$ & $0.8423\pm 0.0183$\\
        DAGMM & $0.2165\pm 0.0$ & $0.1213\pm 0.0001$ & $0.9999\pm 0.0$ & $0.8533\pm 0.0097$ & $0.8602\pm 0.0116$ & $0.8464\pm 0.0079$\\
        USAD & $0.6807\pm 0.0977$ & $0.8240\pm 0.2933$ & $0.6188\pm 0.0570$ & $0.8534\pm 0.0172$ & $0.9442\pm 0.0540$ & $0.7836\pm 0.0703$\\
        OmniAnomaly & $0.7540\pm 0.0018$ & $0.9962\pm 0.0007$ & $0.6066\pm 0.0026$ & $0.7993\pm 0.0053$ & $0.9993\pm 0.0009$ & $0.6660\pm 0.0076$\\
        NSIBF & $0.6983\pm 0.0053$ & $0.7725\pm 0.0116$ & $0.6371\pm 0.0017$ & $0.8691\pm 0.0096$ & $0.8273\pm 0.0156$ & $0.9164\pm 0.0362$\\
        NSIBF-RECON & $0.7603\pm 0.0037$ & $0.9669\pm 0.0316$ & $0.6270\pm 0.0182$ & $0.8690\pm 0.0217$ & $0.9481\pm 0.0103$ & $0.8023\pm 0.0297$\\
        NSIBF-PRED & $0.7417\pm 0.0037$ & $0.9468\pm 0.0438$ & $0.6106\pm 0.0216$ & $0.8680\pm 0.0125$ & $0.8660\pm 0.0337$ & $0.8710\pm 0.0216$\\
        SRLR (Ours) & $\bm{0.8329\pm 0.0007}$ & $0.9816\pm 0.0007$ & $0.7233\pm 0.0015$ & $\bm{0.9299\pm 0.0096}$ & $0.9353\pm 0.0197$ & $0.9247\pm 0.0$\\
        \bottomrule
    \end{tabular}
        \begin{tablenotes}
        \footnotesize
        \item[1] F1$^*$, Precision$^*$ and Recall$^*$ refer to the metrics calculated using the point-adjusted evaluation strategy.
        \end{tablenotes}
    \end{threeparttable}
    }
    \end{center}
\end{table*}

\noindent\textbf{Comparison with Baselines. } {We compare our results with various deep learning-based methods and neural system identification models.
As depicted in Table~\ref{table_comparison_swatattack}, our model surpasses EncDec-AD, generation-based models USAD, DAGMM, OmniAnomaly, and neural system identification models CascadedNN, RNN and NSIBF. Specifically, although NSIBF (neural system identification method) can numerically identify systems, it fails to generate a clear expression, thereby lacking explanatory power for ICS operators.} Once the true rules are generated by SRLR, attacks can be detected effectively, as most attacks involve control logic modifications, output value tampering, or sensor readings falling outside valid ranges. Consequently, the number of false negatives is significantly lower compared to deep learning methods, as evidenced by the recall metric in Table~\ref{table_comparison_swatattack}. 
As previously discussed, the rare cases of misdetection are due to environmental sampling delays. As illustrated in Figure~\ref{fig_false_positives}, these false positives occur either due to the persistence of the hazard (e.g., tank overflow) or due to the discrepancies of actuator-sensor relationship rules generated by SRLR. Although the data is labeled as normal in the gray area shown in Figure~\ref{fig_false_positives}, these false alarms are justifiable in real-world systems. For the first false alarm, despite the valve being closed, the water level takes time to return to normal, leading to a delay that results in false positives (grey area). For the second alarm, our model detects an immediate violation of the actuator values as soon as the pump is activated, even though the pipe has not yet burst. This early detection leads to false positives. Furthermore, the false positive rate of SRLR is $0.0068$. We already consider the noise and outliers, hence the number of false positives can be kept to a minimum.
This results in a precision score consistently over 90\%. The ICS-specific design of SRLR enables it to handle complex systems effectively, allowing it to generate rules for attack detection that minimize both false positives and false negatives. This design ensures superior F1 performance, providing a robust solution for detecting deviations in PLC behavior.

\section{Related Work and Limitations}

\subsection{{Logic Recovery}}
Logic recovery aims to infer the underlying PLC logic solely from input-output data. This concept overlaps with system identification and symbolic regression, as all three approaches focus on deriving mathematical representations from observed data. In this section, we review these related methods to contextualize our approach.

\textbf{System Identification.} System identification (SI) has been a prominent research interest (see~\cite{ljung2010perspectives} and references therein), evolving from the identification of linear systems to encompass non-linear and hybrid system identification. Classical SI methods like ARX, FIR, and ARMAX employ linear combinations of inputs to predict system outputs, and this paradigm extends to nonlinear systems in methods such as NARX~\cite{ljung2020deep} and NARMAX~\cite{billings2013nonlinear}.
Some SI methods detect multi-mode systems by fine-tuning the parameters of a predefined nonlinear model, such as the linear piecewise affine system identification method~\cite{bako2011identification} and the hybrid system identification method~\cite{yuan2019data}. In comparison with these methods, SRLR offers greater flexibility, as it does not require prior knowledge of the model structure.

{\textbf{Symbolic Regression.} Existing symbolic regression methods can be broadly categorized into three groups: heuristic methods, pre-trained methods, and reinforcement learning methods. Heuristic methods~\cite{schmidt2009distilling, mundhenk2021symbolic} employ evolutionary techniques like GP to explore the search space for optimal solutions. However, these methods face scalability issues and are sensitive to hyperparameters~\cite{petersen2021deep}. Pre-trained methods~\cite{holt2023deep, valipour2021symbolicgpt} leverage pre-trained models to provide prior knowledge for guiding the initial search space exploration. However, these methods primarily focus on numerical results, making it challenging to handle symbols due to their discrete characteristics. Reinforcement learning methods~\cite{petersen2021deep, landajuela2021discovering}, on the other hand, sample symbols and address the discrete sampling process using policy gradient training. This approach has proven to be efficient and has achieved SOTA results in symbolic regression benchmarks. Thus, we adopt DSR~\cite{petersen2021deep} in this work as the backbone framework for identifying control logic.}
 
\textbf{Logic Recovery.} The concept of logic recovery has been explored extensively~\cite{ljung2010perspectives, ljung2020deep}. For instance, Schmidt and Lipson~\cite{schmidt2009distilling} employed an evolutionary-based symbolic regression method to uncover physical laws through motion-tracking data, striking a balance between computational complexity and accuracy of discovery with the Pareto Front.
Ly et al.~\cite{ly2012learning} proposed the Cluster SR designed to identify time-invariant nonlinear hybrid systems. They integrated GP and the Expectation Maximization (EM) algorithm to infer the equations governing each mode and introduced a classification method to recover the transition logic among distinct modes within the hybrid system.
In our experiments, existing clustering-based methods performed poorly in identifying complex nonlinear PLC logics. For example, mode memberships determined by Cluster SR were often sparse and inaccurate in complex systems, largely due to their reliance on soft assignments and the absence of continuity constraints. By contrast, SRLR achieved a 39\% improvement in BFR in noisy environments compared with Cluster SR. Furthermore, without key designs used in SRLR like the s-domain enhancement and complexity regularization, existing symbolic regression methods such as PySR and Operon failed to identify subsystems in load frequency control systems and automatic voltage regulators, including PID controllers.

Several studies have explored methods for reverse-engineering PLC logic. Reditus~\cite{qasim2020control} captures network packets generated during PLC program execution and reconstructs the logic using the decompilation tools provided by PLC workstations. However, this approach is limited to a small subset of PLCs; for example, recent platforms such as Siemens TIA Portal and Codesys do not support reverse-engineering functions. Another line of research proposes extracting PLC logic from compiled binaries through database mapping. In this method~\cite{kalle2019clik}, a database maps binary structures to their corresponding control program components, and similarity analysis is used to match target code segments. While effective in some cases, it is impractical to enumerate all mapping relationships needed to reconstruct complex control logic, especially those involving varied arithmetic operations. CLEVER~\cite{sang2024control} addresses this challenge by analyzing PLC control applications directly. Using heuristic algorithms, it identifies variable dependencies and sequentially decompiles the source code. However, its effectiveness is limited when variable dependencies are incomplete, leading to failures in recognizing certain numerical operations, such as cyclic shift functions in motor control programs.

In contrast, SRLR demonstrates strong capabilities in handling numerical calculations. First, it employs s-domain modeling to broaden the scope of recognized operations, enabling the modeling of motor logic in LFC systems. Second, SRLR focuses on recovering mathematical relationships between variables, making the precise code implementation less critical in some contexts. These strengths allow SRLR to operate independently of specific programming platforms while offering robust logic reconstruction capabilities.

\subsection{{Attack Detection}}
{Attack detection methods in ICSs are crucial for addressing threats in both cyber and physical layers and these methods can be broadly categorized into specification-based and ML-based approaches. Specification-based methods~\cite{khraisat2019survey, giraldo2018survey} derive rules from a database of known attacks and employ these rules to identify threats as they occur. These methods typically offer high accuracy and clear explanations; however, their major limitation lies in the difficulty of detecting zero-day attacks, which are not represented in the existing database. In contrast, ML-based methods~\cite{han2022learning, tuli2022tranad, feng2021time, audibert2020usad} leverage statistical or machine learning techniques to develop patterns of normal data, which facilitates the detection of zero-day attacks by recognizing patterns that deviate from the norm. Despite this capability, ML-based methods often suffer from a low true positive rate, particularly when attack data falls within the normal data range or does not significantly deviate from established patterns. Additionally, ML-based methods cannot generally provide clear, easily understandable explanations for their detection decisions. From the perspective of ICS security, related efforts can be broadly categorized into three areas: process-level protection, sensor data integrity, and PLC code integrity verification. In this section, we review representative work in each of these areas and discuss relevant AI explainability methods that related to the goals of our approach.}

{\textbf{Process-level Protection.} Regarding process control attacks, Hadziosmanovi et.al~\cite{hadvziosmanovic2014through} proposed monitoring the status of PLC variables and developing a regression model to predict variable traces, aiming to detect discrepancies that do not match the predicted patterns. However, in modern ICSs, PLC outputs are predominantly binary variables, such as commands for controlling valves or pumps in water treatment systems. This characteristic poses a significant challenge for time-series models, as accurately predicting binary outcomes often proves difficult. Tan et al.~\cite{tan2022cotoru} introduced CoToRu, a method for inferring the behavior of PLCs to detect attacks. Their approach involves parsing the PLC code to generate a state transition table, which is then used to derive rules for detecting threats. While their method is effective and achieves high accuracy, obtaining access to the PLC code can be challenging for third-party security teams in real-world scenarios.}

{\textbf{Sensor Data Integrity. } Numerous studies have focused on protecting sensor data integrity. A common approach involves using prediction models to forecast future sensor values. PyCRA~\cite{shoukry2015pycra}, for example, enhances these prediction methods by comparing current readings with historical data, employing a noise-reduction strategy to detect stealthy integrity attacks. Another widely-used method is the Linear Dynamical State-Space model, 
which simulates process behavior using known signals and compares the simulated values with sensor response. While effective for process-level data integrity detection, these methods struggle with predicting the outputs of controllers. SRLR, on the other hand, is specifically designed to counter logic attacks on controllers. Unlike the methods above, which focus on sensor data integrity, SRLR addresses the core logic of controllers, making it a complementary and orthogonal approach. Deploying SRLR alongside sensor data integrity protection methods can provide comprehensive protection against both process-level attacks (such as false data injection via sensor) and control logic tampering.}

{\textbf{Attesting PLC Code Integrity. } Attesting code integrity is another potential solution for addressing PLC logic tampering. This can be achieved through root-of-trust mechanisms in specialized hardware modules or checksum verification via updating PLC software. However, these methods often require invasive hardware modifications or PLC vendor's support~\cite{lin2023dnattest} and are vulnerable to time-of-check-to-time-of-use (TOCTOU) attacks~\cite{bratus2008toctou}. In contrast, SRLR can be integrated into existing PLC I/O monitoring platforms, such as commercial-ready level 0 monitoring solutions\footnote{https://sigasec.com}, to offer advanced logic attack detection in a non-invasive manner.}



\subsection{{Scope and Limitations.}}

{PLC logic covers a wide range of control strategies. From a control theory perspective, these strategies fall into two broad categories: open-loop control, such as simple switches, and closed-loop control, including PID controllers. Our experiments show that the SRLR framework effectively recovers certain open-loop behaviors, such as hysteresis relay logic, and accommodates variations of PID control commonly used in closed-loop systems. However, SRLR has limitations when applied to complex control strategies such as Model Predictive Control and $\mathcal{H}_\infty$ control. These approaches involve solving quadratic programming or semi-definite programming problems with intricate constraints, making it difficult to represent their logic through simple mathematical expressions. As a result, such advanced control methods fall outside the current scope of SRLR.}

{From a syntactic standpoint, PLC programs rely on a set of standardized function blocks, which can be grouped into key categories: Bitwise operations; Timers; Counters; Edge detection; Data operations; Control functions. Many of these constructs—such as bitwise logic and arithmetic—map directly to mathematical symbols, expanding the SRLR framework’s symbolic library, enabling effective handling of such logic. Timer functions can also be expressed mathematically. For example, the TON (on-delay) timer activates when its input signal remains high for a duration longer than a predefined threshold. This behavior can be represented as:}
\begin{equation}
    {Q(t) = H(I(t)\Delta t \sum_{\tau=0}^N I(t-\tau) - PT),}
\end{equation}
{where $I(t)$ is the input signal, $PT$ the preset time, $\Delta t$ the PLC scan cycle, and $H$ the Heaviside step function. The accumulated time $T_{acc}(t)=I(t)\Delta t \sum_{\tau=0}^N I(t-\tau)$, similar to an EWMA, is reset whenever the input is low. SRLR has been shown to recover EWMA-like patterns even in noisy environments, supporting its ability to model timer behaviors.}

{Other PLC structures can be similarly translated into mathematical representations, such as counters and edge triggers. These results collectively demonstrate that SRLR is capable of capturing basic PLC syntax. However, the mathematical modeling of more complex process control and sequential structures like interrupt and asynchronous event handling becomes significantly more complicated. These advanced control logics are currently beyond the scope of SRLR.}


While SRLR shows promising results in our experiments across various ICS domains, thorough evaluation with a more diverse set of control logic is needed to characterize SRLR's capability and applicability --- specifically, future work is needed to answer the following questions: which type of control logic SRLR can handle well, and which types it cannot? Is there a complexity bound for the control logic, beyond which, SRLR can no longer work reliably? If that is the case, can we find a reliable way to divide and conquer a complicated system using SRLR? We hope our work can inspire future research in these directions.

\section{Conclusion}
We presented SRLR, a novel approach to recover logic of PLC and generate rules for detecting logic manipulation attacks in ICSs. SRLR accurately infers formulas for both single-mode and multi-mode processes, recovering the control logic of PLCs. The recovered logic helps generate rules capable of detecting attacks such as controller configuration tampering, malicious controller output injection, and controller disabling. SRLR is tested on both a grid frequency control system and a water treatment testbed. Results indicate that SRLR outperforms existing deep anomaly detection models, while providing explainability to its decisions. In future work, we will evaluate SRLR under more complex control logic and determine the applicable range of such methods.

\section*{Acknowledgments}
This research is supported in part by the National Natural Science Foundation of China under the Grant No. 62371057; and in part by the National Research Foundation, Singapore, under its National Satellite of Excellence Programme “Design Science and Technology for Secure Critical Infrastructure: Phase II", under Award NRF-NCR25-NSOE05-0001. Any opinions, findings and conclusions or recommendations expressed in this material are those of the author(s) and do not reflect the views of National Research Foundation, Singapore. The work was done when Hao Zhou was a visiting PhD student in Singapore University of Technology and Design (SUTD).


\normalem
\bibliographystyle{IEEEtran}
\bibliography{ref}

@inproceedings{petersen2021deep,
  title={Deep symbolic regression: Recovering mathematical expressions from data via risk-seeking policy gradients},
  author={Petersen, Brenden K and Landajuela, Mikel and Mundhenk, T Nathan and Santiago, Claudio P and Kim, Soo K and Kim, Joanne T},
  booktitle={Proc. of the International Conference on Learning Representations},
  year={2021}
}

@article{yuan2019data,
  title={Data driven discovery of cyber physical systems},
  author={Yuan, Ye and Tang, Xiuchuan and Zhou, Wei and Pan, Wei and Li, Xiuting and Zhang, Hai-Tao and Ding, Han and Goncalves, Jorge},
  journal={Nature communications},
  volume={10},
  number={1},
  pages={4894},
  year={2019},
  publisher={Nature Publishing Group UK London}
}

@article{ly2012learning,
  title={Learning symbolic representations of hybrid dynamical systems},
  author={Ly, Daniel L and Lipson, Hod},
  journal={The Journal of Machine Learning Research},
  volume={13},
  number={1},
  pages={3585--3618},
  year={2012},
  publisher={JMLR. org}
}

@inproceedings{landajuela2021discovering,
  title={Discovering symbolic policies with deep reinforcement learning},
  author={Landajuela, Mikel and Petersen, Brenden K and Kim, Sookyung and Santiago, Claudio P and Glatt, Ruben and Mundhenk, Nathan and Pettit, Jacob F and Faissol, Daniel},
  booktitle={International Conference on Machine Learning},
  pages={5979--5989},
  year={2021},
  organization={PMLR}
}

@article{schmidt2009distilling,
  title={Distilling free-form natural laws from experimental data},
  author={Schmidt, Michael and Lipson, Hod},
  journal={science},
  volume={324},
  number={5923},
  pages={81--85},
  year={2009},
  publisher={American Association for the Advancement of Science}
}

@inproceedings{kumar2010self,
 author = {Kumar, M. and Packer, Benjamin and Koller, Daphne},
 booktitle = {Advances in Neural Information Processing Systems},
 editor = {J. Lafferty and C. Williams and J. Shawe-Taylor and R. Zemel and A. Culotta},
 pages = {},
 publisher = {Curran Associates, Inc.},
 title = {Self-Paced Learning for Latent Variable Models},
 url = {https://proceedings.neurips.cc/paper_files/paper/2010/file/e57c6b956a6521b28495f2886ca0977a-Paper.pdf},
 volume = {23},
 year = {2010}
}

@article{akaike1974new,
  title={A new look at the statistical model identification},
  author={Akaike, Hirotugu},
  journal={IEEE transactions on automatic control},
  volume={19},
  number={6},
  pages={716--723},
  year={1974},
  publisher={Ieee}
}

@book{saadat1999power,
  title={Power system analysis},
  author={Saadat, Hadi and others},
  volume={2},
  year={1999},
  publisher={McGraw-hill}
}

@article{ljung2009experiments,
  title={Experiments with identification of continuous time models},
  author={Ljung, Lennart},
  journal={IFAC Proceedings Volumes},
  volume={42},
  number={10},
  pages={1175--1180},
  year={2009},
  publisher={Elsevier}
}

@article{giraldo2018survey,
  title={A survey of physics-based attack detection in cyber-physical systems},
  author={Giraldo, Jairo and Urbina, David and Cardenas, Alvaro and Valente, Junia and Faisal, Mustafa and Ruths, Justin and Tippenhauer, Nils Ole and Sandberg, Henrik and Candell, Richard},
  journal={ACM Computing Surveys (CSUR)},
  volume={51},
  number={4},
  pages={1--36},
  year={2018},
  publisher={ACM New York, NY, USA}
}

@article{ljung2010perspectives,
  title={Perspectives on system identification},
  author={Ljung, Lennart},
  journal={Annual Reviews in Control},
  volume={34},
  number={1},
  pages={1--12},
  year={2010},
  publisher={Elsevier}
}

@article{ljung2020deep,
  title={Deep learning and system identification},
  author={Ljung, Lennart and Andersson, Carl and Tiels, Koen and Sch{\"o}n, Thomas B},
  journal={IFAC-PapersOnLine},
  volume={53},
  number={2},
  pages={1175--1181},
  year={2020},
  publisher={Elsevier}
}

@book{billings2013nonlinear,
  title={Nonlinear system identification: NARMAX methods in the time, frequency, and spatio-temporal domains},
  author={Billings, Stephen A},
  year={2013},
  publisher={John Wiley \& Sons}
}

@inproceedings{tan2022cotoru,
  title={CoToRu: automatic generation of network intrusion detection rules from code},
  author={Tan, Heng Chuan and Cheh, Carmen and Chen, Binbin},
  booktitle={IEEE INFOCOM 2022-IEEE Conference on Computer Communications},
  pages={720--729},
  year={2022},
  organization={IEEE}
}

@misc{mundhenk2021symbolic,
      title={Symbolic Regression via Neural-Guided Genetic Programming Population Seeding}, 
      author={T. Nathan Mundhenk and Mikel Landajuela and Ruben Glatt and Claudio P. Santiago and Daniel M. Faissol and Brenden K. Petersen},
      year={2021},
      eprint={2111.00053},
      archivePrefix={arXiv},
      primaryClass={cs.NE}
}

@misc{holt2023deep,
      title={Deep Generative Symbolic Regression}, 
      author={Samuel Holt and Zhaozhi Qian and Mihaela van der Schaar},
      year={2023},
      eprint={2401.00282},
      archivePrefix={arXiv},
      primaryClass={cs.LG}
}

@misc{valipour2021symbolicgpt,
      title={SymbolicGPT: A Generative Transformer Model for Symbolic Regression}, 
      author={Mojtaba Valipour and Bowen You and Maysum Panju and Ali Ghodsi},
      year={2021},
      eprint={2106.14131},
      archivePrefix={arXiv},
      primaryClass={cs.LG}
}

@article{masti2021learning,
  title={Learning nonlinear state--space models using autoencoders},
  author={Masti, Daniele and Bemporad, Alberto},
  journal={Automatica},
  volume={129},
  pages={109666},
  year={2021},
  publisher={Elsevier}
}

@inproceedings{goh2017dataset,
  title={A dataset to support research in the design of secure water treatment systems},
  author={Goh, Jonathan and Adepu, Sridhar and Junejo, Khurum Nazir and Mathur, Aditya},
  booktitle={Critical Information Infrastructures Security: 11th International Conference, CRITIS 2016, Paris, France, October 10--12, 2016, Revised Selected Papers 11},
  pages={88--99},
  year={2017},
  organization={Springer}
}

@inproceedings{hundman2018detecting,
  title={Detecting spacecraft anomalies using lstms and nonparametric dynamic thresholding},
  author={Hundman, Kyle and Constantinou, Valentino and Laporte, Christopher and Colwell, Ian and Soderstrom, Tom},
  booktitle={Proceedings of the 24th ACM SIGKDD international conference on knowledge discovery \& data mining},
  pages={387--395},
  year={2018}
}

@article{obaid2016fuzzy,
  title={Fuzzy hierarchal approach-based optimal frequency control in the Great Britain power system},
  author={Obaid, Zeyad Assi and Cipcigan, Liana Mirela and Muhssin, Mazin T},
  journal={Electric power Systems research},
  volume={141},
  pages={529--537},
  year={2016},
  publisher={Elsevier}
}

@article{shin2011anti,
  title={Anti-windup PID controller with integral state predictor for variable-speed motor drives},
  author={Shin, Hwi-Beom and Park, Jong-Gyu},
  journal={IEEE Transactions on Industrial Electronics},
  volume={59},
  number={3},
  pages={1509--1516},
  year={2011},
  publisher={IEEE}
}

@article{da2018analysis,
  title={Analysis of anti-windup techniques in PID control of processes with measurement noise},
  author={da Silva, Lucian R and Flesch, Rodolfo CC and Normey-Rico, Julio E},
  journal={IFAC-PapersOnLine},
  volume={51},
  number={4},
  pages={948--953},
  year={2018},
  publisher={Elsevier}
}

@article{jia2019improved,
  title={An improved particle swarm optimization (PSO) optimized integral separation PID and its application on central position control system},
  author={Jia, Lin and Zhao, Xinqiu},
  journal={IEEE Sensors Journal},
  volume={19},
  number={16},
  pages={7064--7071},
  year={2019},
  publisher={IEEE}
}

@article{liu2019event,
  title={Event-triggered $ H\infty $ load frequency control for multiarea power systems under hybrid cyber attacks},
  author={Liu, Jinliang and Gu, Yuanyuan and Zha, Lijuan and Liu, Yajuan and Cao, Jie},
  journal={IEEE Transactions on Systems, Man, and Cybernetics: Systems},
  volume={49},
  number={8},
  pages={1665--1678},
  year={2019},
  publisher={IEEE}
}

@misc{malhotra2016lstm,
      title={LSTM-based Encoder-Decoder for Multi-sensor Anomaly Detection}, 
      author={Pankaj Malhotra and Anusha Ramakrishnan and Gaurangi Anand and Lovekesh Vig and Puneet Agarwal and Gautam Shroff},
      year={2016},
      eprint={1607.00148},
      archivePrefix={arXiv},
      primaryClass={cs.AI}
}

@article{tuli2022tranad,
  title={Tranad: Deep transformer networks for anomaly detection in multivariate time series data},
  author={Tuli, Shreshth and Casale, Giuliano and Jennings, Nicholas R},
  journal={arXiv preprint arXiv:2201.07284},
  year={2022}
}

@inproceedings{feng2021time,
  title={Time series anomaly detection for cyber-physical systems via neural system identification and bayesian filtering},
  author={Feng, Cheng and Tian, Pengwei},
  booktitle={Proceedings of the 27th ACM SIGKDD Conference on Knowledge Discovery \& Data Mining},
  pages={2858--2867},
  year={2021}
}

@inproceedings{audibert2020usad,
  title={Usad: Unsupervised anomaly detection on multivariate time series},
  author={Audibert, Julien and Michiardi, Pietro and Guyard, Fr{\'e}d{\'e}ric and Marti, S{\'e}bastien and Zuluaga, Maria A},
  booktitle={Proceedings of the 26th ACM SIGKDD International Conference on Knowledge Discovery \& Data Mining},
  pages={3395--3404},
  year={2020}
}

@inproceedings{
zong2018deep,
title={Deep Autoencoding Gaussian Mixture Model for Unsupervised Anomaly Detection},
author={Bo Zong and Qi Song and Martin Renqiang Min and Wei Cheng and Cristian Lumezanu and Daeki Cho and Haifeng Chen},
booktitle={International Conference on Learning Representations},
pages={1--19},
year={2018},
url={https://openreview.net/forum?id=BJJLHbb0-},
}

@inproceedings{su2019robust,
  title={Robust anomaly detection for multivariate time series through stochastic recurrent neural network},
  author={Su, Ya and Zhao, Youjian and Niu, Chenhao and Liu, Rong and Sun, Wei and Pei, Dan},
  booktitle={Proceedings of the 25th ACM SIGKDD international conference on knowledge discovery \& data mining},
  pages={2828--2837},
  year={2019}
}

@inproceedings{han2022learning,
  title={Learning sparse latent graph representations for anomaly detection in multivariate time series},
  author={Han, Siho and Woo, Simon S},
  booktitle={Proceedings of the 28th ACM SIGKDD Conference on Knowledge Discovery and Data Mining},
  pages={2977--2986},
  year={2022}
}

@inproceedings{deng2021graph,
  title={Graph neural network-based anomaly detection in multivariate time series},
  author={Deng, Ailin and Hooi, Bryan},
  booktitle={Proceedings of the AAAI conference on artificial intelligence},
  pages={4027--4035},
  year={2021}
}

@article{khraisat2019survey,
  title={Survey of intrusion detection systems: techniques, datasets and challenges},
  author={Khraisat, Ansam and Gondal, Iqbal and Vamplew, Peter and Kamruzzaman, Joarder},
  journal={Cybersecurity},
  volume={2},
  number={1},
  pages={1--22},
  year={2019},
  publisher={Springer}
}

@inproceedings{hadvziosmanovic2014through,
  title={Through the eye of the PLC: semantic security monitoring for industrial processes},
  author={Had{\v{z}}iosmanovi{\'c}, Dina and Sommer, Robin and Zambon, Emmanuele and Hartel, Pieter H},
  booktitle={Proceedings of the 30th Annual Computer Security Applications Conference},
  pages={126--135},
  year={2014}
}

@article{kushner2013real,
  title={The real story of stuxnet},
  author={Kushner, David},
  journal={IEEE Spectrum},
  volume={50},
  number={3},
  pages={48--53},
  year={2013},
  publisher={Institute of Electrical and Electronics Engineers (IEEE)}
}

@article{stellios2018survey,
  title={A survey of iot-enabled cyberattacks: Assessing attack paths to critical infrastructures and services},
  author={Stellios, Ioannis and Kotzanikolaou, Panayiotis and Psarakis, Mihalis and Alcaraz, Cristina and Lopez, Javier},
  journal={IEEE Communications Surveys \& Tutorials},
  volume={20},
  number={4},
  pages={3453--3495},
  year={2018},
  publisher={IEEE}
}

@article{wang2023scientific,
  title={Scientific discovery in the age of artificial intelligence},
  author={Wang, Hanchen and Fu, Tianfan and Du, Yuanqi and Gao, Wenhao and Huang, Kexin and Liu, Ziming and Chandak, Payal and Liu, Shengchao and Van Katwyk, Peter and Deac, Andreea and others},
  journal={Nature},
  volume={620},
  number={7972},
  pages={47--60},
  year={2023},
  publisher={Nature Publishing Group UK London}
}

@article{newhart2019data,
  title={Data-driven performance analyses of wastewater treatment plants: A review},
  author={Newhart, Kathryn B and Holloway, Ryan W and Hering, Amanda S and Cath, Tzahi Y},
  journal={Water research},
  volume={157},
  pages={498--513},
  year={2019},
  publisher={Elsevier}
}

@article{estimator1999fast,
  title={A Fast Algorithm for the Minimum Covariance},
  author={EStimator, Determinant},
  journal={Technometrics},
  volume={41},
  number={3},
  pages={212},
  year={1999}
}

@inproceedings{aoudi2018truth,
  title={Truth will out: Departure-based process-level detection of stealthy attacks on control systems},
  author={Aoudi, Wissam and Iturbe, Mikel and Almgren, Magnus},
  booktitle={Proceedings of the 2018 ACM SIGSAC Conference on Computer and Communications Security},
  pages={817--831},
  year={2018}
}

@article{spenneberg2016plc,
  title={Plc-blaster: A worm living solely in the plc},
  author={Spenneberg, Ralf and Br{\"u}ggemann, Maik and Schwartke, Hendrik},
  journal={Black Hat Asia},
  volume={16},
  pages={1--16},
  year={2016}
}

@inproceedings{ribeiro2016should,
  title={" Why should i trust you?" Explaining the predictions of any classifier},
  author={Ribeiro, Marco Tulio and Singh, Sameer and Guestrin, Carlos},
  booktitle={Proceedings of the 22nd ACM SIGKDD international conference on knowledge discovery and data mining},
  pages={1135--1144},
  year={2016}
}

@article{lundberg2017unified,
  title={A unified approach to interpreting model predictions},
  author={Lundberg, Scott M and Lee, Su-In},
  journal={Advances in neural information processing systems},
  volume={30},
  year={2017}
}

@article{apley2020visualizing,
  title={Visualizing the effects of predictor variables in black box supervised learning models},
  author={Apley, Daniel W and Zhu, Jingyu},
  journal={Journal of the Royal Statistical Society Series B: Statistical Methodology},
  volume={82},
  number={4},
  pages={1059--1086},
  year={2020},
  publisher={Oxford University Press}
}

@inproceedings{han2021deepaid,
  title={Deepaid: Interpreting and improving deep learning-based anomaly detection in security applications},
  author={Han, Dongqi and Wang, Zhiliang and Chen, Wenqi and Zhong, Ying and Wang, Su and Zhang, Han and Yang, Jiahai and Shi, Xingang and Yin, Xia},
  booktitle={Proceedings of the 2021 ACM SIGSAC Conference on Computer and Communications Security},
  pages={3197--3217},
  year={2021}
}

@inproceedings{shoukry2015pycra,
  title={Pycra: Physical challenge-response authentication for active sensors under spoofing attacks},
  author={Shoukry, Yasser and Martin, Paul and Yona, Yair and Diggavi, Suhas and Srivastava, Mani},
  booktitle={Proceedings of the 22nd ACM SIGSAC Conference on Computer and Communications Security},
  pages={1004--1015},
  year={2015}
}

@inproceedings{lin2023dnattest,
  title={DNAttest: Digital-twin-based Non-intrusive Attestation under Transient Uncertainty},
  author={Lin, Wei and Tan, Heng Chuan and Chen, Binbin and Zhang, Fan},
  booktitle={2023 53rd Annual IEEE/IFIP International Conference on Dependable Systems and Networks (DSN)},
  pages={376--388},
  year={2023},
  organization={IEEE}
}

@inproceedings{bratus2008toctou,
  title={TOCTOU, traps, and trusted computing},
  author={Bratus, Sergey and D’Cunha, Nihal and Sparks, Evan and Smith, Sean W},
  booktitle={International Conference on Trusted Computing},
  pages={14--32},
  year={2008},
  organization={Springer}
}

@article{zhu2024honeyjudge,
  title={HoneyJudge: A PLC Honeypot Identification Framework Based on Device Memory Testing},
  author={Zhu, Hengye and Liu, Mengxiang and Chen, Binbin and Che, Xin and Cheng, Peng and Deng, Ruilong},
  journal={IEEE Transactions on Information Forensics and Security},
  year={2024},
  publisher={IEEE}
}

@article{bako2011identification,
  title={Identification of switched linear systems via sparse optimization},
  author={Bako, Laurent},
  journal={Automatica},
  volume={47},
  number={4},
  pages={668--677},
  year={2011},
  publisher={Elsevier}
}

@article{pearson2002outliers,
  title={Outliers in process modeling and identification},
  author={Pearson, Ronald K},
  journal={IEEE Transactions on control systems technology},
  volume={10},
  number={1},
  pages={55--63},
  year={2002},
  publisher={IEEE}
}

@inproceedings{python-control2021,
  title={The Python Control Systems Library (python-control)},
  author={Fuller, Sawyer and Greiner, Ben and Moore, Jason and
          Murray, Richard and van Paassen, Ren{\'e} and Yorke, Rory},
  booktitle={60th IEEE Conference on Decision and Control (CDC)},
  pages={4875--4881},
  year={2021},
  organization={IEEE}
}

@article{langner2011stuxnet,
  title={Stuxnet: Dissecting a cyberwarfare weapon},
  author={Langner, Ralph},
  journal={IEEE Security \& Privacy},
  volume={9},
  number={3},
  pages={49--51},
  year={2011},
  publisher={IEEE}
}

@article{di2018triton,
  title={TRITON: The first ICS cyber attack on safety instrument systems},
  author={Di Pinto, Alessandro and Dragoni, Younes and Carcano, Andrea},
  journal={Proc. Black Hat USA},
  volume={2018},
  pages={1--26},
  year={2018}
}

@article{geng2023control,
  title={Control logic attack detection and forensics through reverse-engineering and verifying PLC control applications},
  author={Geng, Yangyang and Che, Xin and Ma, Rongkuan and Wei, Qiang and Wang, Mufeng and Chen, Yuqi},
  journal={IEEE Internet of Things Journal},
  volume={11},
  number={5},
  pages={8386--8400},
  year={2023},
  publisher={IEEE}
}

@article{cranmer2023interpretable,
  title={Interpretable machine learning for science with PySR and SymbolicRegression. jl},
  author={Cranmer, Miles},
  journal={arXiv preprint arXiv:2305.01582},
  year={2023}
}

@inproceedings{burlacu2020operon,
  title={Operon C++ an efficient genetic programming framework for symbolic regression},
  author={Burlacu, Bogdan and Kronberger, Gabriel and Kommenda, Michael},
  booktitle={Proceedings of the 2020 genetic and evolutionary computation conference companion},
  pages={1562--1570},
  year={2020}
}

@article{qasim2020control,
  title={Control logic forensics framework using built-in decompiler of engineering software in industrial control systems},
  author={Qasim, Syed Ali and Smith, Jared M and Ahmed, Irfan},
  journal={Forensic Science International: Digital Investigation},
  volume={33},
  pages={301013},
  year={2020},
  publisher={Elsevier}
}

@inproceedings{kalle2019clik,
  title={CLIK on PLCs! Attacking control logic with decompilation and virtual PLC},
  author={Kalle, Sushma and Ameen, Nehal and Yoo, Hyunguk and Ahmed, Irfan},
  booktitle={Binary Analysis Research (BAR) Workshop, Network and Distributed System Security Symposium (NDSS)},
  pages={1--12},
  year={2019}
}

@article{sang2024control,
  title={From Control Application to Control Logic: PLC Decompile Framework for Industrial Control System},
  author={Sang, Chao and Wu, Jun and Li, Jianhua and Guizani, Mohsen},
  journal={IEEE Transactions on Information Forensics and Security},
  volume={19},
  pages={8685--8700},
  year={2024},
  publisher={IEEE}
}
%



\begin{IEEEbiography}[{\includegraphics[width=1in,height=1.25in,clip,keepaspectratio]{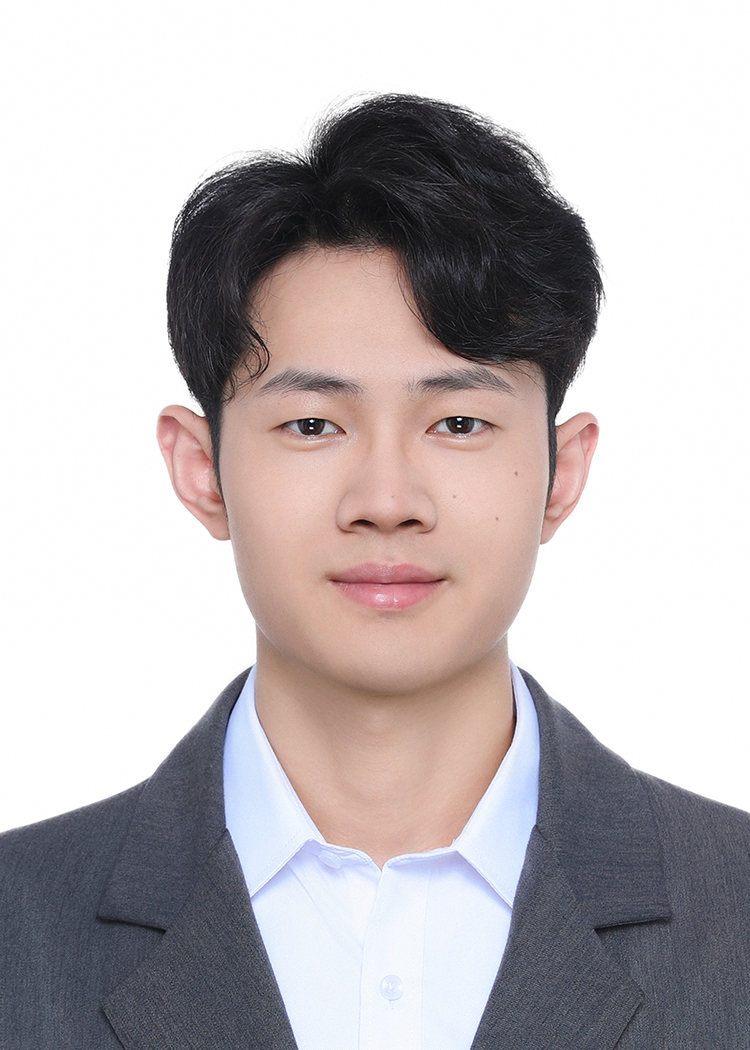}}]{Hao Zhou}
received the B.Sc. degree in communication engineering from Wuhan University of Technology in 2019 and the Ph.D. degree in information and communication engineering from Beijing University of Posts and Telecommunications in 2025. He was a visiting Ph.D. student at the Singapore University of Technology and Design from 2023 to 2024. His current research interests include AI-native radio access networks and time series modeling for forecasting and anomaly detection.
\end{IEEEbiography}
\begin{IEEEbiography}
[{\includegraphics[width=1in,height=1.25in,clip,keepaspectratio]{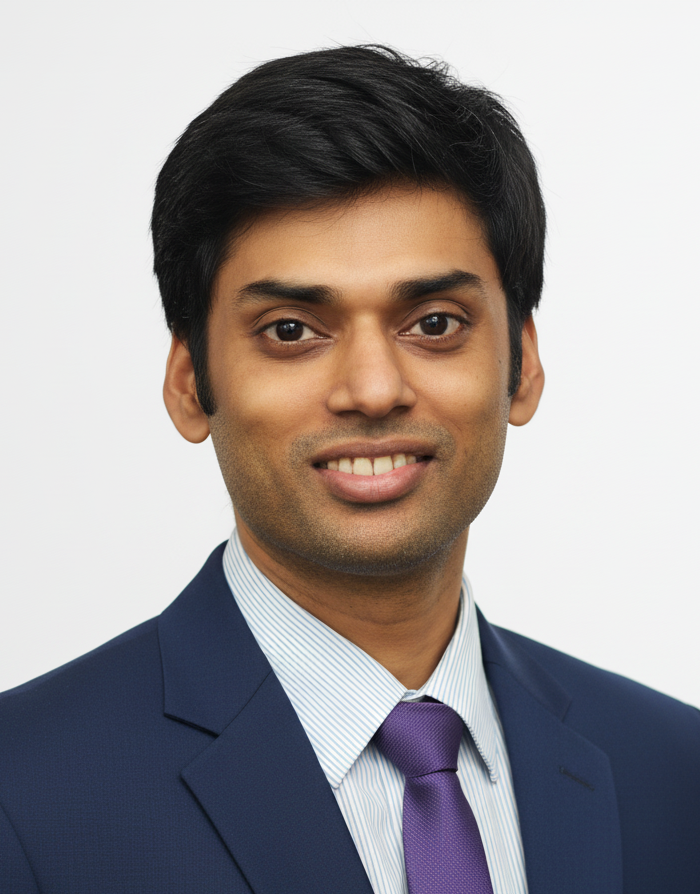}}]{Suman Sourav}
 is an Assistant Professor at Aalborg University, Denmark. He earned his Ph.D. from the National University of Singapore, where he also received the Research Achievement Award. He was previously a Research Fellow at the Singapore University of Technology and Design and at Illinois Advanced Research Centre, Singapore. His research interest lies in the interplay of theory and application, considering both theoretical as well as practically implemented algorithmic aspects of real-world systems, including topics such as satellite networks, edge-cloud systems, distributed computing, and cyber-physical systems.
\end{IEEEbiography}
\begin{IEEEbiography}
[{\includegraphics[width=1in,height=1.25in,clip,keepaspectratio]{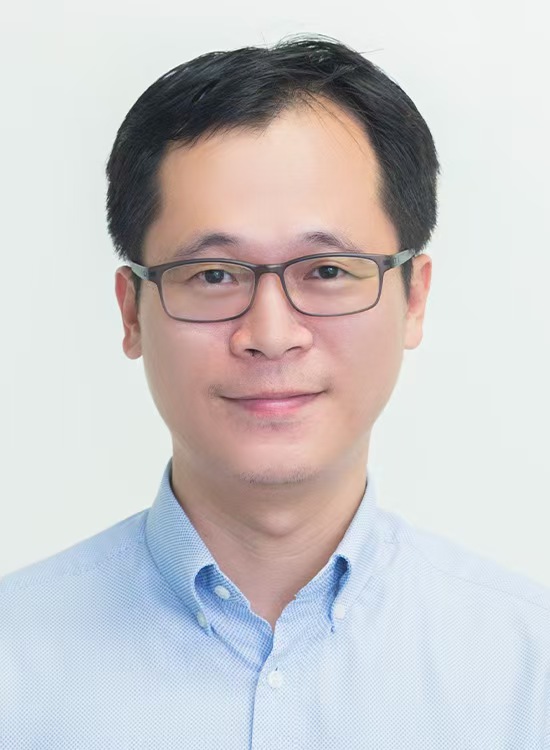}}]{Binbin Chen}
(M'11) received the B.Sc. degree in computer science from Peking University in 2003 and the Ph.D. degree in computer science from the National University of Singapore in 2010. Since July 2019, he has been an Associate Professor in the Information Systems Technology and Design (ISTD) pillar, Singapore University of Technology and Design (SUTD). His current research interests include wireless networks, cyber-physical systems, and cyber security for critical infrastructures.
\end{IEEEbiography}
\begin{IEEEbiography}
[{\includegraphics[width=1in,height=1.25in,clip,keepaspectratio]{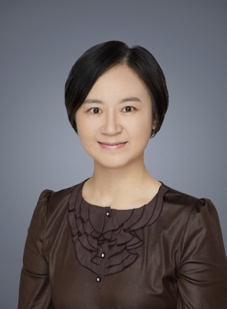}}]{Ke Yu}
received the B.S. degree in computer science and the Ph.D. degree in signal and information processing from Beijing University of Posts and Telecommunications (BUPT), China, in 2000 and 2005, respectively. She worked as a visiting researcher at the University of Agder, Norway in 2011. From 2015 to 2016, she has been a visiting scholar in the University of Illinois at Chicago, USA. She is currently a professor at the School of Artificial Intelligence, BUPT. Her research interests include communication network theory, network data mining and anomaly detection, Internet of Agents.
\end{IEEEbiography}
\vfill

\balance
\newpage

{\appendices
\section*{Appendix I: Detailed Algorithms of SRLR}
\subsection{Multi-mode Logic Recovery Algorithm}
\label{app:multi_mode_alg}
\begin{algorithm}[htbp]
    \renewcommand{\algorithmicrequire}{\textbf{Input:}}
	\renewcommand{\algorithmicensure}{\textbf{Output:}}
	\caption{\small Alternative optimization strategy for multi-mode logic recovery}
    \label{algorithm_aos}
    \begin{algorithmic}[1]
        \REQUIRE  Dataset $\mathcal{D}=\{(x_t,y_t)|t\in \mathcal{T}=\{1,2,...,T\}\}$ and $max\_iter$
	    \ENSURE The total number of modes $K$, index of each mode $ind_k$ and dynamic equation of each mode $f_k(.)$
        
        \STATE Initialization: $K\leftarrow 1, \{\gamma_{t}\}_{t\in \mathcal{W}} \leftarrow 1, \{\gamma_{t}\}_{t\in \mathcal{T} \diagdown \mathcal{W}} \leftarrow 0$
        
        \REPEAT 
            \IF {$K = 1$}
                \STATE $\mathcal{T}_K \leftarrow \mathcal{T}$
            \ELSE
                \STATE $\mathcal{T}_K \leftarrow \mathcal{T}\diagdown \bigcup_{k=1}^{K-1} \mathcal{T}_k$
            \ENDIF
            \FORALL {$i=1,2,\cdots,max\_iter$}
                \STATE /* Update parameters of DSR */ \\
                $\zeta^{(i+1)}_K=\mathop{\arg\min}\limits_{\zeta_K} J_{risk}(\zeta^{(i)}_K;\varepsilon)$

                \STATE Calculate current loss $\{l_t\}_{t \in \mathcal{T}_K}$

                \STATE /* Update $\gamma$ */\\
                $\gamma^{(i+1)}_K = \mathop{\arg\min}\limits_{\gamma_{K} \in \{0,1\}^{|\mathcal{T}_K|}} \quad \sum_{t\in \mathcal{T}_K}\gamma^{(i)}_{K,t}l^{(i)}_{t} - \lambda \gamma^{(i)}_{K,t}$ 
            \ENDFOR
            \STATE $\{ind_K\} \leftarrow \{t | \gamma_{k,t}=1, t\in \mathcal{T}_K\}$

            \IF {$\mathcal{T} \diagdown \bigcup_{k=1}^{K}\{ind_k\} = \varnothing$}
                \STATE \textbf{return} ${K}, f_1,..., f_{K}, ind_1,..., ind_{K}$.
            \ELSE
                \STATE $K \leftarrow K+1$
            \ENDIF
        \UNTIL all the points are identified.
    \end{algorithmic}
\end{algorithm}
The details of multi-mode logic recovery are illustrated in Algorithm~\ref{algorithm_aos}, where the input consists of the training dataset and the maximum allowed iteration for each mode identification ($max\_itr$). The algorithm outputs the total number of modes, the indexes of mode changes, and the mode expressions. Initialization (Row 1)  involves assigning the first $w$ points ($\mathcal{W}=\{1,2,...,w\}$) to mode 1, which serves as prior information (follows from Assumption 1). Rows 3-7 check the unmarked dataset, while Rows 8-12 represent the identification process. Row 9 optimizes the DSR model with the fixed mode membership $\gamma$, followed by Row 10, which computes the prediction loss for each point. Row 11 updates the mode membership $\gamma$ with a fixed mode expression. Row 13 indicates the identification index for the $K$-th mode. This process is repeated until all points are successfully identified. An adjustment method is proposed to enhance the precision of index identification. It is assumed that the points generated by each mode exhibits temporal continuity. We utilize a sliding window approach with a size and sliding step of 50 to carefully adjust identified points in high-level noisy data. If the proportion of identified points exceeds 80\% of all points within the window, we classify all points within that window as identified; otherwise, they are classified as unidentified. This large sliding step was chosen not only to minimize misclassified points in switching time step of the subsystem, but also to enhance computational efficiency.

\subsection{Outlier-aware Training}
\label{app:outlier_aware_train}
The outlier-aware training algorithm is shown in Algorithm~\ref{algorithm_outlier}, where $N$ is the batch size.
\begin{algorithm}[htbp]
    \renewcommand{\algorithmicrequire}{\textbf{Input:}}
	\renewcommand{\algorithmicensure}{\textbf{Output:}}
	\caption{\small Outlier-aware method for dynamic system identification}
    \label{algorithm_outlier}
    \begin{algorithmic}[2]
        \REQUIRE  Dataset $\mathcal{D}=\{(x_t,y_t)|t\in \mathcal{T}=\{1,2,...,T\}\}$ and $max\_itr$
	    \ENSURE The best fitting expression $\tau^*$
        \STATE Initialization: The parameters of RNN and the initial distribution of $p(.|\zeta)$
        \REPEAT 
            \STATE /* Sample a batch of expressions using RNN */ \\
            $\mathcal{O}=\{\tau^{(i)}\}_{i=1}^{N}$
            \STATE /* Initialize constant items */ \\
            $\mathcal{O} \leftarrow \{InitializeConstant(\tau^{(i)})\}_{i=1}^{N}$
            \STATE /* Compute loss for each training point */ \\
            $\mathcal{L}^{(i)}\leftarrow \{l_t^{(i)}=|y_t-\hat{y}^{(i)}_t| \ \vert \ t\in \mathcal{T}\}, i=1,2,...,N$
            \STATE /* Remove training points with higher losses, $th$ is the (1-$\alpha$)-quartile of losses */ \\
            $\mathcal{T}^{(i)}\leftarrow \{t \ \vert \ l_t^{(i)}\leq th, t\in \mathcal{T}\}, i=1,2,...,N$
            \STATE /* Optimize constant items */ \\
            $\mathcal{O} \leftarrow \{OptimizeConstant(\tau^{(i)}, R, \mathcal{T}^{(i)})\}_{i=1}^{N}$
            \STATE /* Update DSR parameters */ \\
            $\zeta \leftarrow \mathop{\arg\min}\limits_{\zeta} J_{risk}(\zeta;\varepsilon, \mathcal{T}^{(i)}), i=1,2,...,N$
            \STATE /* Evaluate current best expression */ \\
            if $\max R > R(\tau^*)$ then $\tau^* \leftarrow \tau^{\arg\max R}$
        \UNTIL {evaluation error smaller than the threshold or maximum iterations $max\_itr$ is reached.}
    \end{algorithmic}
\end{algorithm}

\subsection{Mode Switch Detection Procedure}

The mode switch detection process consists of three main steps: identifying boundary points, filtering these points, and identifying the mode-switching logic. Each step is described in detail below.

Step 1: Identifying Boundary Points. We begin by using the proposed SRLR to determine the behavioral logic associated with all data points. Mode-switching points (those at which the system's behavior changes) are then identified and labeled as boundary points.

Step 2: Filtering Boundary Points. The boundary points collected in the first step vary in their proximity to the true switching boundaries. To refine the selection, we apply a filtering process using the Elliptic Envelope method, chosen for its balance of efficiency and accuracy in identifying outliers. This involves estimating the center of the point cluster using the Minimum Covariance Determinant, computing the Mahalanobis distance of each point from this center, ranking the points based on these distances, and discarding those that exceed a threshold (empirically set to 0.4 in this study).

Step 3: Identifying Mode-Switch Logic. In the final step, we reconfigure SRLR for single-mode, time-domain logic identification. The filtered boundary points are then input into SRLR, which learns a logic function $f(x)$, where $f(x)=0$ defines the switching condition. The system mode at any point is determined by the sign of $f(x)$: values where $f(x)>0$ or $f(x) \leq 0$ correspond to different operational modes.

\section*{Appendix II: Additional Experiments}
\subsection{Identification results for EWMA}
\label{app:ewma}
We compare our model with widely-used system identification methods: Nonlinear Autoregressive Models with Moving Average and Exogenous Input (NARMAX)~\footnote{https://github.com/wilsonrljr/sysidentpy} and Genetic Programming (GP)~\footnote{https://github.com/trevorstephens/gplearn}, under different percentages of injected outliers. NARMAX is tailored for a broad class of nonlinear dynamic systems, comprising two steps: first, identifying the system structure, such as polynomial terms, and second, combining input-output lags to estimate the structure parameters. GP is a symbolic regression method inspired by evolutionary principles, exploring the symbolic space to select the fittest individuals capable of explaining the input-output data relationship.

The detailed results are in Table~\ref{table_comparison_ewma}. NARMAX, GP, and SRLR can accurately identify EWMA systems with high Best Fit Ratio (BFR) and provide precise predictive expressions. However, when outliers are introduced into the training data, both NARMAX and GP fail to identify the correct expression. For a contamination fraction of 0.01, GP achieves a higher BFR than NARMAX. However, upon plotting the dynamics of the predictive expressions generated by the baselines and our proposed model on the test data shown in Figure~\ref{fig_ewma_compare}, it becomes evident that the curve generated by GP deviates more from the ground truth compared to the NARMAX model. Our proposed model not only generates accurate expressions with a higher BFR but also effectively fits the true curve on the test set.

\begin{figure}[t]
    \centering
    \includegraphics[scale=0.5]{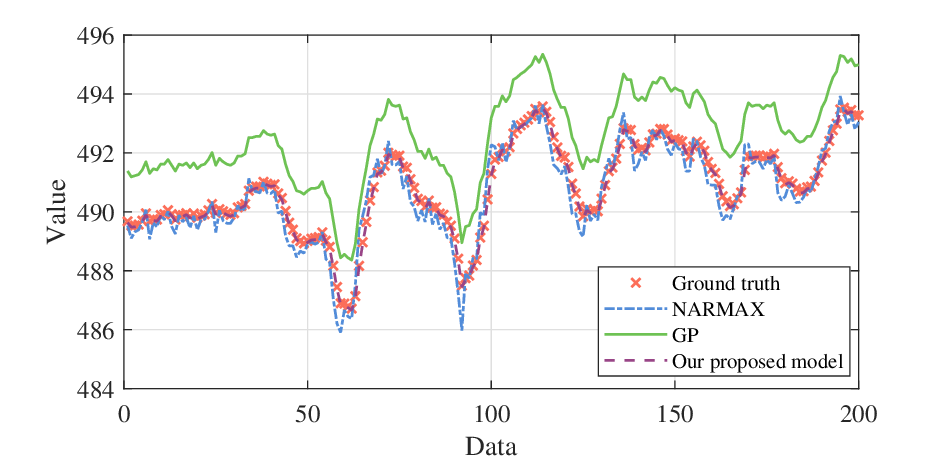}
    \caption{The comparison results of the simulated ewma data.}
    \label{fig_ewma_compare}
\end{figure}

\begin{table*}[htbp]
    \begin{center}
    \caption{The comparison of logic recovery for EWMA. Ground truth: $y(t)=0.8x(t)+0.16x(t-1)+0.032x(t-2)$.}
    \label{table_comparison_ewma}
    \resizebox{\textwidth}{!}{
    \begin{tabular}{ c|cc|cc|cc }
        \toprule
        Outlier & \multicolumn{2}{c}{NARMAX} & \multicolumn{2}{c}{Genetic Programming} & \multicolumn{2}{c}{SRLR(Our model)}\\
        fraction & Prediction & BFR & Prediction & BFR & Prediction & BFR\\
        \midrule
        0.0 & $y(t)=0.8x(t)+0.16x(t-1)+0.032x(t-2)+5.8e^{-11}y(t-1)$ & 1.0 & $y(t)=0.995x(t)-0.968$ & 0.989 & $y(t)=0.7958x(t)+0.1653x(t-1)+0.0308x(t-2)$ & 0.9959\\
        0.01 & $y(t)=1.7329x(t-1)-0.7470y(t-1)$ & 0.9728 & $y(t)=0.995x(t)-0.979$ & 0.9883 & $y(t)=0.7985x(t)+0.1628x(t-1)+0.0305x(t-2)$ & 0.9945\\
        0.02 & $y(t)=1.8563x(t)-0.8715y(t-1)$ & 0.9690 & $y(t)=0.995x(t)-0.979$ & 0.9883 & $y(t)=0.7987x(t)+0.1635x(t-1)+0.0297x(t-2)$ & 0.9959\\
        0.03 & $y(t)=1.8916x(t)-0.9070y(t-1)$ & 0.9681 & $y(t)=0.995x(t)-0.979$ & 0.9883 & $y(t)=0.7961x(t)+0.1665x(t-1)+0.0292x(t-2)$ & 0.9945\\
        0.04 & $y(t)=1.9181x(t)-0.9338y(t-1)$ & 0.9672 & $y(t)=x(t)-x(t)/(0.446x(t-2))$ & 0.9904 & $y(t)=0.7986x(t)+0.1637x(t-1)+0.0296x(t-2)$ & 0.9959\\
        0.05 & $y(t)=1.9286x(t)-0.9449y(t-1)$ & 0.9643 & $y(t)=x(t)-2.2176$ & 0.9900 & $y(t)=0.7986x(t)+0.1620x(t-1)+0.0313x(t-2)$ & 0.9959\\
        0.06 & $y(t)=1.9393x(t)-0.9554y(t-1)$ & 0.9658 & $y(t)=x(t)-2.2590$ & 0.9901 & $y(t)=0.8004x(t)+0.1588x(t-1)+0.0327x(t-2)$ & 0.9960\\
        \bottomrule
    \end{tabular}
    }
    \end{center}
\end{table*}

{We also evaluate recovery performance by comparing our approach with baseline models that incorporate data cleaning techniques. Hampel filtering, a commonly used method, mitigates outliers by adjusting them based on median absolute deviation. As shown in Table~\ref{table_comparison_ewma_hampelfilter}, applying this filtering improves the accuracy of the NARMAX model in predicting the EWMA expression. However, its performance declines as the proportion of outliers increases. Genetic Programming demonstrates greater stability in the structure of its predictive expressions, but it struggles to accurately identify the true underlying format.}
\begin{table*}[htbp]
    \begin{center}
    \caption{Comparison of logic recovery for EWMA with data cleaned using the Hampel filter.}
    \label{table_comparison_ewma_hampelfilter}
    \begin{threeparttable}
    \begin{tabular}{ l|cc|cc }
        \toprule
        Outlier & \multicolumn{2}{c|}{NARMAX} & \multicolumn{2}{c}{Genetic Programming}\\
        fraction & Prediction & BFR & Prediction & BFR\\
        \midrule
        0.0 & $y(t)=0.7929x(t)+0.2006y(t-1)$ & 0.9978 & $y(t)=0.993x(t)-0.3744$ & 0.9913\\
        0.01 & $y(t)=0.8123x(t)+0.1810y(t-1)$ & 0.9970 & $y(t)=0.993x(t)-0.3744$ & 0.9913\\
        0.02 & $y(t)=0.8247x(t)+0.1685y(t-1)$ & 0.9970 & $y(t)=0.993x(t)-0.3744$ & 0.9913\\
        0.03 & $y(t)=0.8301x(t)+0.1631y(t-1)$ & 0.9978 & $y(t)=x(t)-2.2542$ & 0.9902\\
        0.04 & $y(t)=0.8417x(t)+0.1514y(t-1)$ & 0.9975 & $y(t)=0.995x(t)-0.982$ & 0.9881\\
        0.05 & $y(t)=0.8522x(t)+0.1408y(t-1)$ & 0.9971 & $y(t)=0.995x(t)-0.982$ & 0.9881\\
        0.06 & $y(t)=0.9122x(t)+0.0803y(t-1)$ & 0.9958 & $y(t)=0.995x(t)-0.997$ & 0.9872\\
        \bottomrule
    \end{tabular}
    \begin{tablenotes}
    \footnotesize
    \item[1] Ground truth expression: $y(t)=0.8x(t)+0.16x(t-1)+0.032x(t-2)$.
    \item[2] Predictive expression by SRLR under outlier fraction of 0.06: $y(t)=0.8004x(t)+0.1588x(t-1)+0.0327x(t-2)$.
    \end{tablenotes}
    \end{threeparttable}
    \end{center}
\end{table*}

{Data cleaning techniques are commonly used in system identification to eliminate extreme values or smooth data, helping to prevent the optimization process from being misled. However, most traditional cleaning methods—such as distance-based or clustering-based approaches—do not account for the performance of the system identification itself. As a result, they may incorrectly label valid data points as outliers. Our proposed outlier-aware training method addresses this limitation by jointly optimizing for both system identification and outlier removal. This integrated approach ensures that outlier detection supports the ultimate goal of accurate system modeling.}

{Figure~\ref{fig:outlier_removal_alg} presents a comparison between our method and two widely used data cleaning techniques~\cite{pearson2002outliers}: MT Cleaner and the Hampel filter. MT Cleaner combines a regression model with a Kalman filter, using prediction errors to assess outlier severity. The Hampel filter relies on median absolute deviation, while our method, SRLR, uses the reciprocal of the learned reward to quantify outlier likelihood. As illustrated, both MT Cleaner and the Hampel filter effectively identify global and collective outliers but struggle with contextual outliers—those that appear anomalous in local segments yet fall within the normal range when viewed globally. In contrast, SRLR demonstrates higher sensitivity to all types of outliers, owing to its joint optimization strategy.}

\begin{figure*}[htbp]
    \centering
    \includegraphics[width=1.\linewidth]{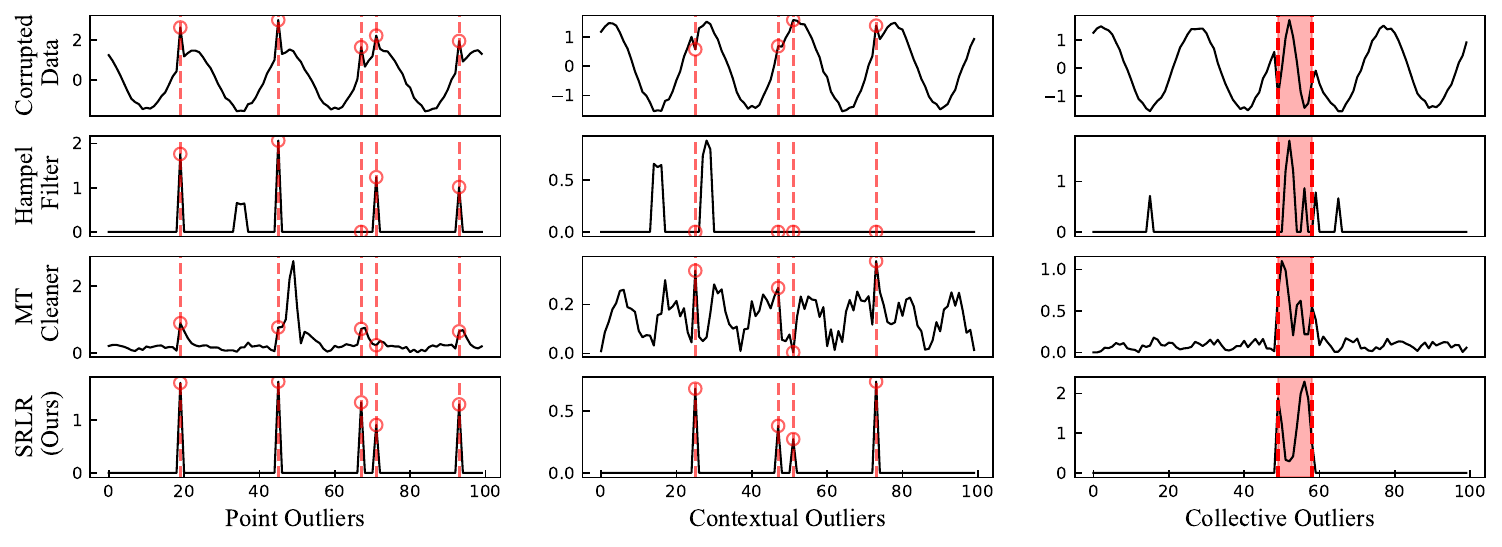}
    \caption{Comparison of data cleaning algorithms, including the Hampel Filter, MT Cleaner, and the proposed SRLR’s outlier-aware filter. Red regions indicate anomalies. The first row shows the raw data, followed by anomaly scores from each algorithm.}
    \label{fig:outlier_removal_alg}
\end{figure*}

When considering contaminated data with outliers, as depicted in Figure~\ref{fig_ewma_residual}, the cumulative distribution probability of residual errors for the candidate expression is presented for cases involving the injection of 1\% and 2\% outliers. The results reveal that more than 90\% of the training data exhibit lower residual errors. This suggests that training data with higher residual errors are likely to be outliers, constituting only a very small percentage of the dataset. Therefore, the proposed outlier-aware filter method effectively identifies and reduces the impact of outlier training data on system identification results.


\begin{figure}[htbp]
    \centering
    \includegraphics[scale=0.5]{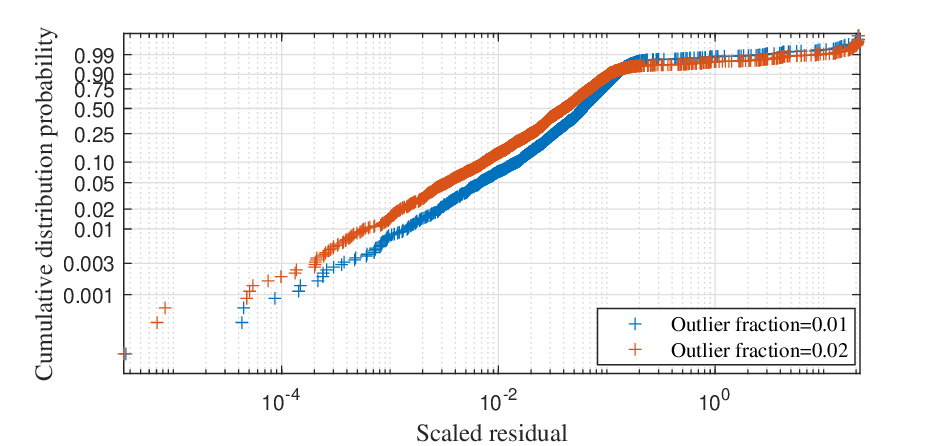}
    \caption{The residual distribution for each point of the candidate expressions.}
    \label{fig_ewma_residual}
\end{figure}

\subsection{Identification Results in Load Frequency Control}
\label{app:freq_control}
The Load Frequency Control (LFC) System employs several mathematical models, including generator, load, prime mover and governor model. The generator can be characterized by the swing equation. The load comprises frequency-independent and frequency-sensitive devices, allowing us to model the load as a combination of these two types of devices. Sudden load changes can cause deviations in frequency. The governor model is designed to adjust mechanical power, steering the system towards a new steady-state. The formulations are expressed as follows:
\begin{equation}
    \label{eq5}
    \begin{aligned}
    \frac{\triangle \Omega(s)}{\triangle P_M(s)-\triangle P_L(s)} &= \frac{1}{2Hs+D},\ \ 
    \frac{\triangle P_M(s)}{\triangle P_V(s)} = \frac{1}{1+\tau_Ts}\\
    \frac{\triangle P_V(s)}{\triangle P_G(s)} &= \frac{1}{1+\tau_Gs},\ \ 
    \frac{\triangle P_G(s)}{\triangle \Omega(s)} = -\frac{1}{R}
    \end{aligned}
\end{equation}
where the equations are in the S-domain. The first is the equation for the load and inertia, $\triangle \Omega(s), \triangle P_M(s), \triangle P_L(s)$ is the change in frequency, mechanical power, and load power, respectively, and $H, D$ is the load angle and inertia constant, respectively. The second is the equation for prime mover(turbine), $\triangle P_V(s)$ is the changes in steam valve position, and $\tau_T$ is the turbine time constant. The third is the equation for the governor, $\triangle P_G(s)$ is the difference between the set power and the feedback power, and $\tau_G$ is the governor time constant. The fourth is the equation for the feedback gain and $R$ is the governor speed regulation.

\begin{figure}[t]
    \centering
    \includegraphics[scale=0.35]{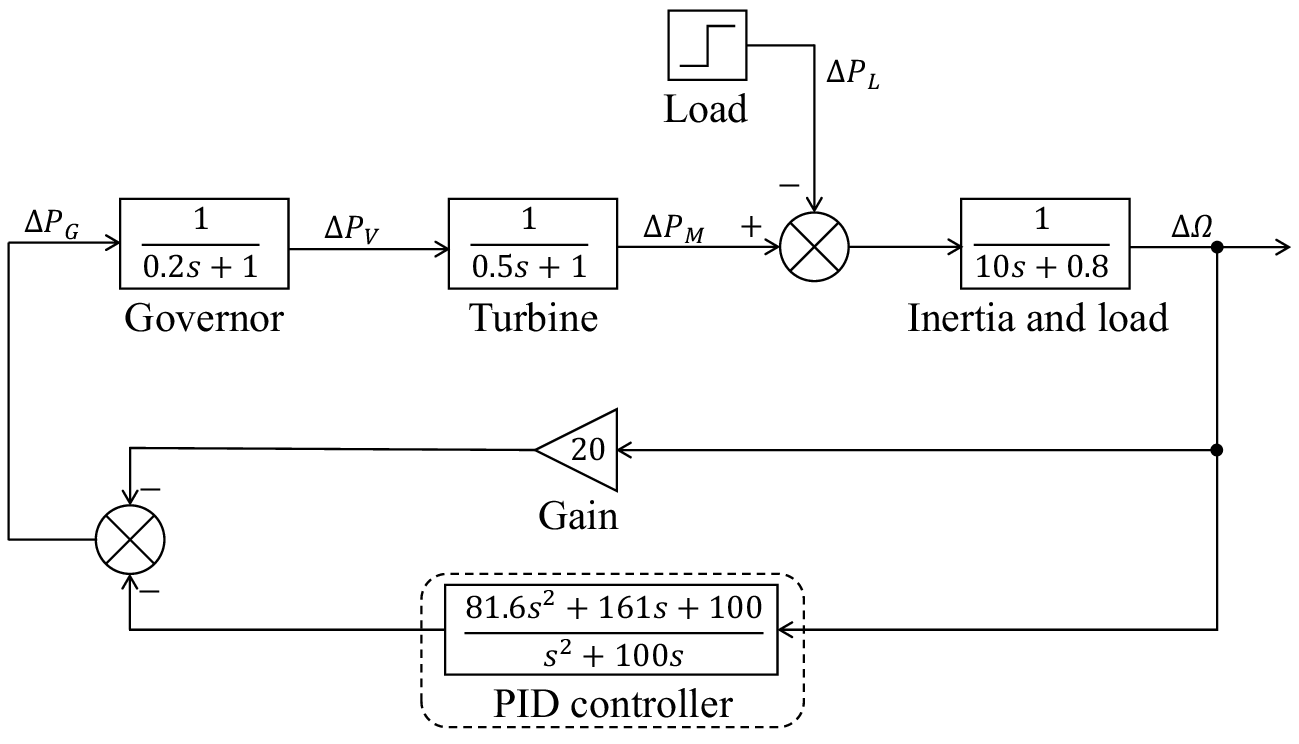}
    \caption{Framework of the load frequency control system, comprising five components: governor, turbine, generator, gain, and controller.}
    \label{fig_frawmework_freqPID}
\end{figure}
Standard frequency in load applications remains constant, such as 60Hz or 50Hz. However, increased loads may convert more kinetic energy into electrical energy, impacting rotor angles of generators and subsequently leading to changes in system frequency. Consequently, control strategies within power systems become crucial to achieve a steady-state system. 
Here, we analyze the capabilities of SRLR in identifying power control systems and the associated power control logic.

We present a simulation of a load frequency controlled system as shown in Figure~\ref{fig_frawmework_freqPID}, whose parameters are configured according to the reference~\cite{saadat1999power}. 
When given a constant input, i.e. $\triangle P_L=0.2$ per unit, the frequency tending towards a new steady state after a deviation interval.
The identification results, as presented in Table~\ref{table_identified_system}, demonstrate that the proposed framework accurately identifies the system. Moreover, the complexity regularization method effectively aids in discovering simpler structures. 

The droop alone fails to control the frequency changes to zero. Therefore, an additional controller needs to be introduced to control the frequency change towards zero. We incorporate a Proportional-Integral-Derivative (PID) controller into the feedback loop, as depicted in the dashed box in Figure~\ref{fig_frawmework_freqPID}. The PID controller leverages proportional, integral, and derivative influences for feedback, optimizing control over frequency. The frequency will tend toward the reference frequency (60Hz) when employing the PID controller in the feedback loop. Table~\ref{table_identified_system} records the identification results, revealing that the proposed framework can accurately identify the ground truth PID controller.
\begin{table*}[htbp]
    \begin{center}
    \caption{The identification results of SRLR in frequency control system with/without AIC regularization.}
    \label{table_identified_system}
    \resizebox{\textwidth}{!}{
    \begin{tabular}{ c|c|ccc|cccc }
        \hline
         & Ground truth mode & Identified mode(without AIC) &Complexity&BFR & Identified mode(with AIC) & simplified mode(with AIC) &Complexity&BFR \\
        \hline
        Governor&$\frac{1}{0.2s+1}$&$\frac{5.001s-1.327e-5}{s^2+5.001s}$& 20 & 0.9999 &$\frac{5.001}{s+5.001}$ & $\frac{1}{0.1999s+1}$& 13 & 0.9999\\
        \hline
        Turbine & $\frac{1}{0.5s+1}$ & $\frac{2s-2.108e-6}{s^2+2s}$ & 20 & 0.9999 & $\frac{2}{s+2}$ &$\frac{1}{0.5s+1}$ & 13 & 1\\
        \hline
        Rotating mass and load & $\frac{1}{10s+0.8}$ & $\frac{0.1s+3.797e-6}{s^2+0.08s}$ & 20 & 0.9998 & $\frac{0.1}{s+0.08}$ & $\frac{1}{10s+0.8}$ & 17 & 1\\
        \hline
        Gain & $-20$ & $-20$ & 19 & 1 & $-20$ & $-20$ & 18 & 1\\
        \hline
        PID controller & $\frac{1050s^2+5030s+3000}{s^2+100s}$ &$\frac{1049.9995s^2+5029.9949s+2999.9982}{s^2+100s}$  & 17 & 1 & $\frac{1049.9991s^2+5029.9849s+2999.9973}{s^2+100s}$ & $\frac{1049.9991s^2+5029.9849s+2999.9973}{s^2+100s}$ & 17 & 1\\
        \hline
    \end{tabular}}
    \end{center}
\end{table*}

\subsection{Identification Results in Voltage Regulator}
\label{app:volt_control}
\begin{figure}[t]
    \centering
    \includegraphics[scale=0.4]{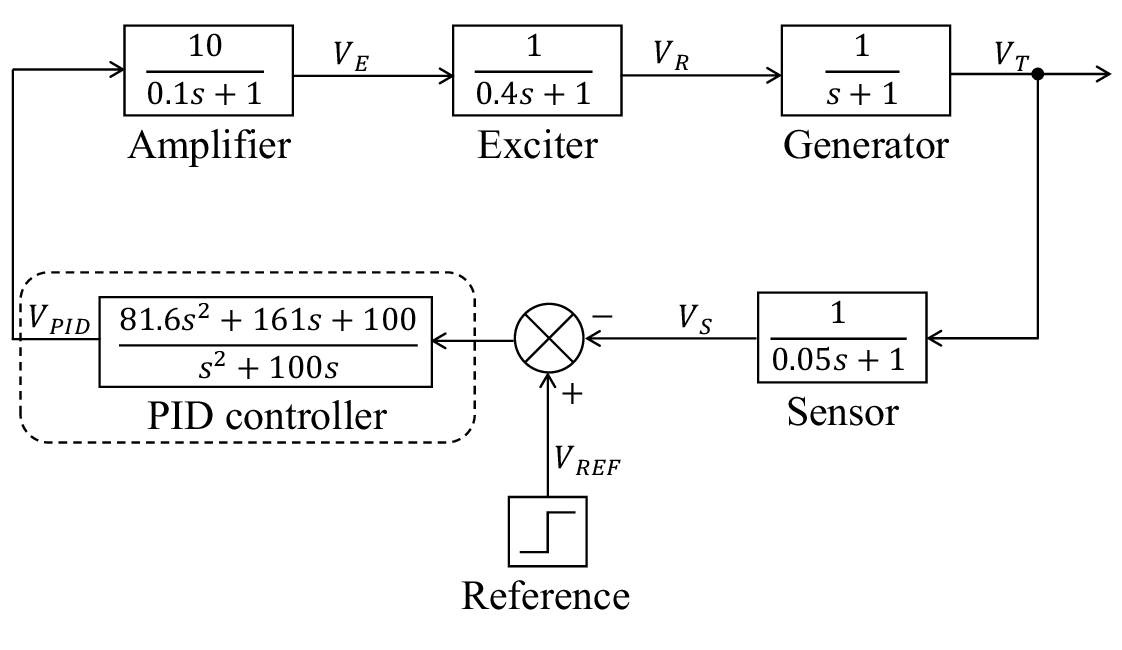}
    \caption{Framework of the voltage regulator system, comprising five components: amplifier, exciter, generator, sensor, and controller.}
    \label{fig_frawmework_voltPID}
\end{figure}
\begin{figure}[t]
    \centering
    \subfloat[Frequency signals in load frequency control system.]{
    \begin{minipage}[b]{0.45\textwidth}
    \includegraphics[width=1\linewidth]{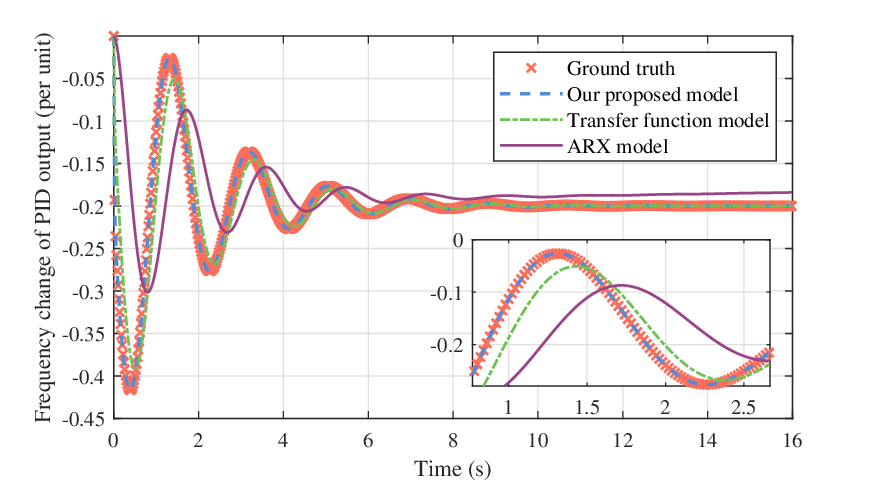}
    \end{minipage}}
    \hfill
    \subfloat[Voltage signals in voltage regulator system.]{
    \begin{minipage}[b]{0.45\textwidth}
    \includegraphics[width=1\linewidth]{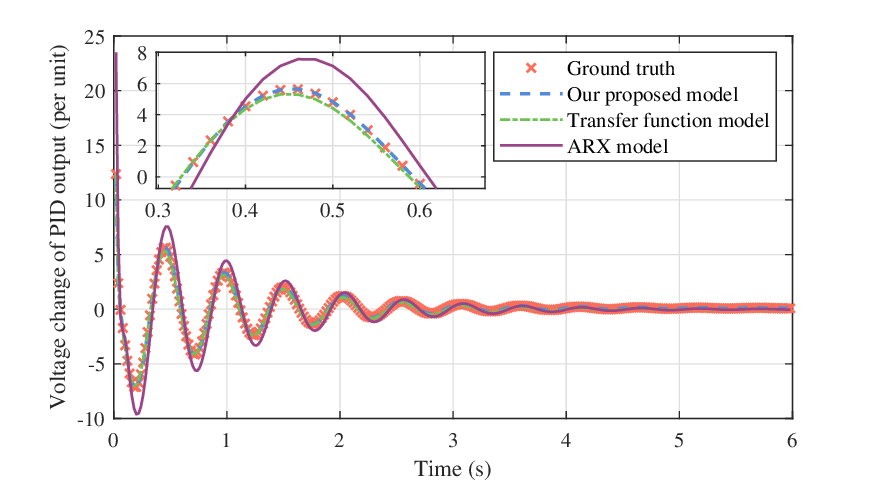}
    \end{minipage}}
	
    \caption{The comparison of the simulated frequency changes and voltage changes for PID output.}
    \label{fig_compare_power}
\end{figure}
\begin{figure*}[t]
    \centering
    \subfloat[Hysteresis Relay]{
    \begin{minipage}[b]{0.49\textwidth}
    \includegraphics[width=1\linewidth]{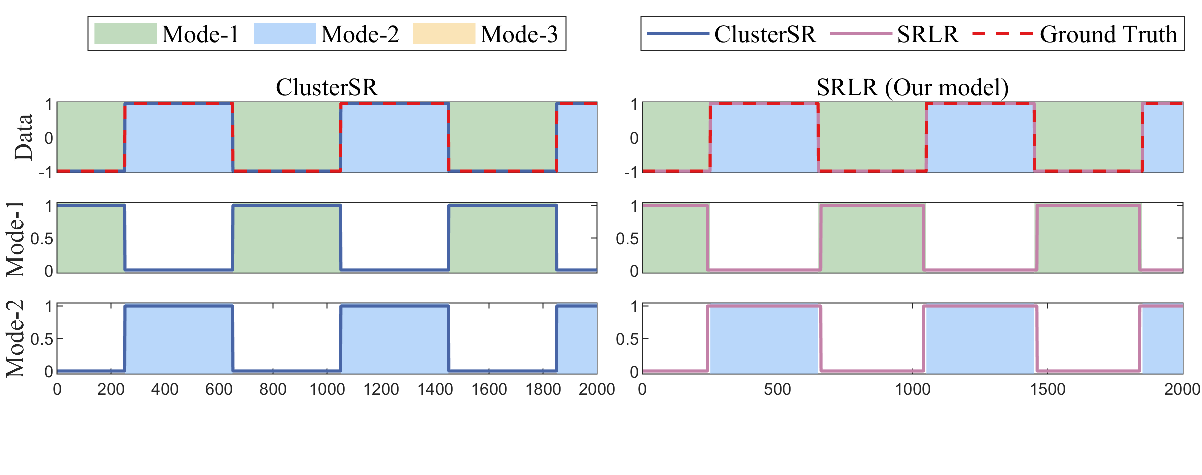}
    \end{minipage}}
    \subfloat[Phototaxic Robot]{
    \begin{minipage}[b]{0.49\textwidth}
    \includegraphics[width=1\linewidth]{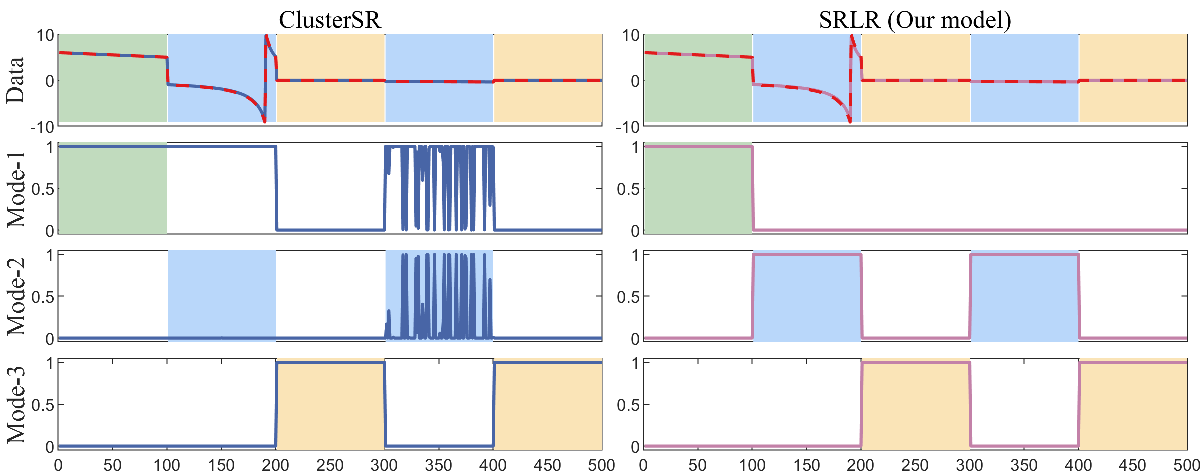}
    \end{minipage}}
	
    \caption{Comparison of identified index for Cluster SR and proposed method in Hysteresis Relay and Phototaxic Robot datasets. The first row shows simulated data from both methods alongside ground truth. Subsequent rows present the identified mode indices for each method.}
    \label{fig_compare_index_app}
\end{figure*}

\begin{table*}[htbp]
    \begin{center}
    \caption{The identification results of SRLR in voltage regulator system with/without AIC regularization.}
    \label{table_identified_voltage}
    \renewcommand{\arraystretch}{1.5}
    \resizebox{\textwidth}{!}{
    \begin{tabular}{ c|c|ccc|cccc }
        \hline
        & Ground truth mode & Identified mode(without AIC)&Complexity&BFR & Identified mode(with AIC) & simplified mode(with AIC) &Complexity&BFR\\
        \hline
        Amplifier&$\frac{10}{0.1s+1}$&$\frac{100.0656s-0.0039}{s^2+10.0009s}$& 20 & 0.9998 &$\frac{100.0674}{s+10.002}$ &$\frac{10}{0.0999s+0.9995}$& 13 & 0.9997\\
        \hline
        Exciter & $\frac{1}{0.4s+1}$ & $\frac{2.5022s-0.0001}{s^2+2.5005s}$ & 20 & 0.9994 & $\frac{2.5022}{s+2.5015}$ &$\frac{1}{0.3996s+0.9997}$& 13 & 0.9993\\
        \hline
        Generator & $\frac{1}{s+1}$ & $\frac{1.425}{4.912e-5s^2+1.4244s+1.425}$ & 16 & 0.9997 & $\frac{1}{s+1}$ & $\frac{1}{s+1}$ &13 & 1\\
        \hline
        Sensor & $\frac{1}{0.05s+1}$ & $\frac{-0.0136s+20.29}{s+20.2868}$ & 18 & 0.9990 & $\frac{20.01}{s+20.01}$ & $\frac{1}{0.0499s+1}$ & 13 & 0.9996\\
        \hline
        PID & $\frac{81.6s^2+161s+100}{s^2+100s}$ & $\frac{81.7941s^2+131.9117s+104.8382}{s^2+100.1764s}$ & 21 & 0.9883 & $\frac{81.57s^2+131.6s+104.5}{s^2+99.91s}$ & $\frac{81.57s^2+131.6s+104.5}{s^2+99.91s}$ & 17& 0.9888\\
        \hline
    \end{tabular}}
    \end{center}
\end{table*}

Another simulation involves automatic voltage regulator (AVR) system. Changes in reactive powers can influence the deviations of voltage magnitude. We model the reactive power using an amplifier, exciter, generator and sensor, with the formulations constructed as follows:
\begin{equation}
    \label{eq6}
    \begin{aligned}
    \frac{V_R(s)}{V_E(s)} &= \frac{K_A}{\tau_As+1},\ \  
    \frac{V_F(s)}{V_R(s)} &= \frac{K_E}{\tau_Es+1}\\
    \frac{V_T(s)}{V_F(s)} &= \frac{K_G}{\tau_Gs+1},\ \  
    \frac{V_S(s)}{V_T(s)} &= \frac{K_R}{\tau_Rs+1}
    \end{aligned}
\end{equation}
where the first equation represents an amplifier in the excitation system. The second equation is the exciter model, which is a linearized model disregarding saturation or other nonlinearities. The third equation is the generator and the fourth equation is a sensor model, where the sensed voltage is rectified through a bridge rectifier. The parameters $K_A, K_E, K_G, K_R$ are gains and $\tau_A, \tau_E, \tau_G, \tau_R$ are time constants.

A simulation is conducted as illustrated in Figure~\ref{fig_frawmework_voltPID}, with parameters configured according to the reference~\cite{akaike1974new}. Specifically, $K_A=10, \tau_A=0.1; K_E=1, \tau_E=0.4, K_G=1, \tau_G=1; K_R=1, \tau_R=0.05$. The voltage tends to be stable when PID is employed.
The results presented in Table~\ref{table_identified_voltage} demonstrate that our proposed model accurately identifies each part of the voltage regulator system. The application of complexity regularization proves to be superior, facilitating the model in generating more simple expression structures with high accuracy and mitigating the risk of overfitting.

\subsection{Comparison with Baselines}
We also conduct a comprehensive comparison between our proposed method and well-established system identification models. The first model is the Transfer Function Model, implemented using the Matlab System Identification Tool~\cite{ljung2009experiments}. This model initializes its parameters through the Instrument Variable method, and the Sequential Quadratic Programming algorithm is employed for optimal parameter search. We also include the autoregressive model with exogenous input (ARX), a widely used method for identifying linear-time invariant systems.

The detailed results of the logic recovery for the frequency control system and voltage regulator system are presented in Table~\ref{table_comparison_freq} and Table~\ref{table_comparison_volt}. The transfer function model, ARX model, and our proposed method demonstrate accurate identification of the physical parts in the control system. However, it's noteworthy that these baseline models struggle to identify the PID controller component in both voltage regulator and frequency control scenarios. The simulated PID outputs are depicted in Figure~\ref{fig_compare_power}(a) and Figure~\ref{fig_compare_power}(b), showcasing the precision of our proposed method in simulating frequency and voltage changes of PID controller outputs. While the transfer function model can fit the PID output with relatively small errors, the identified expressions show a significant disparity compared to the ground truth expressions. The ARX model performs less effectively in identifying the expressions and exhibits large errors in numerical simulations.

We compare SRLR with symbolic regression methods, including PySR and Operon. As shown in Figures~\ref{fig_baseline_lfc} and~\ref{fig_baseline_avc}, SRLR consistently outperforms these baselines. The gain subsystem is linear in the time domain, making it relatively easy for the baselines to identify. However, for other subsystems, the results differ. While PySR and Operon can adapt to nonlinear systems, they perform poorly in closed-form control systems such as PID controllers, as they do not incorporate designs like using the right domain, Use of continuity constraint, outlier-aware training and complexity regularization.

\begin{figure}[htbp]
    \centering
    \includegraphics[scale=0.42]{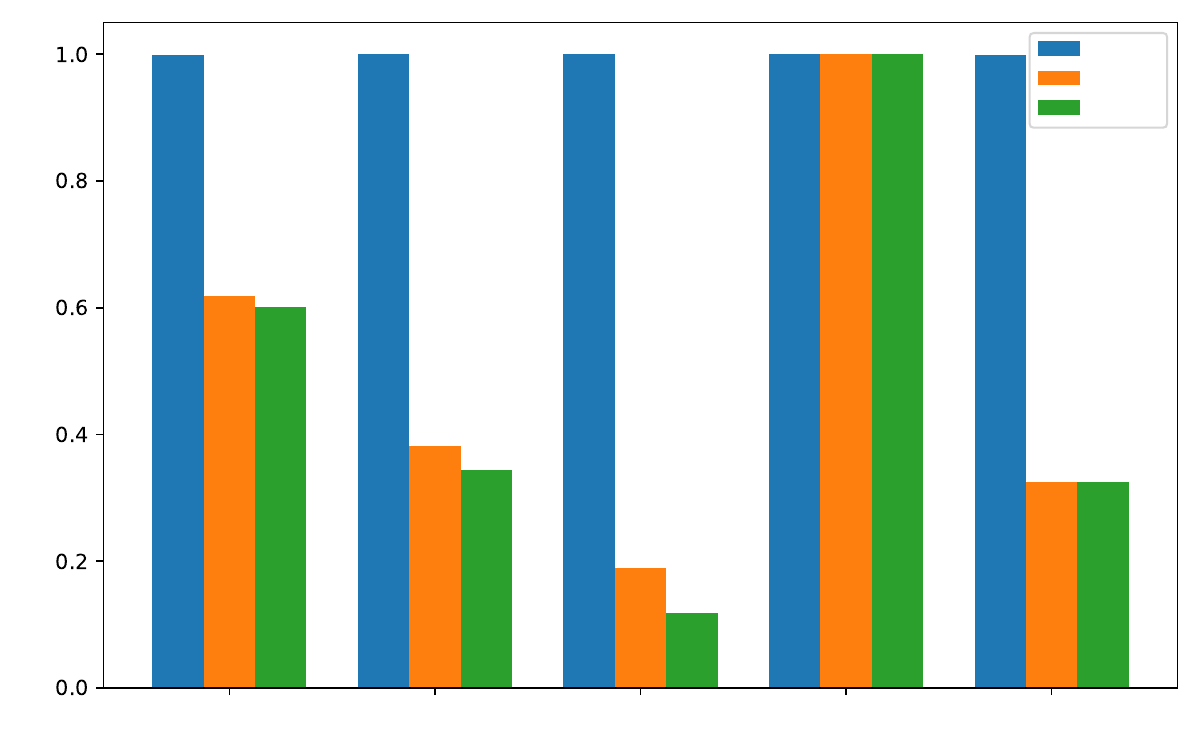}
    \caption{Performance comparison of SRLR and baselines on the load frequency control system.}
    \label{fig_baseline_lfc}
\end{figure}

\begin{figure}[htbp]
    \centering
    \includegraphics[scale=0.42]{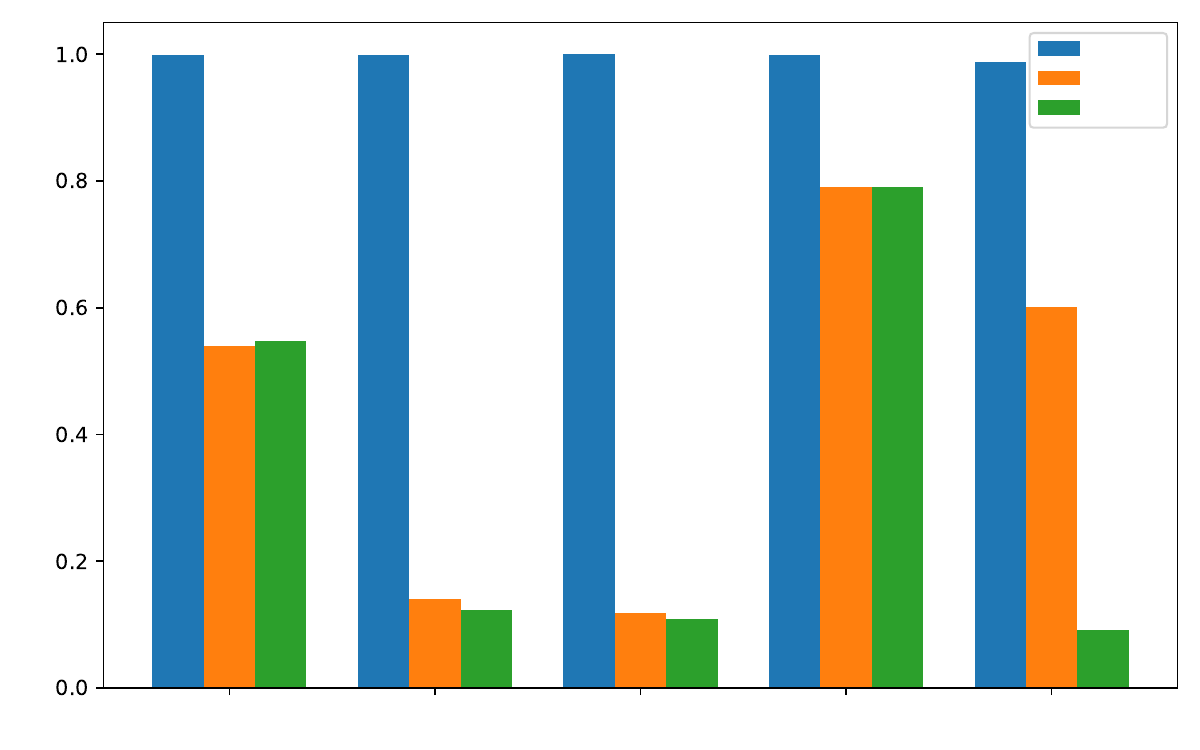}
    \caption{Performance comparison of SRLR and baselines on the automatic voltage regulator system.}
    \label{fig_baseline_avc}
\end{figure}

\begin{table*}[htbp]
    \begin{center}
    \caption{The comparison of SRLR and baselines for logic recovery in load frequency control system.}
    \renewcommand{\arraystretch}{1.5}
    \label{table_comparison_freq}
    \resizebox{\textwidth}{!}{
    \begin{tabular}{ c|c|cc|cc|cc }
        \hline
        \multirow{2}*{ } & \multirow{2}*{Ground truth mode} & \multicolumn{2}{|c|}{Transfer function model} & \multicolumn{2}{|c|}{ARX model} & \multicolumn{2}{|c}{SRLR(our model)}\\ \cline{3-8}
        & & Identified mode & BFR & Identified mode & BFR & Identified mode & BFR\\ 
        \hline
        Governor & $\frac{1}{0.2s+1}$ & $\frac{1}{0.1901s+1.0003}$ & 0.9814 & $\frac{1}{0.1898s+1.0004}$ & 0.9807 & $\frac{1}{0.1999s+1}$ & 0.9999\\
        \hline
        Turbine & $\frac{1}{0.5s+1}$ & $\frac{1}{0.4904s+1.0005}$ & 0.9893 & $\frac{1}{0.4895s+1}$ & 0.9885 & $\frac{1}{0.5s+1}$ & 1\\
        \hline
        Inertia & $\frac{1}{10s+0.8}$ & $\frac{1}{10.1215s+0.8078}$ & 0.9450 & $\frac{1}{10.1420s+0.8085}$ & 0.9373 & $\frac{1}{10s+0.8}$ & 1\\
        \hline
        Gain & -20 & -20 & 1 & -20 & 1 & -20 & 1\\
        \hline
        PID & $\frac{1050s^2+5030s+3000}{s^2+100s}$ & $\frac{482.6s^2+5060s+3000}{s^2+100s}$ & 0.5900 & $\frac{3.773s^2+199.1s+140.6}{s^2+4.748s+0.0217}$ & 0 & $\frac{1051.57s^2+5039.15s+3005.73}{s^2+100.41}$ & 0.9997\\
        \hline
    \end{tabular}
    }
    \end{center}
\end{table*}

\begin{table*}[htbp]
    \begin{center}
   \renewcommand{\arraystretch}{1.5}
    \caption{The comparison of SRLR and baselines for logic recovery in voltage regulator system.}
    \label{table_comparison_volt}
    \begin{tabular}{ c|c|cc|cc|cc }
        \hline
        \multirow{2}*{ } & \multirow{2}*{Ground truth mode} & \multicolumn{2}{|c|}{Transfer function model} & \multicolumn{2}{|c|}{ARX model} & \multicolumn{2}{|c}{SRLR(our model)}\\ \cline{3-8}
        & & Identified mode & BFR & Identified mode & BFR & Identified mode & BFR\\
        \hline
        Amplifier & $\frac{10}{0.1s+1}$ & $\frac{10}{0.0908s+1.0191}$ & 0.9567 & $\frac{10}{0.0915s+1.0201}$ & 0.9587 & $\frac{10.1}{0.1s+1}$ & 0.9997\\
        \hline
        Exciter & $\frac{1}{0.4s+1}$ & $\frac{1}{0.3908s+1.0191}$ & 0.9716 & $\frac{1}{0.3902s+1.0176}$ & 0.9716 & $\frac{1}{0.3996s+1}$ & 0.9993\\
        \hline
        Generator & $\frac{1}{1s+1}$ & $\frac{1}{0.994s+1.001}$ & 0.9931 & $\frac{1}{0.9911s+1.001}$ & 0.9898 & $\frac{1}{s+1}$ & 1\\
        \hline
        Sensor & $\frac{1}{0.05s+1}$ & $\frac{1}{0.0391s+1.0004}$ & 0.9548 & $\frac{1}{0.0392s+1.0004}$ & 0.9552 & $\frac{1}{0.0499s+1}$ & 0.9996\\
        \hline
        PID & $\frac{81.6s^2+161s+100}{s^2+100s}$ & $\frac{81.6s^2+1.22e^{18}s+1.04e^{18}}{s^2+1.8e^{16}s+1.5e^{18}}$ & 0.9742 & $\frac{38.03s^2+2788s+16510}{s^2+73.14s+3720}$ & 0.4731 & $\frac{81.57s^2+131.6s+104.5}{s^2+99.91s}$ & 0.9888\\
        \hline
    \end{tabular}
    \end{center}
\end{table*}
\subsection{Additional Results in Multi-Mode Logic Recovery}
\label{app:multi_mode_recovery}
To evaluate the identification accuracy of the index for each mode, we define the index accuracy as follows:
\begin{equation}
    \label{eq_index_acc}
    \text{Index Accuracy} = \frac{1}{T}\sum_{t=1}^T{(\mathbb{I}_{ind_t=\hat{ind}_t})}
\end{equation}
where $\mathbb{I}$ is an indicator varaible: if the predicted mode $ind_t$ at time step $t$ is correct, $\mathbb{I}$ is 1; otherwise, $\mathbb{I}$ equals zero.\\ 

The details of the baselines are described as follows:\\
\begin{itemize}
    \item \textit{Cluster SR}~\cite{ly2012learning}: A clustering symbolic regression approach that utilizes GP as the backbone for symbolic regression. This method iteratively optimizes soft membership values for each point and the GP parameters through the expectation-maximization algorithm, employing a weighted error summation as the fitness function.

\item \textit{Cluster DSR}: A comparative approach that substitutes GP with Deep Symbolic Regression~\cite{petersen2021deep} within the Cluster SR model, maintaining consistency in all other configurations.

\item \textit{LSTM Network}~\cite{ljung2020deep}: A LSTM based regression method that comprises one LSTM layer and two linear projection layers to predict the next system output. The number of hidden units is set to 8, as the default configuration in~\cite{ljung2020deep}, and the input data is segmented using a sliding window with size 100.

\item \textit{Deep Cascaded Network}~\cite{ljung2020deep}: A Neural Network based regression method with 7 cascaded feedforward networks and each with ReLU activation layers, which is used to predict the next system output. The sliding window size is similarly set to 100.\\
\end{itemize}

\begin{table*}[htbp]
    \begin{center}
    \caption{The identification results of SRLR in multi-mode process systems.}
    \label{table_hybrid_system}
    \resizebox{\textwidth}{!}{
    \begin{tabular}{ c|c|c|c|c|c|c|c }
        \hline
        Dataset & Mode ID & The ground truth mode & No. of points & Identified mode & Complexity & BFR & Index accuracy\\
        \hline
        \multirow{2}*{Hysteresis Relay} & 1 & $y=1$ & 1200 & $y=1$ & 1 &1.0 & \multirow{2}*{100\%} \\
        & 2 & $y=-1$ & 1200 & $y=-1$ & 6 & 1.0 &\\
        \hline
        \multirow{2}*{Continuous Hysteresis} & 1 & $y=0.5x^2+x-0.5$ & 2000 & $y=0.4999x^2+x-0.4999$ & 11 & 0.9999 & \multirow{2}*{99\%}\\
        & 2 & $y=-0.5x^2+x+0.5$ & 2000 & $y=-0.5x^2+x+0.5$ & 15 & 1.0 &\\
        \hline
        \multirow{2}*{Phototaxic Robot} & 1 & $y=x_2-x_1$ & 840 & $y=x_2-x_1$ & 7 & 1.0 & \multirow{3}*{100\%}\\
        & 2 & $y=1/(x_1-x_2)$ & 1500 & $y=1/(x_1-x_2)$ & 10 & 1.0 &\\
        & 3 & $y=0$ & 1200 & $y=0$ & 1 & 1.0 &\\
        \hline
        \multirow{2}*{Non-linear System} & 1 & $y=x_1x_2$ & 1500 & $y=x_1x_2$ & 7 & 1.0 & \multirow{3}*{98.51\%}\\
        & 2 & $y=6x_1/(6+x_2)$ & 1000 & $y=6x_1/(6+x_2)$ & 11 & 1.0 &\\
        & 3 & $y=(x_1+x_2)/(x_1-x_2)$ & 1000 & $y=(x_1+x_2)/(x_1-x_2)$ & 17 & 0.9996 &\\
        \hline
    \end{tabular}
    }
    \end{center}
\end{table*}

As depicted in Figure~\ref{fig_compare_index_app}, both our SRLR and Cluster SR successfully identified the discrete hysteresis for the switch-off and switch-on modes, as each mode exhibits a constant behavior that is relatively straightforward to discern. However, it is essential to note that the point distribution within each mode possesses local continuity, a characteristic overlooked by Cluster SR, which does not account for the temporal dynamics of systems. Thus, in the case of phototaxic robot, Cluster SR fails to identify the indexes. 

{We also compare the runtime performance of SRLR against the baseline methods across four multi-mode process systems. All experiments use the same amount of training data, with a fixed maximum of 200,000 expressions processed each method. Runtime is measured until all modes are successfully recovered. As shown in Table~\ref{table_runtime}, SRLR achieves the fastest performance, attributed to its effective mode membership assignment and accurate model identification. While DSR demonstrates better recovery than GP, its Cluster DSR variant suffers from less accurate mode membership assignment. As a result, Cluster SR slightly outperforms Cluster DSR, highlighting the critical role of precise mode membership and expression identification.}

\begin{table}[htbp]
    \centering
    \caption{Average Runtime Comparison}
    \label{table_runtime}
    \begin{tabular}{l|ccc}
    \toprule
    System & SRLR & Cluster SR & Cluster DSR \\
    \midrule
    Hysteresis Relay & 205.92 s & 414.34 s & 4916.48 s \\
    Continuous Hysteresis & 642.70 s & 3537.18 s & 3131.23 s \\
    Phototaxic Robot & 494.94 s & 7490.50 s & 10317.71 s \\
    Non-linear System & 3839.66 s & 8598.47 s & 7144.81 s \\
    \bottomrule
    \end{tabular}
\end{table}

\subsection{{Additional Results in Switch Linear Systems Identification}}
\begin{figure}[htbp]
    \centering
    \includegraphics[scale=0.35]{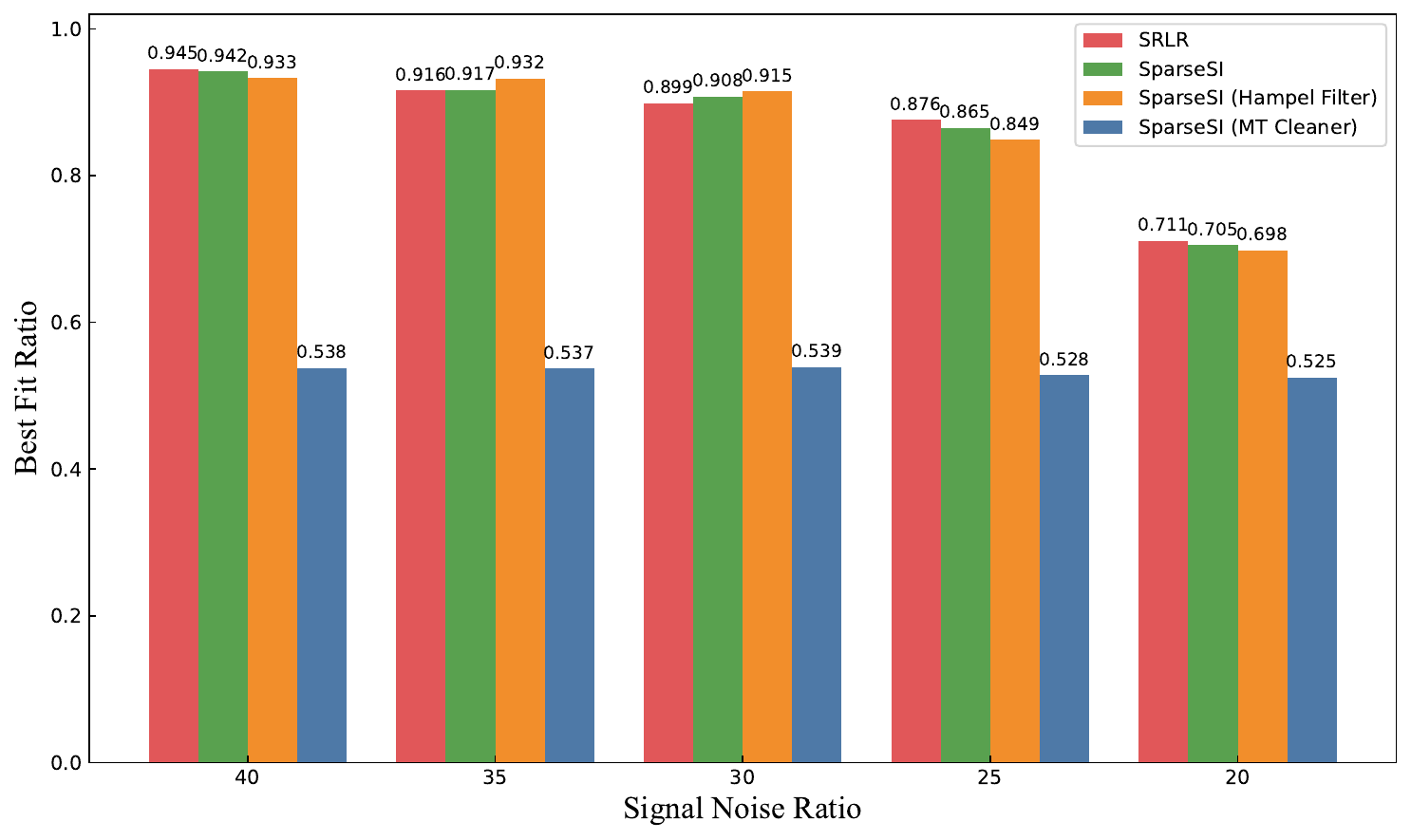}
    \caption{Performance comparison between SRLR and sparse optimization-based identification.}
    \label{fig_compare_with_sparse_si2}
\end{figure}

{System identification for switched linear systems shares similarities with the identification of multi-mode control logic, as both approaches yield mathematical representations for distinct system stages. In our study, we simulate a switched linear system comprising three linear submodels, defined by the following equation:}
\begin{equation}
    \label{switch_linearsys}
    y(t)=\theta^T_{k}[y(t-1),x(t-1),x(t-2)]^T+e(t)].
\end{equation}
{where the parameters for each submodel are: $\theta_{1}=[-0.277, -1.779, 1.154], \theta_{2} = [-0.747, -1.816, 0.707] , \theta_{3} = [-0.376, 1.803, 0.928]$. The noise term $e(t)$ is a white Gaussian noise, with a Signal-to-Noise Ratio (SNR) ranging from 40 dB to 20 dB, relative to the output signal.}

{We compare the performance of our proposed SRLR, with a sparse optimization-based system identification method (SparseSI)~\cite{bako2011identification}. To evaluate both methods, we generate synthetic data for each subsystem, totaling 3,600 data points for training. On average, SRLR requires approximately one hour for model identification, while SparseSI completes the process in about one second. This difference is expected, as SRLR includes a neural network training phase, which is inherently more computationally intensive. SparseSI, by contrast, is optimized for linear systems and is thus more efficient in these scenarios.}

{As illustrated in Figure~\ref{fig_compare_with_sparse_si2}, both methods achieve strong BFR performance under high SNR conditions. When combined with Hampel filtering, SparseSI’s accuracy improves slightly. However, applying the MT Cleaner significantly degrades its performance. This drop occurs because MT Cleaner uses a regressor to fit the dataset. Since the data is randomly generated, the regressor cannot generalize well, often distorting normal data points instead of isolating anomalies. This outcome underscores that data cleaning methods may not always yield improvements and must be chosen carefully based on data characteristics.}

{Although SRLR incurs a higher computational cost, it does not rely on predefined structural assumptions about the system, unlike SparseSI. This flexibility allows SRLR to be applied to a broader range of systems, including nonlinear ones. Therefore, while SRLR may require more processing time, its adaptability offers a significant advantage for modeling complex or unknown system behaviors.}

\section*{Appendix III: Practical Implications}
\subsection{Overhead Evaluation}
\begin{figure}[htbp]
    \centering
    \includegraphics[scale=0.3]{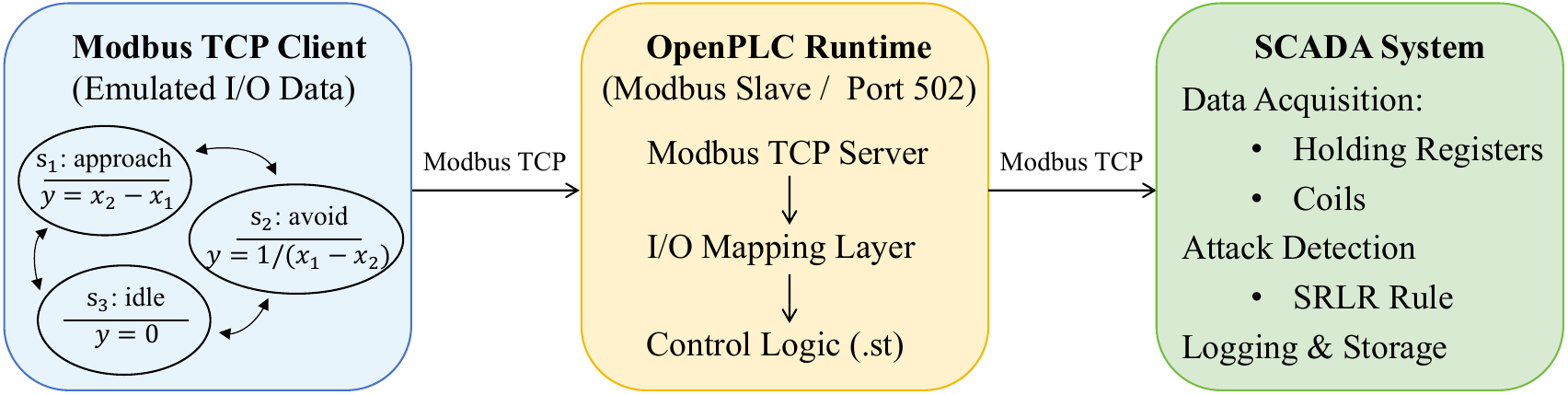}
    \caption{The overall framework of simulated PLC plantform.}
    \label{fig_openplc_sim}
\end{figure}

To evaluate the overhead of deploying SRLR in real-world industrial environments, we implemented a software-based PLC platform using the open-source OpenPLC project~\footnote{https://autonomylogic.com}. OpenPLC adheres to the IEC 61131-3 standard, providing standard hardware and software functionalities of real PLCs.

Our test framework, illustrated in Figure~\ref{fig_openplc_sim}, comprises three components: a Modbus TCP client, the OpenPLC runtime, and a SCADA system. The Modbus client simulates randomized sensor values. These values are read by the OpenPLC runtime, which emulates the phototaxic robot system described in earlier experiments. The PLC input and output data are then collected by the SCADA system. SRLR-generated detection rules are deployed within the SCADA system to identify potential attack events.

The deployment of SRLR involves two main steps. The first is offline training, where the necessary logs and data are collected, and the DSR model is executed. This process takes approximately one hour. The resulting rules, along with the associated code, require about 2 MB of storage. In the second step, the trained rules are installed on an attack detection system, potentially integrated into a SCADA environment. Performance evaluation indicates that collecting a single sample in SCADA incurs an average communication delay of 1.44 milliseconds. The detection process itself, when using SRLR, adds an average delay of 1.91 milliseconds. Memory usage during detection was also assessed: over a 20-minute observation period, SRLR required only 67.6 KB to store collected data and detection logs.


\begin{table*}[!t]
    \centering
    \begin{threeparttable}
    \caption{Explainable Attack Detection via SRLR Rules: Case Studies}
    \label{tab_explanable_ad}
    \rowcolors{2}{white}{gray!10}
    
    \begin{tabularx}{\textwidth}{
    >{\raggedright\arraybackslash}m{3cm}  
    >{\raggedright\arraybackslash}m{3.3cm} 
    >{\raggedright\arraybackslash}X        
    >{\raggedright\arraybackslash}m{2cm}   
    }
    \rowcolor{gray!25}
    \textbf{Attack Type} & \textbf{Mutation Operation} & \textbf{Example (Before $\rightsquigarrow$ After)} & \textbf{Violations} \\
    
    \rowcolor{Green!10}
    Sensor Value Tampering & Scalar variable replacement & $x_1 := 1 \rightsquigarrow \colorbox{red!20}{$x_1 := 10$}$ & rule1, rule2\\

    \rowcolor{Blue!10}
    Actuator Value Tampering & Scalar variable replacement & $y := 0 \rightsquigarrow \colorbox{red!20}{$y := 1$}$ & rule3\\
    
    \rowcolor{Blue!10}
     & Arithmetic operator replacement & $\frac{1}{x_2 - x_1} \rightsquigarrow \colorbox{red!20}{$\frac{1}{x_2 + x_1}$}$ & rule2\\

    \rowcolor{Orange!10}
    Switch Logic Attack & Subroutine deletion & \texttt{IF $s_2$ AND $x_3$ THEN $s_2$ $\rightarrow$ $s_1$} $\rightsquigarrow$ \colorbox{red!20}{\texttt{...}} & rule4\\
    
    \rowcolor{Orange!10}
      & Logical operator replacement & \texttt{IF $s_2$ AND $x_5$ THEN $s_2$ $\rightarrow$ $s_3$} $\rightsquigarrow$ \colorbox{red!20}{\texttt{IF $s_2$ OR $x_5$ THEN $s_2 \rightarrow s_3$}} & rule5\\

    \rowcolor{Orange!10}
     & Logical connector insertion & \texttt{IF $s_2$ AND $x_5$ THEN $s_2 \rightarrow s_3$} $\rightsquigarrow$ \colorbox{red!20}{\texttt{IF $s_2$ AND $x_5$ OR $x_3$ THEN $s_2 \rightarrow s_3$}} & rule4, rule5\\
    \end{tabularx}

    \vspace{0.5em}
    \begin{tablenotes}[para,flushleft]
    \footnotesize
    \textbf{SRLR Rules:} \\
    rule1: in mode $s_1$, $y := x_2 - x_1$; rule2: in mode $s_1$, $y := \frac{1}{x_2 - x_1}$; rule3: in mode $s_3$, $y := 0$; rule4: if \texttt{$s_2$ AND $x_3$}, mode switch $s_2 \rightarrow s_1$ \\
    rule5: if \texttt{$s_2$ AND $x_5$}, mode switch $s_2 \rightarrow s_3$; other rules ...
    \end{tablenotes}
    
    \end{threeparttable}
    
\end{table*}

\subsection{Case Study: Explainability in Attack Detection}

Various AI explainability techniques have been developed to interpret black-box, deep learning models. For instance, LIME~\cite{ribeiro2016should} approximates complex models with locally interpretable ones, such as decision trees, to explain individual predictions. Similarly, SHAP~\cite{lundberg2017unified} and ALE~\cite{apley2020visualizing} estimate feature contributions to model outputs. While these methods offer insight at the feature level, they fall short of generating interpretable rules that capture system behavior.

More advanced approaches, such as DeepAID~\cite{han2021deepaid}, attempt to produce human-readable rules by contrasting normal behavior with detected anomalies. However, these methods still do not reconstruct underlying system logic in a way that supports root-cause analysis and operational response. In contrast, SRLR offers structured, rule-based explainability by modeling interactions among variables. This enables operators to trace the propagation of attacks through system logic, facilitating identification of tampered variables or compromised control logic.

To illustrate, consider the phototaxic robot system described earlier. Using SRLR, interpretable rules governing behaviors such as movement, obstacle avoidance, and idle states can be derived, along with their logical transitions. As summarized in Table~\ref{tab_explanable_ad}, attackers can exploit protocol vulnerabilities to manipulate PLC logic. For instance, by altering scalar values~\cite{langner2011stuxnet, geng2023control}, substituting arithmetic or logical operators~\cite{spenneberg2016plc}, deleting or injecting malicious instructions~\cite{di2018triton} to disrupt control flow. SRLR detects these manipulations by flagging corresponding rule violations. Security personnel can then pinpoint the affected variables or identify the relevant code segments in the PLC’s ST (structured text) files responsible for the anomaly. Thanks to SRLR’s explainable framework, tampered code and variables can be located more quickly, significantly reducing the response time to logic-level attacks and improving overall system resilience.

}

\end{document}